%% file: main.tex
\documentclass[10pt,twocolumn,letterpaper]{article}

\usepackage{cvpr}              

\usepackage{graphicx}
\usepackage{amsmath}
\usepackage{amssymb}
\usepackage{booktabs}
\usepackage{bm}
\usepackage{multirow}
\usepackage[export]{adjustbox}
\usepackage{makecell}
\usepackage{enumitem}
\usepackage[dvipsnames]{xcolor}
\makeatletter
\@namedef{ver@everyshi.sty}{}
\makeatother
\usepackage{tikz}
\usepackage{bbm}

\makeatletter
\def\@fnsymbol#1{\ensuremath{\ifcase#1\or \dagger\or \ddagger\or
   \mathsection\or \mathparagraph\or \|\or **\or \dagger\dagger
   \or \ddagger\ddagger \else\@ctrerr\fi}}
\makeatother

\makeatletter
\renewcommand\thefootnote{\textsuperscript{\@fnsymbol\c@footnote}} 
\DeclareRobustCommand\thanks[2][]{%
  \if\relax#1\relax
    \footnotemark
  \else%
    \protect\refstepcounter{footnote}\protect\label{#1} 
  \fi%
  \protected@xdef\@thanks{
    \@thanks\protect\footnotetext[\the\c@footnote]{#2}%
  }%
}
\makeatother

\usepackage[pagebackref,breaklinks,colorlinks]{hyperref}

\usepackage[capitalize]{cleveref}
\crefname{section}{Sec.}{Secs.}
\Crefname{section}{Section}{Sections}
\Crefname{table}{Table}{Tables}
\crefname{table}{Tab.}{Tabs.}

\def\cvprPaperID{2904} 
\def\confName{CVPR}
\def\confYear{2023}

\begin{document}

\include{defs.tex}

\title{NeuWigs: A Neural Dynamic Model for Volumetric Hair Capture and Animation}

\author{
Ziyan Wang$^{1,2}$~~~~
Giljoo Nam$^{2}$~~~~
Tuur Stuyck$^{2}$~~~~
Stephen Lombardi$^{3}$\ref{thx}~~~~
Chen Cao$^{2}$\\
Jason Saragih$^{2}$~~~~
Michael Zollhöfer$^{2}$~~~~
Jessica Hodgins$^{1}$\ref{thx}~~~~
Christoph Lassner$^{4}$\ref{thx}
\vspace{0.1cm} \\ 
$^{1}$Carnegie Mellon University~~~
$^{2}$Reality Labs Research~~~
$^{3}$Google~~~
$^{4}$Epic Games~~~
}
\thanks[thx]{Work done while at Meta.}

\maketitle


\begin{abstract}
   The capture and animation of human hair are two of the major challenges in the creation of realistic  avatars for the virtual reality.
   Both problems are highly challenging, because hair has complex geometry and appearance, as well as exhibits challenging motion.
   In this paper, we present a two-stage approach that models hair independently from the head to address these challenges in a data-driven manner.
   %
   %
   The first stage, state compression, learns a low-dimensional latent space of 3D hair states containing motion and appearance, via a novel autoencoder-as-a-tracker strategy.
   To better disentangle the hair and head in appearance learning, we employ multi-view hair segmentation masks in combination with a differentiable volumetric renderer.
   The second stage learns a novel hair dynamics model that performs temporal hair transfer based on the discovered latent codes.
   To enforce higher stability while driving our dynamics model, we employ the 3D point-cloud autoencoder from the compression stage for de-noising of the hair state.
   Our model outperforms the state of the art in novel view synthesis and is capable of creating novel hair animations without having to rely on hair observations as a driving signal.
   Project page is here \href{https://ziyanw1.github.io/neuwigs/}{https://ziyanw1.github.io/neuwigs/}.
\end{abstract}

\begin{figure}[h!tb]
\setlength\tabcolsep{0pt}
\renewcommand{\arraystretch}{0}
\centering
\begin{tabular}{ccc}
 \begin{tikzpicture}
    \node[anchor=south west,inner sep=0] (image) at (0,0) {\adjincludegraphics[width=0.16\textwidth, trim={0 {0.2\height} 0 {0.25\height}}, clip]{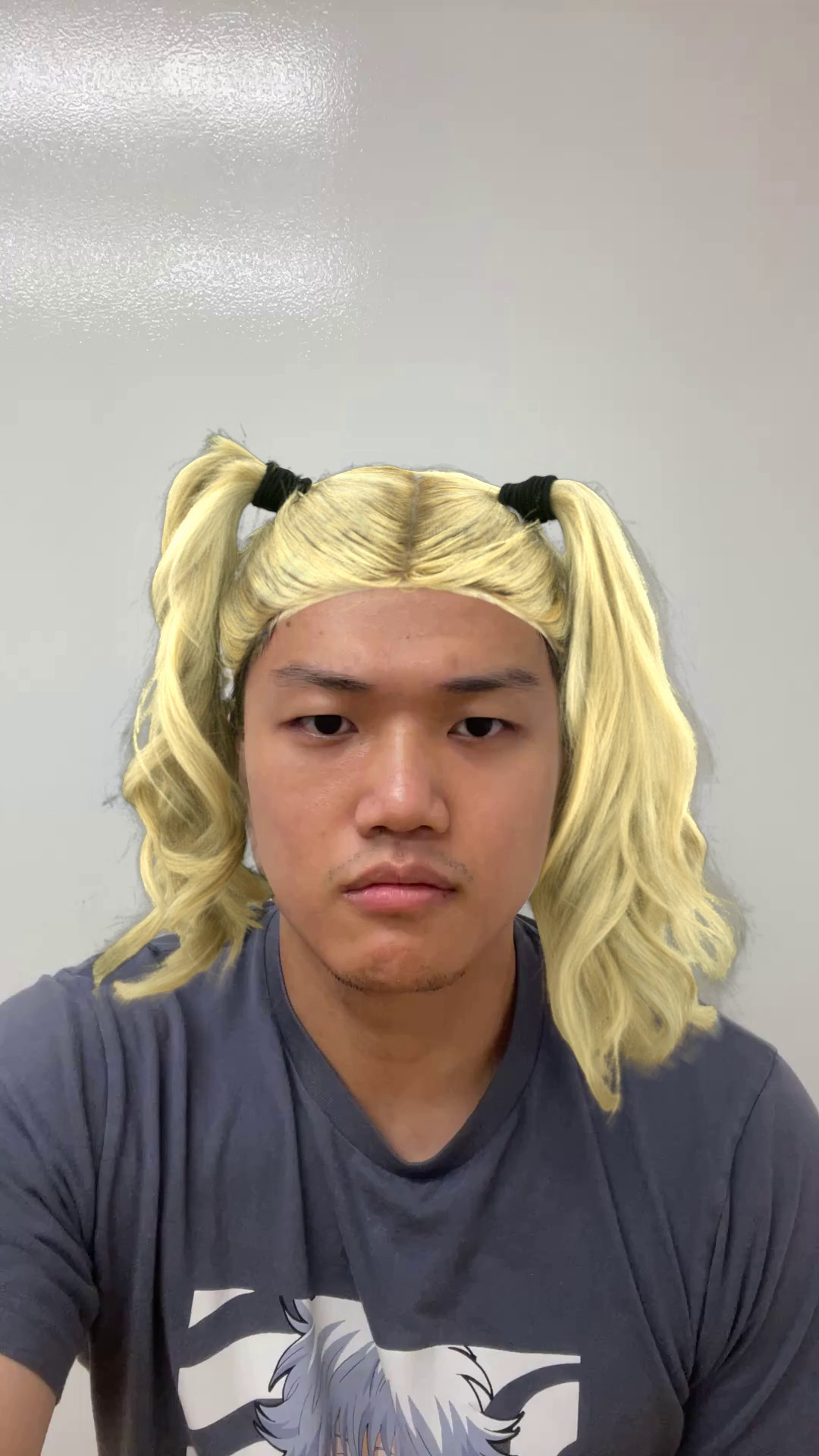}};
    \node[anchor=north west,inner sep=0] at (current bounding box.north west) {\setlength{\fboxsep}{0pt}\setlength{\fboxrule}{0.2pt}\fcolorbox{red}{white}{\adjincludegraphics[width=0.04\textwidth, trim={{0.2\width} {0.3\height} {0.2\width} {0.35\height}}, clip]{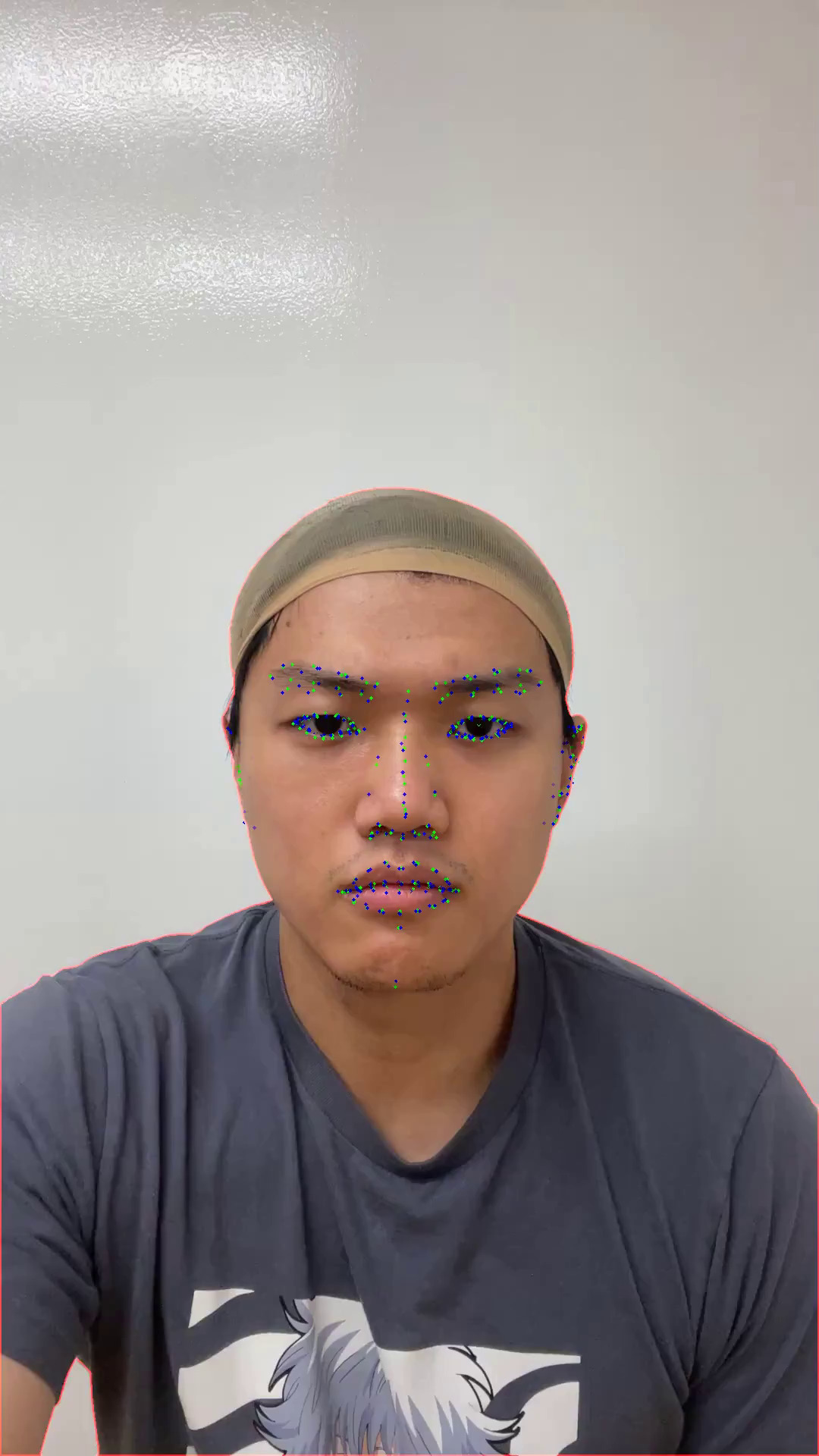}}};
 \end{tikzpicture}
 &
 \begin{tikzpicture}
    \node[anchor=south west,inner sep=0] (image) at (0,0) {\adjincludegraphics[width=0.16\textwidth, trim={0 {0.2\height} 0 {0.25\height}}, clip]{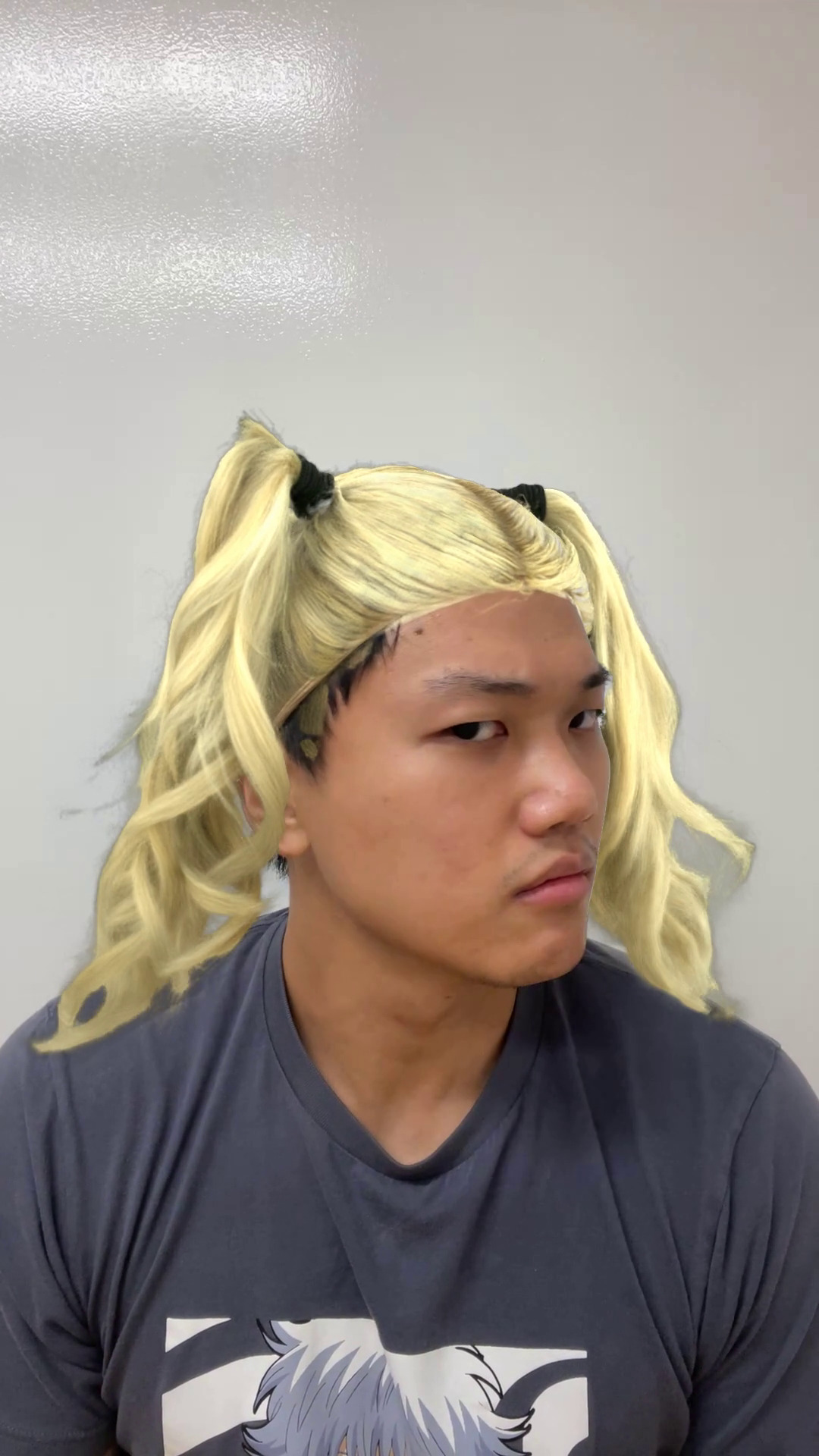}};
    \node[anchor=north west,inner sep=0] at (current bounding box.north west) {\setlength{\fboxsep}{0pt}\setlength{\fboxrule}{0.2pt}\fcolorbox{red}{white}{\adjincludegraphics[width=0.04\textwidth, trim={{0.2\width} {0.3\height} {0.2\width} {0.35\height}}, clip]{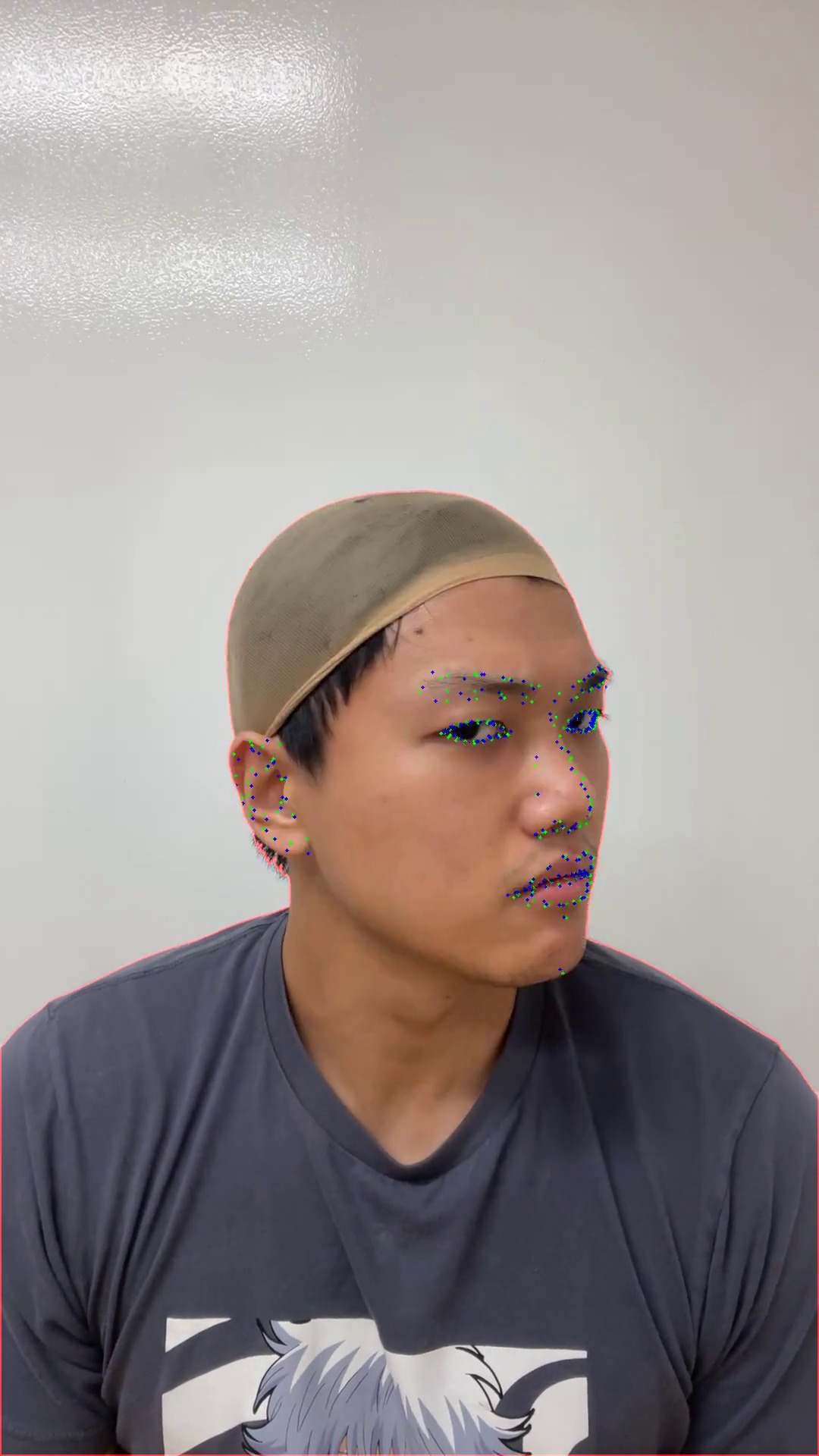}}};
 \end{tikzpicture}
 &
 \begin{tikzpicture}
    \node[anchor=south west,inner sep=0] (image) at (0,0) {\adjincludegraphics[width=0.16\textwidth, trim={0 {0.2\height} 0 {0.25\height}}, clip]{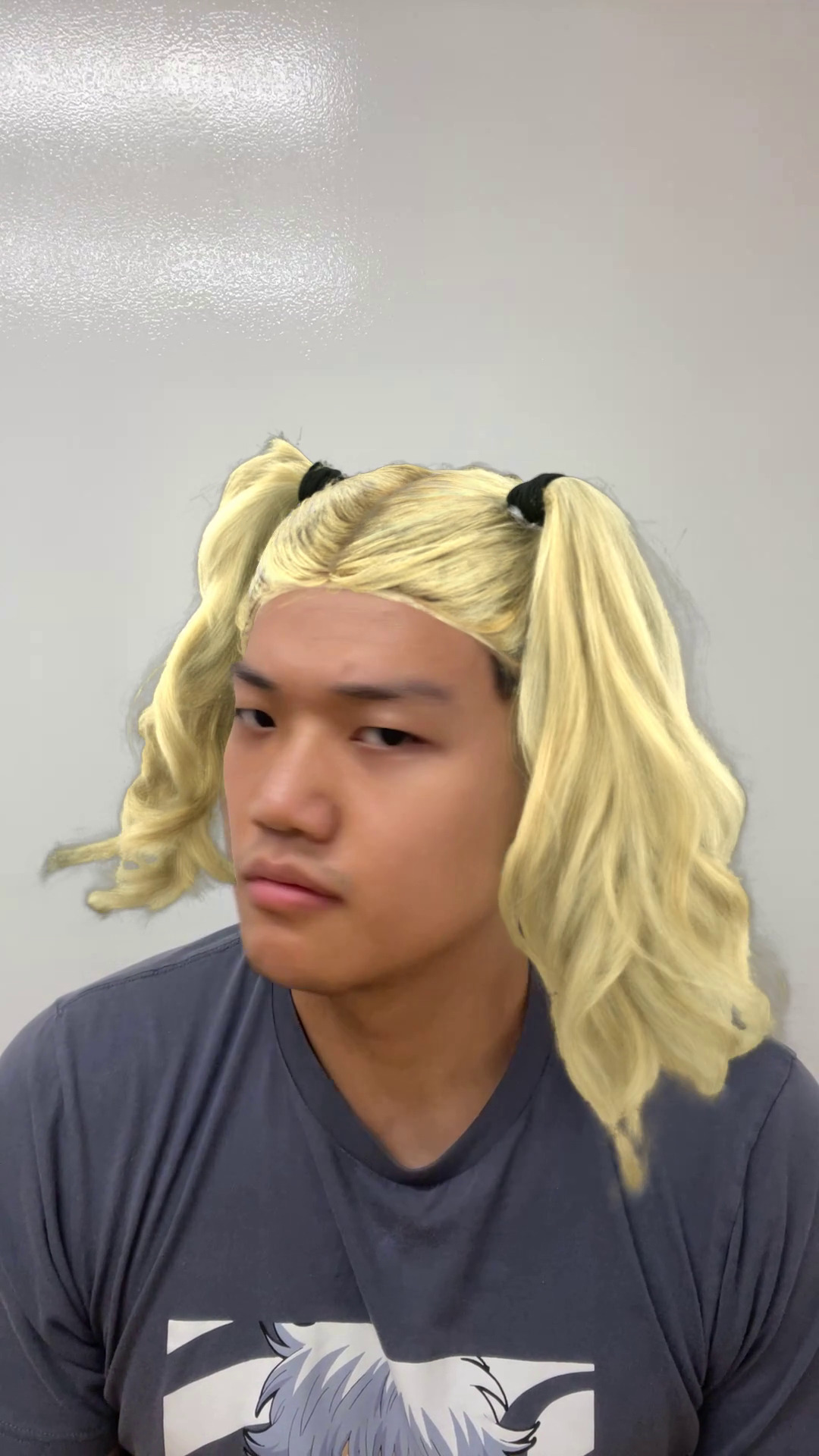}};
    \node[anchor=north west,inner sep=0] at (current bounding box.north west) {\setlength{\fboxsep}{0pt}\setlength{\fboxrule}{0.2pt}\fcolorbox{red}{white}{\adjincludegraphics[width=0.04\textwidth, trim={{0.2\width} {0.3\height} {0.2\width} {0.35\height}}, clip]{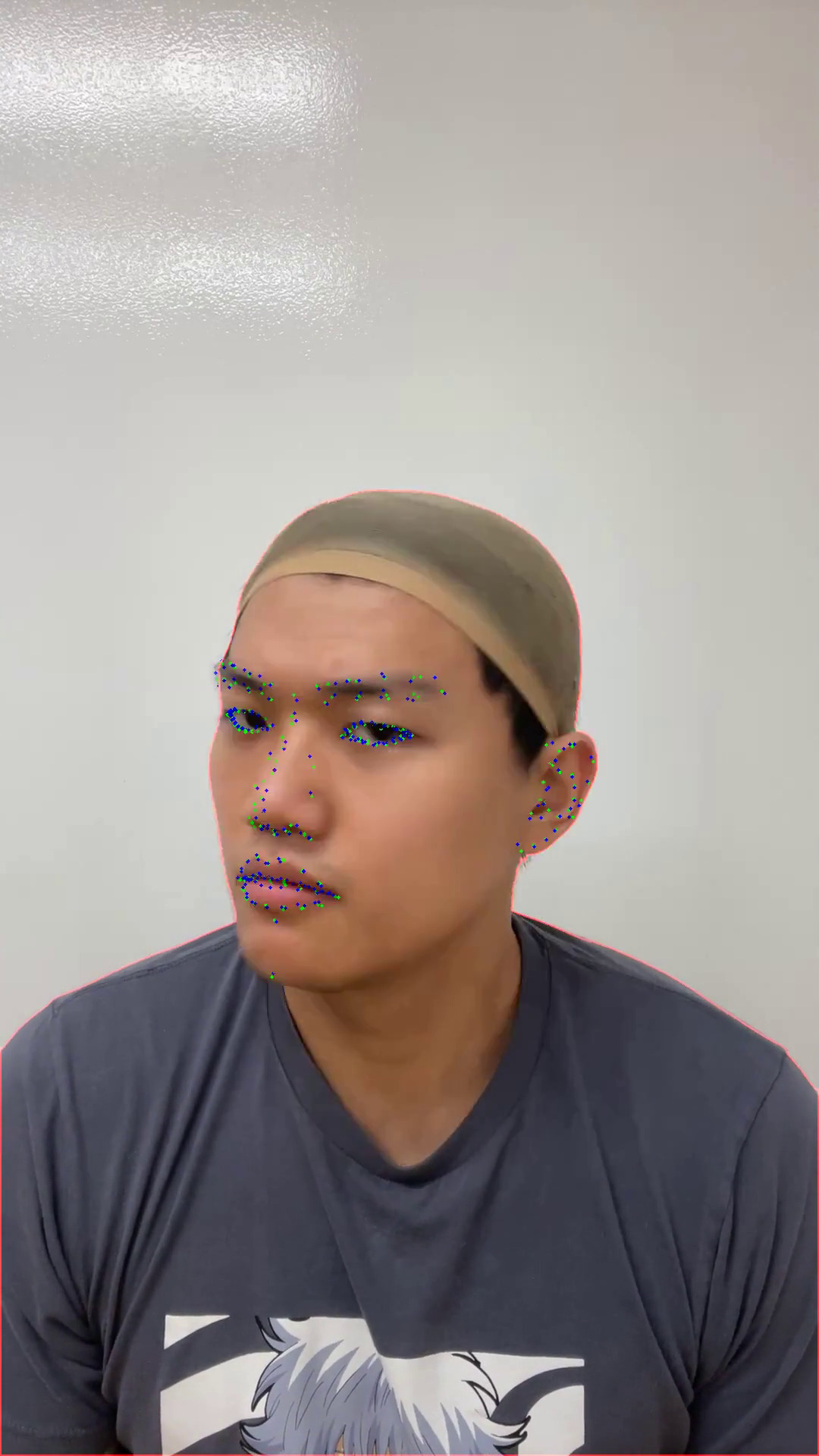}}};
 \end{tikzpicture} 
 \\
 \begin{tikzpicture}
    \node[anchor=south west,inner sep=0] (image) at (0,0) {\adjincludegraphics[width=0.16\textwidth, trim={0 {0.2\height} 0 {0.25\height}}, clip]{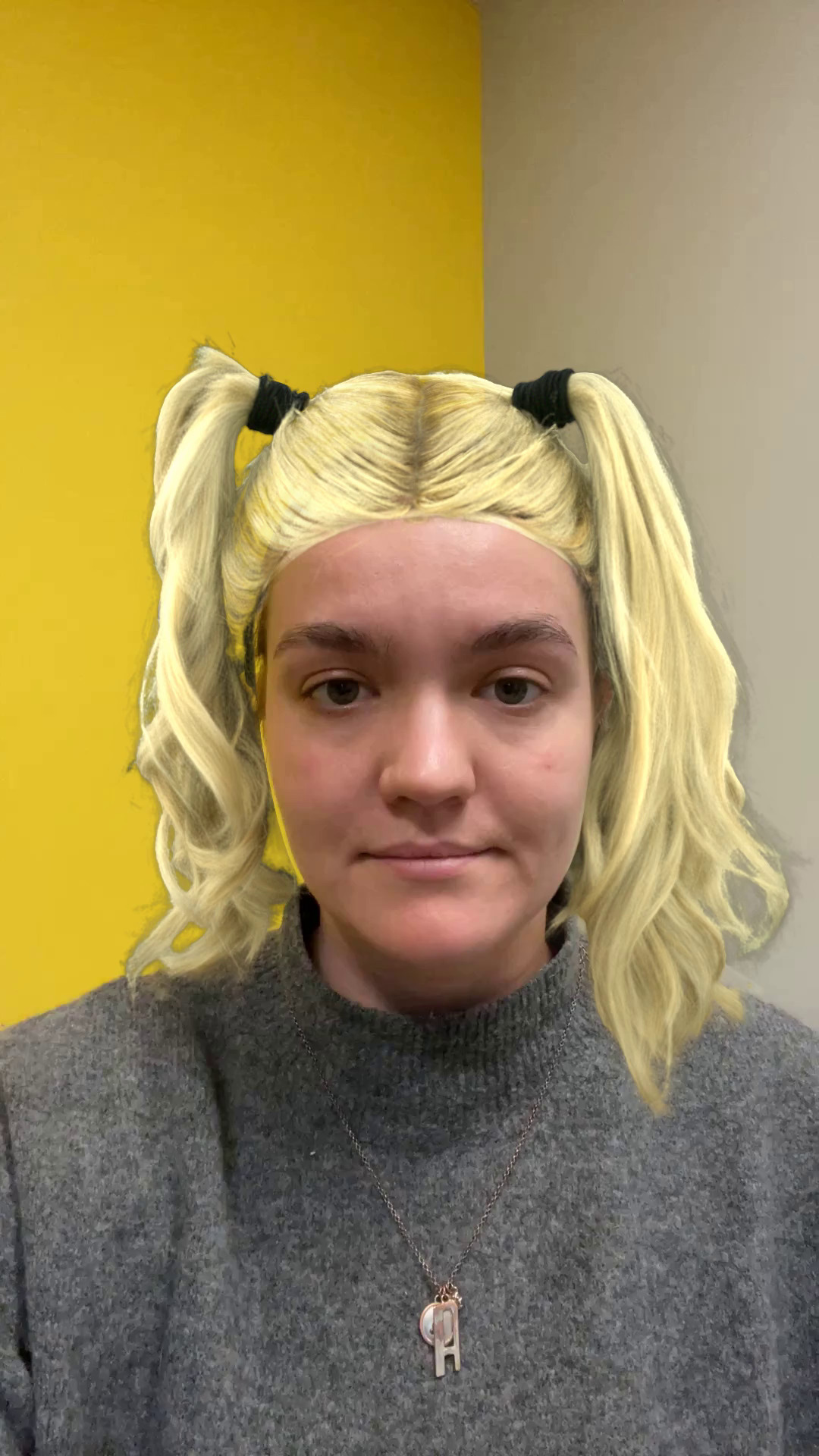}};
    \node[anchor=north west,inner sep=0] at (current bounding box.north west) {\setlength{\fboxsep}{0pt}\setlength{\fboxrule}{0.2pt}\fcolorbox{red}{white}{\adjincludegraphics[width=0.04\textwidth, trim={{0.15\width} {0.3\height} {0.15\width} {0.3\height}}, clip]{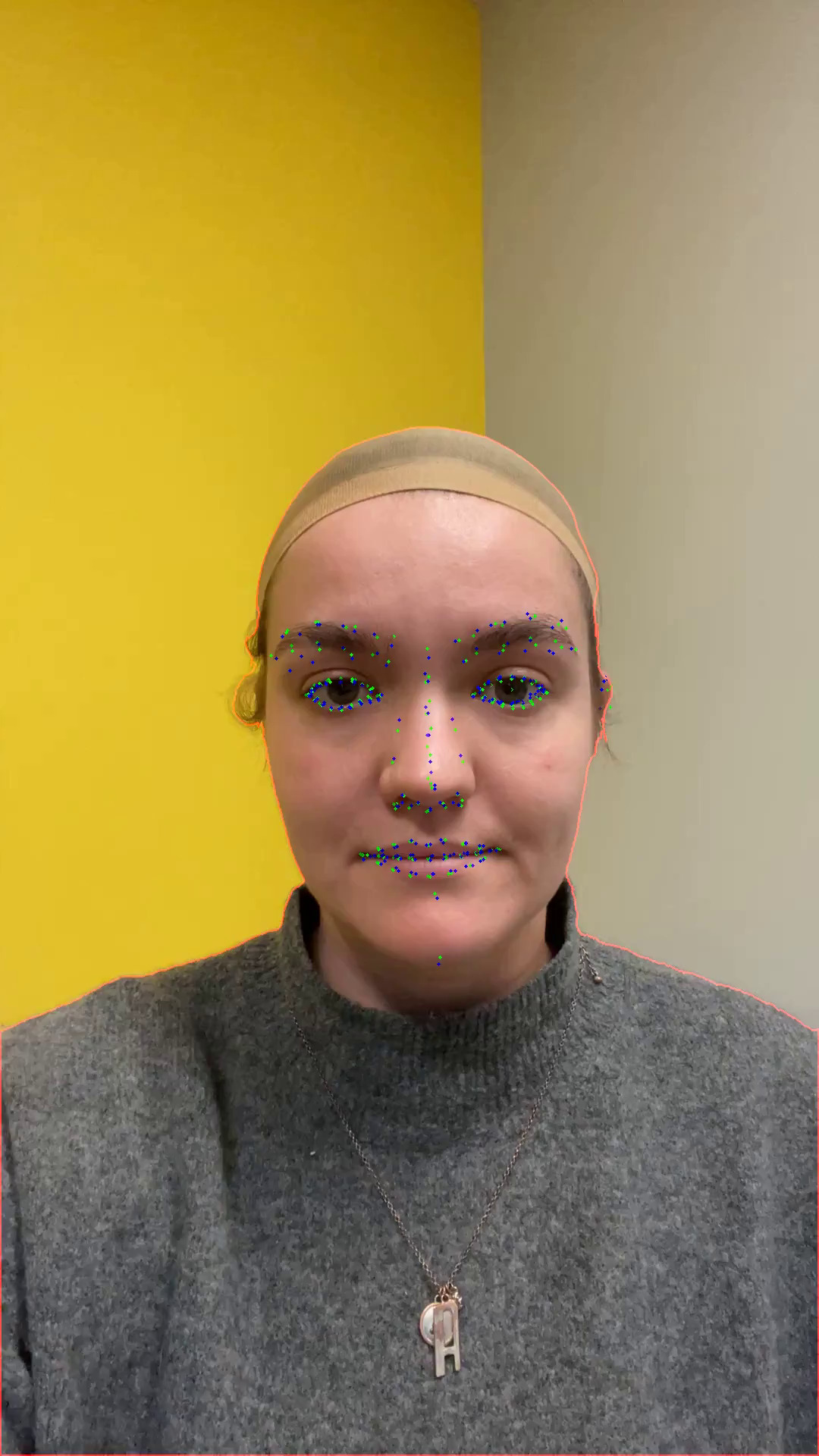}}};
 \end{tikzpicture} &
 \begin{tikzpicture}
    \node[anchor=south west,inner sep=0] (image) at (0,0) {\adjincludegraphics[width=0.16\textwidth, trim={0 {0.2\height} 0 {0.25\height}}, clip]{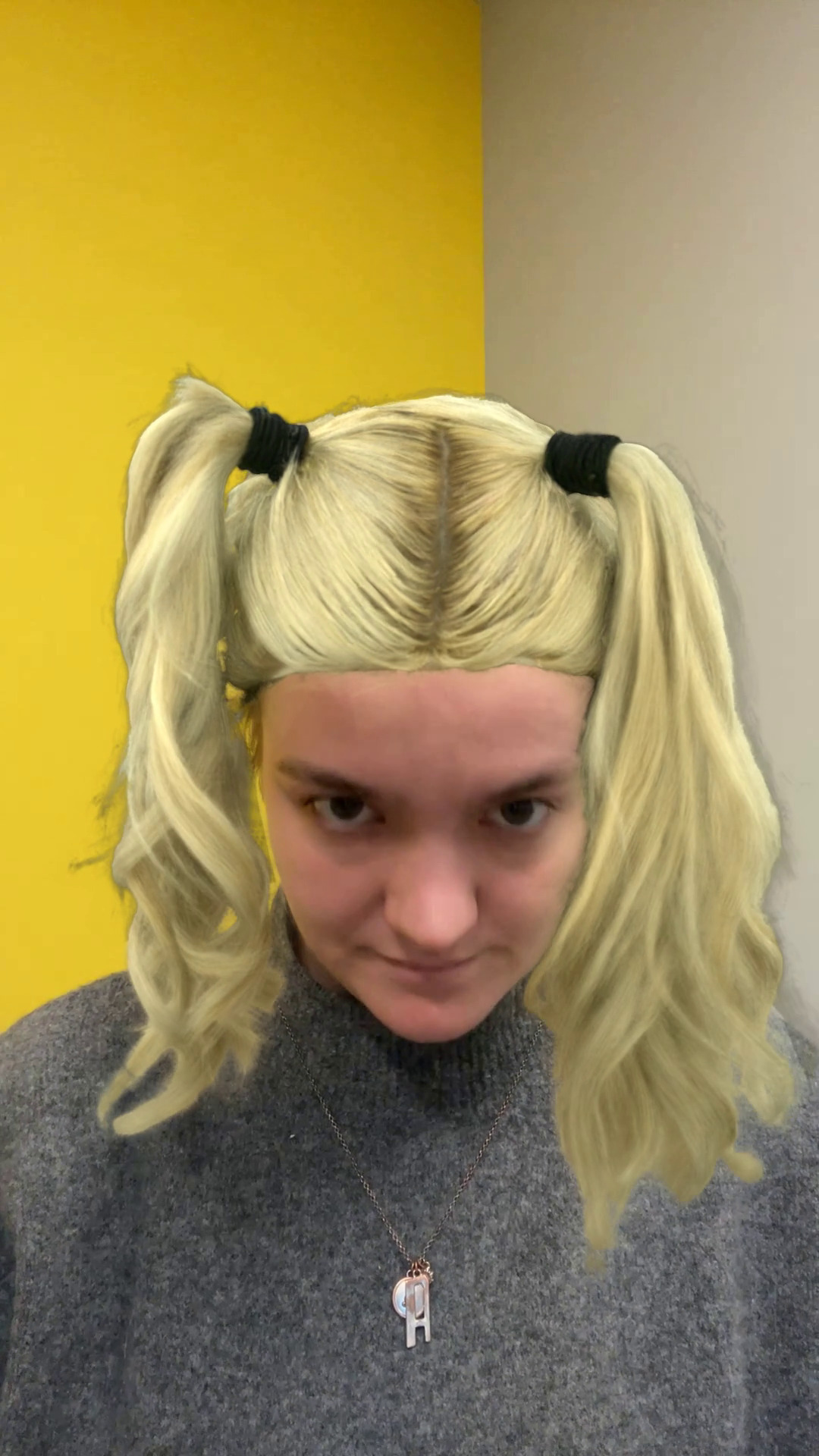}};
    \node[anchor=north west,inner sep=0] at (current bounding box.north west) {\setlength{\fboxsep}{0pt}\setlength{\fboxrule}{0.2pt}\fcolorbox{red}{white}{\adjincludegraphics[width=0.04\textwidth, trim={{0.15\width} {0.3\height} {0.15\width} {0.3\height}}, clip]{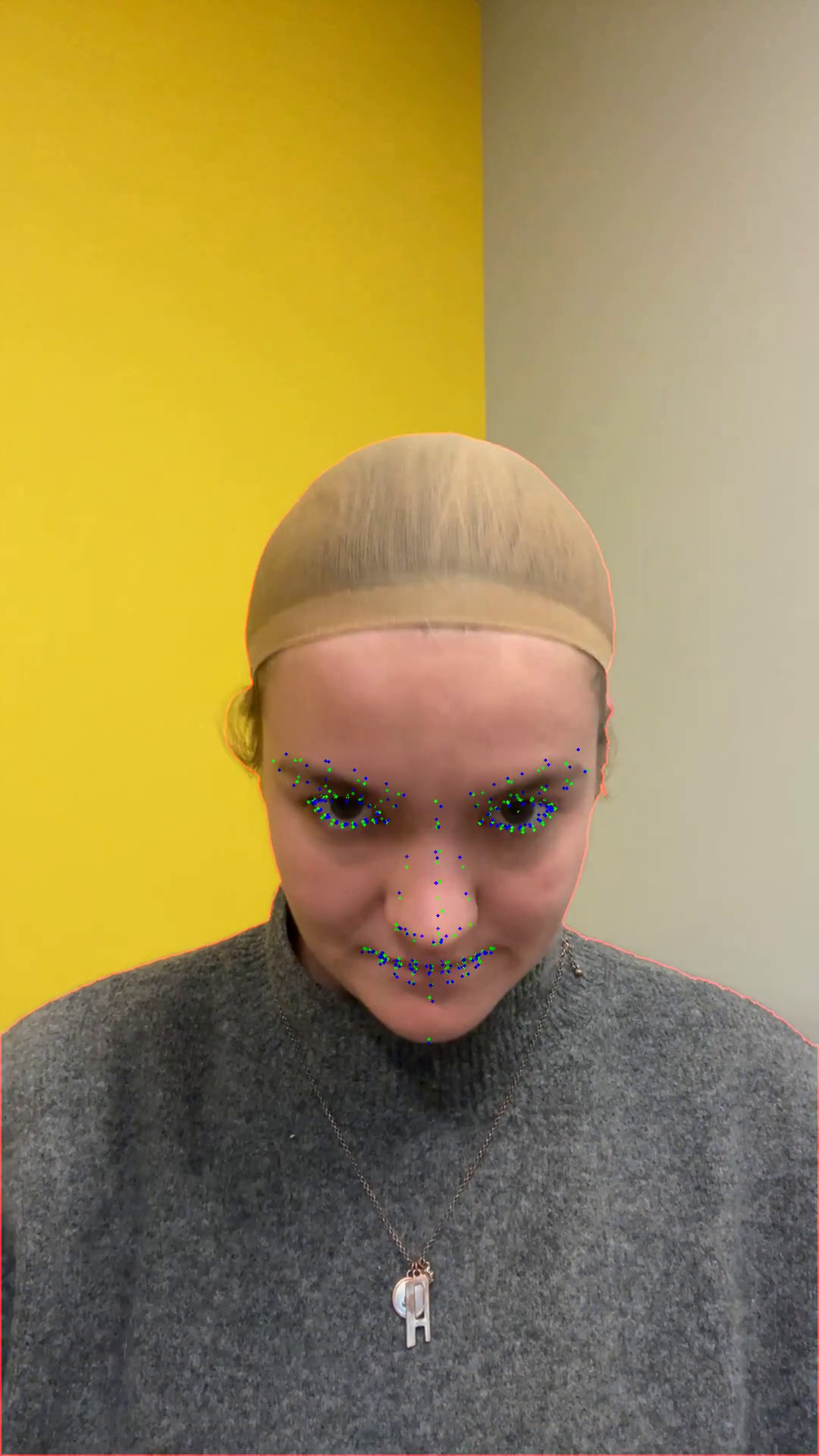}}};
 \end{tikzpicture} &
\begin{tikzpicture}
    \node[anchor=south west,inner sep=0] (image) at (0,0) {\adjincludegraphics[width=0.16\textwidth, trim={0 {0.2\height} 0 {0.25\height}}, clip]{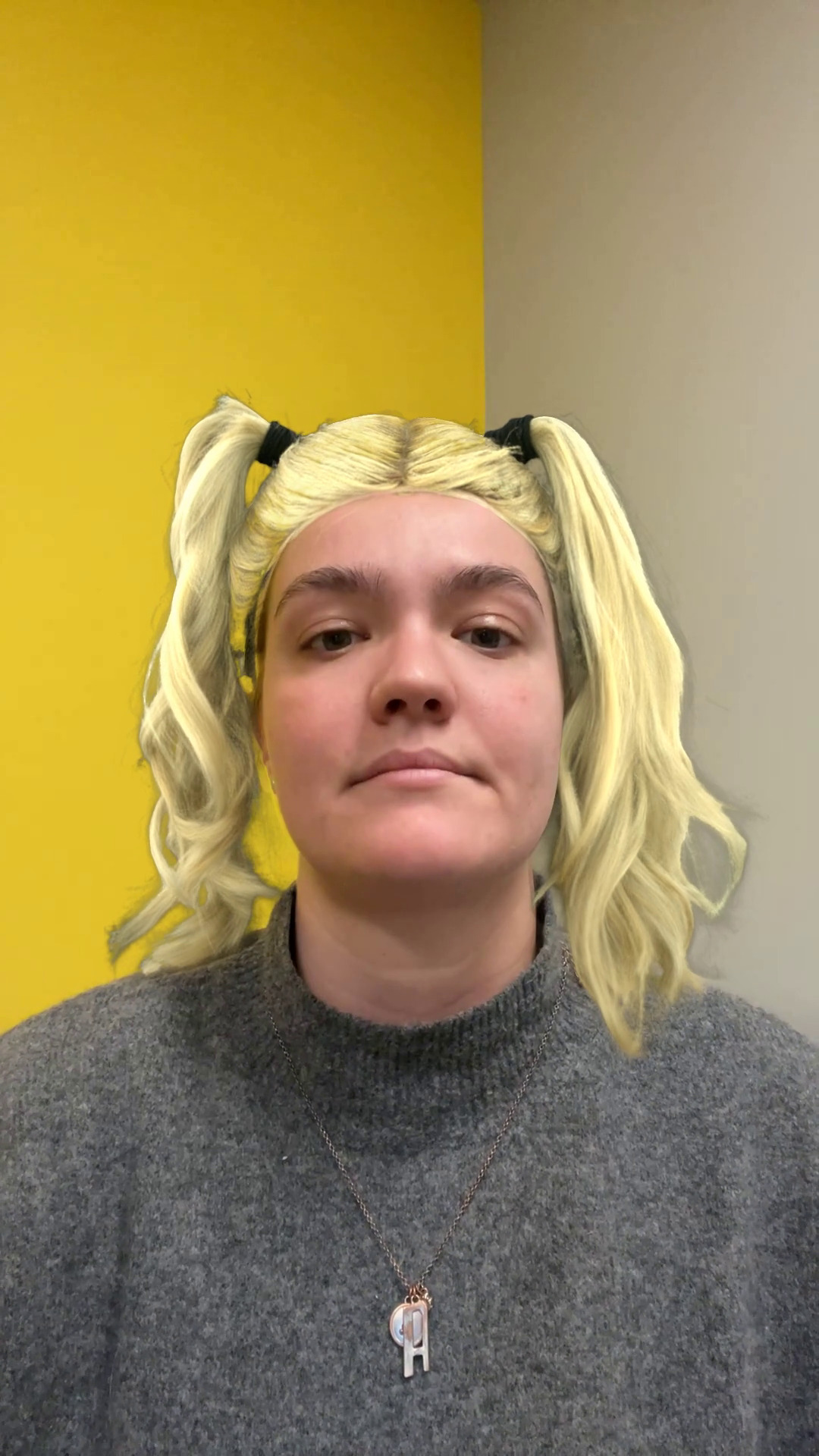}};
    \node[anchor=north west,inner sep=0] at (current bounding box.north west) {\setlength{\fboxsep}{0pt}\setlength{\fboxrule}{0.2pt}\fcolorbox{red}{white}{\adjincludegraphics[width=0.04\textwidth, trim={{0.15\width} {0.3\height} {0.15\width} {0.3\height}}, clip]{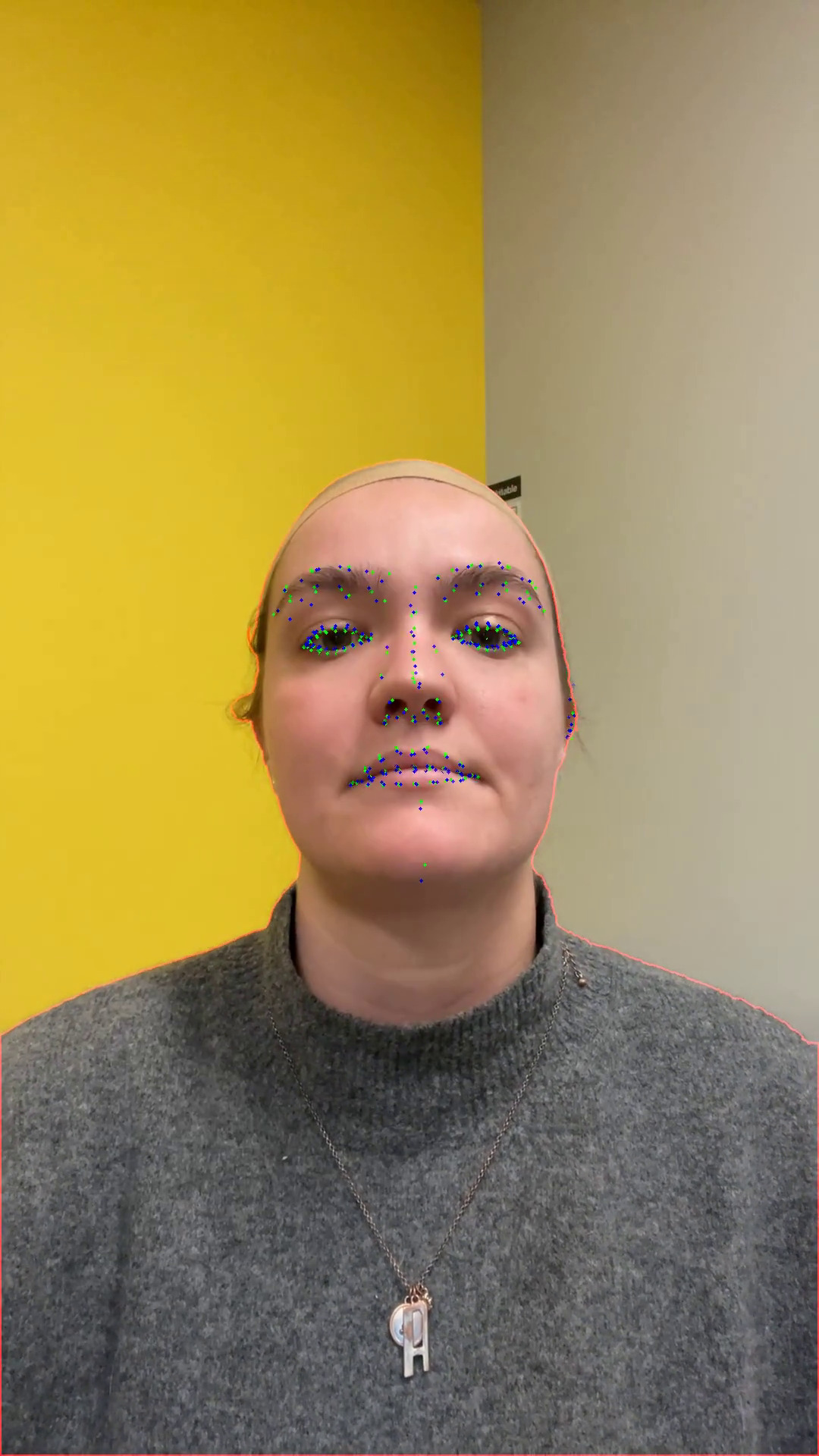}}};
 \end{tikzpicture}
 \\
 \begin{tikzpicture}
    \node[anchor=south west,inner sep=0] (image) at (0,0) {\adjincludegraphics[width=0.16\textwidth, trim={0 {0.2\height} 0 {0.25\height}}, clip]{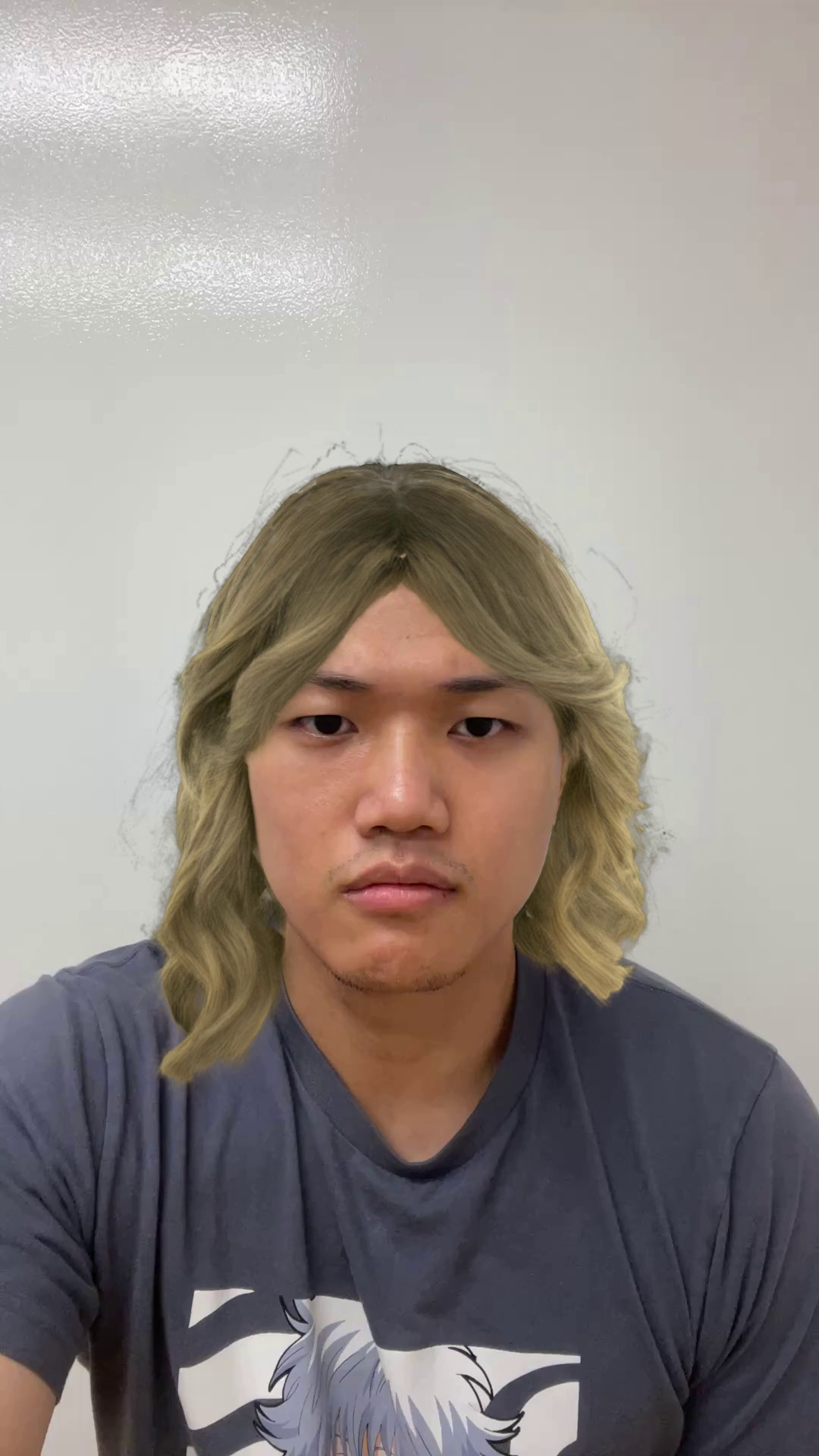}};
    \node[anchor=north west,inner sep=0] at (current bounding box.north west) {\setlength{\fboxsep}{0pt}\setlength{\fboxrule}{0.2pt}\fcolorbox{red}{white}{\adjincludegraphics[width=0.04\textwidth, trim={{0.2\width} {0.3\height} {0.2\width} {0.35\height}}, clip]{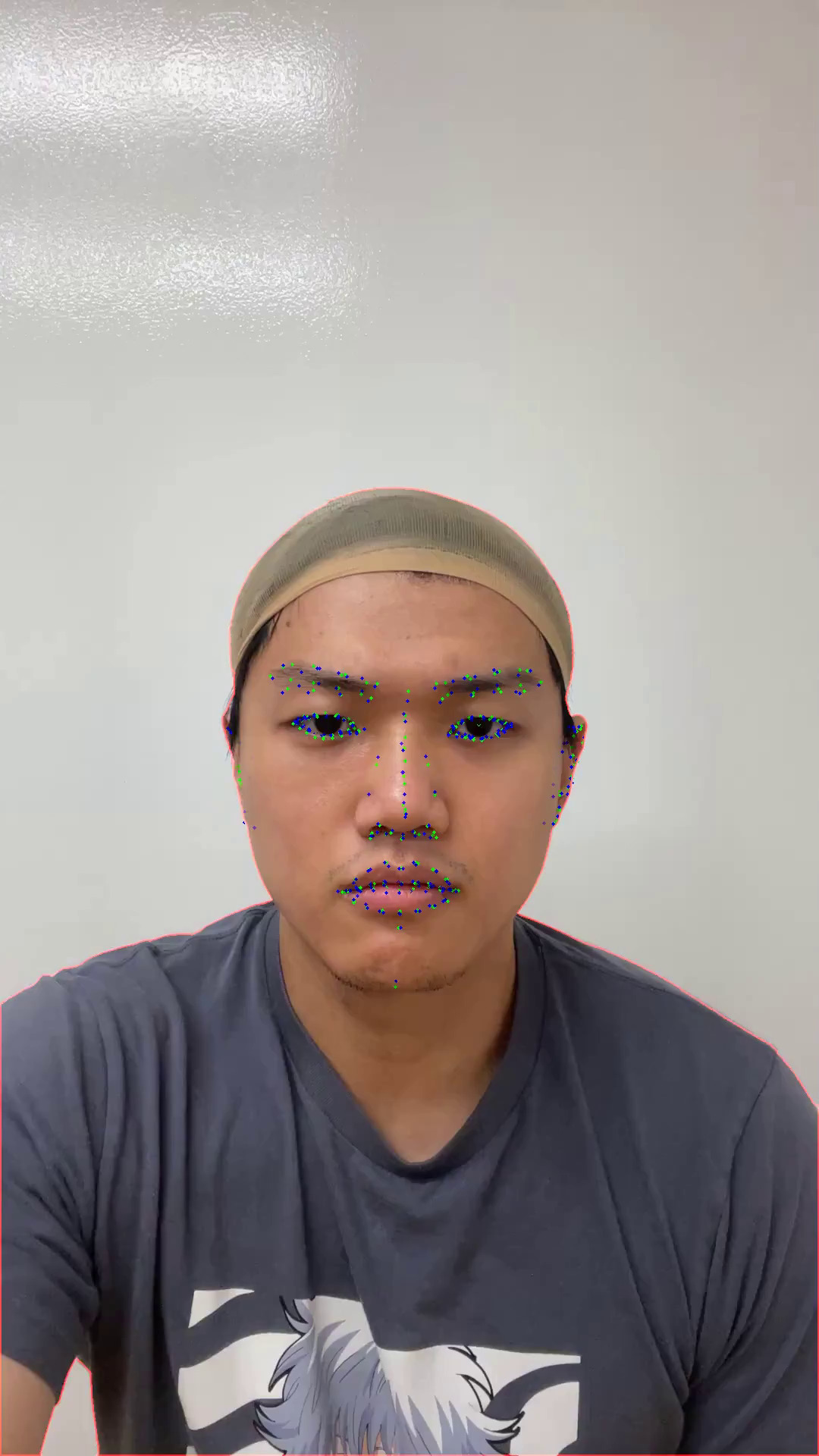}}};
 \end{tikzpicture} &
 \begin{tikzpicture}
    \node[anchor=south west,inner sep=0] (image) at (0,0) {\adjincludegraphics[width=0.16\textwidth, trim={0 {0.2\height} 0 {0.25\height}}, clip]{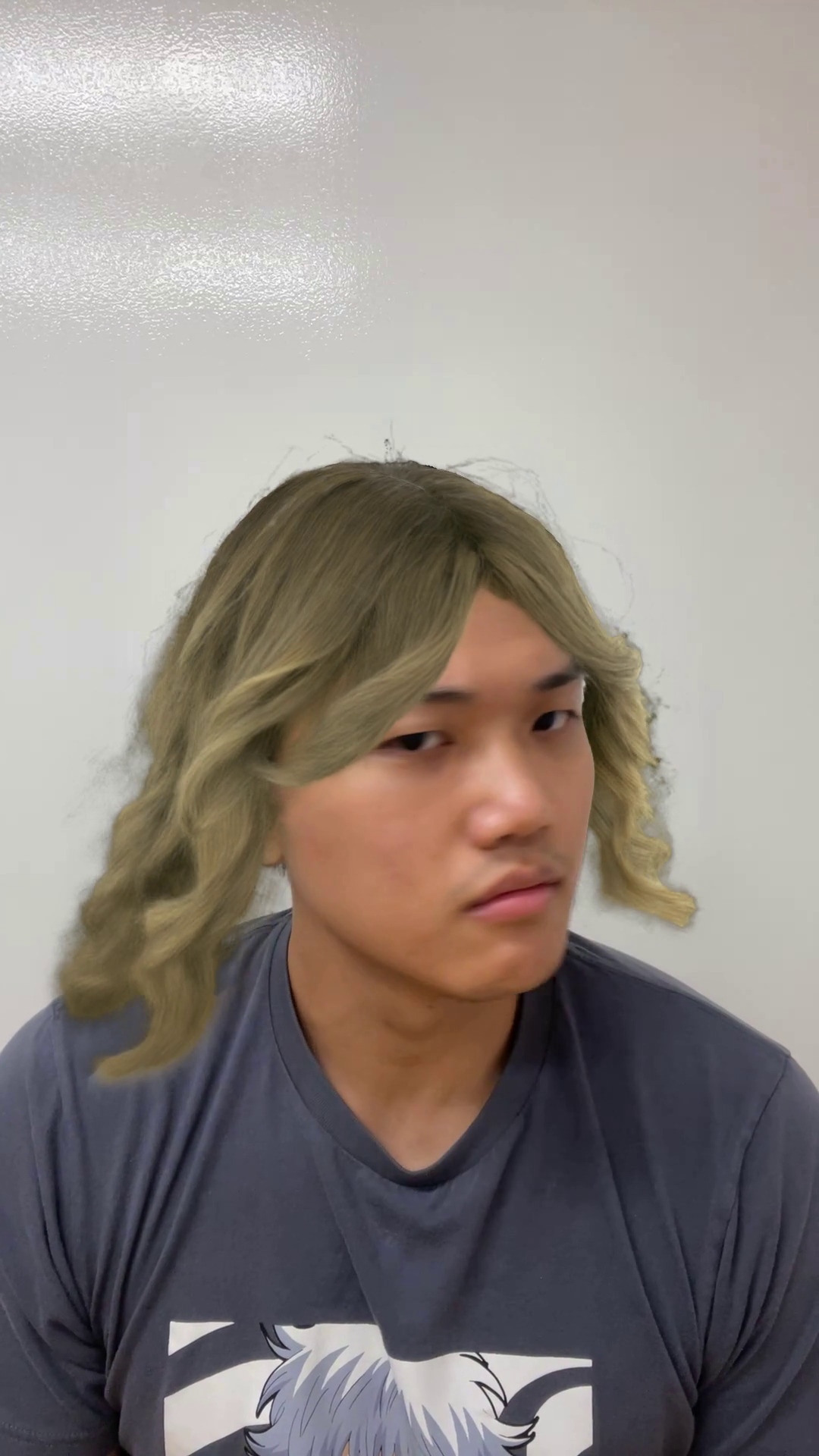}};
    \node[anchor=north west,inner sep=0] at (current bounding box.north west) {\setlength{\fboxsep}{0pt}\setlength{\fboxrule}{0.2pt}\fcolorbox{red}{white}{\adjincludegraphics[width=0.04\textwidth, trim={{0.2\width} {0.3\height} {0.2\width} {0.35\height}}, clip]{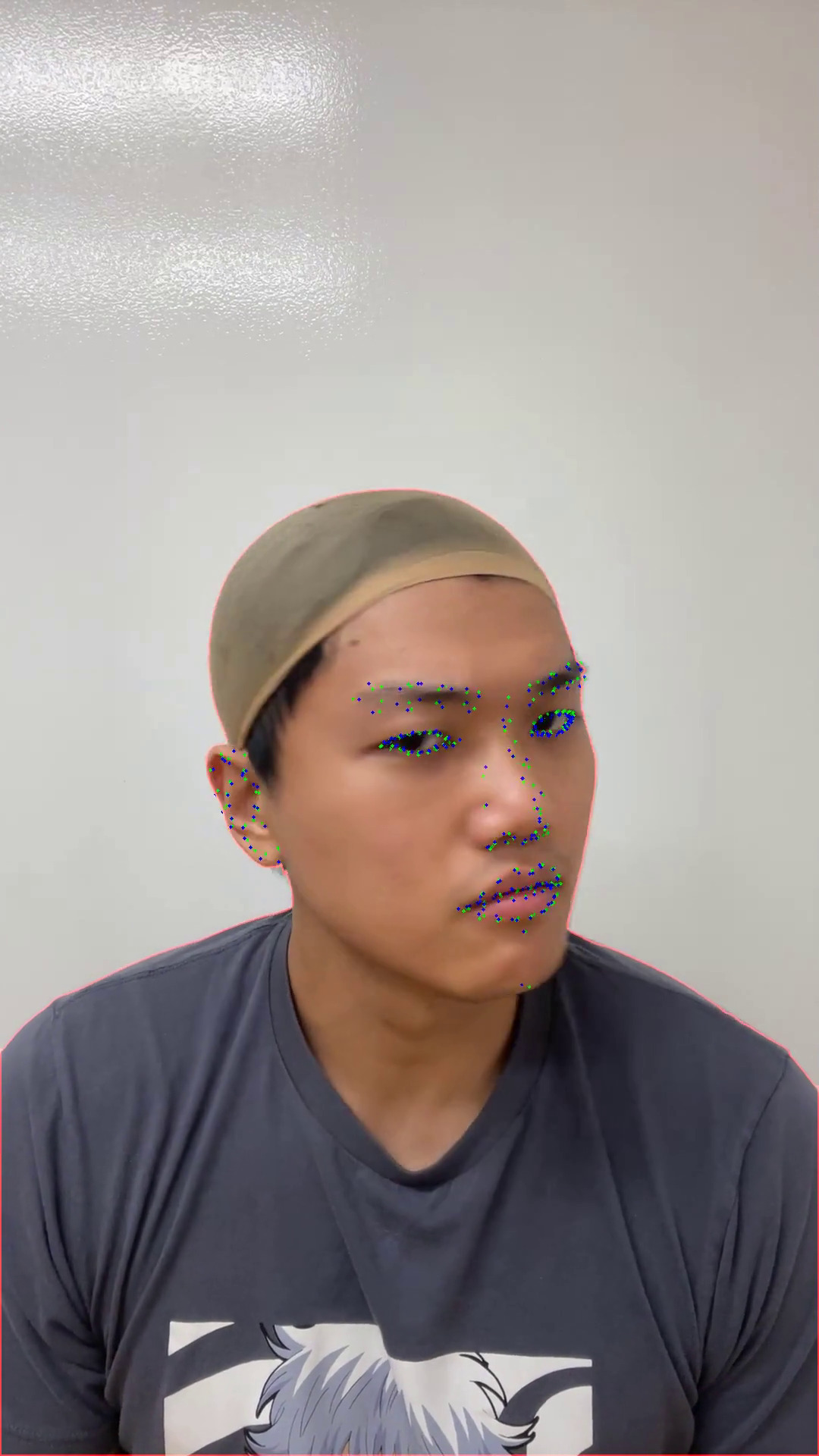}}};
 \end{tikzpicture} &
 \begin{tikzpicture}
    \node[anchor=south west,inner sep=0] (image) at (0,0) {\adjincludegraphics[width=0.16\textwidth, trim={0 {0.2\height} 0 {0.25\height}}, clip]{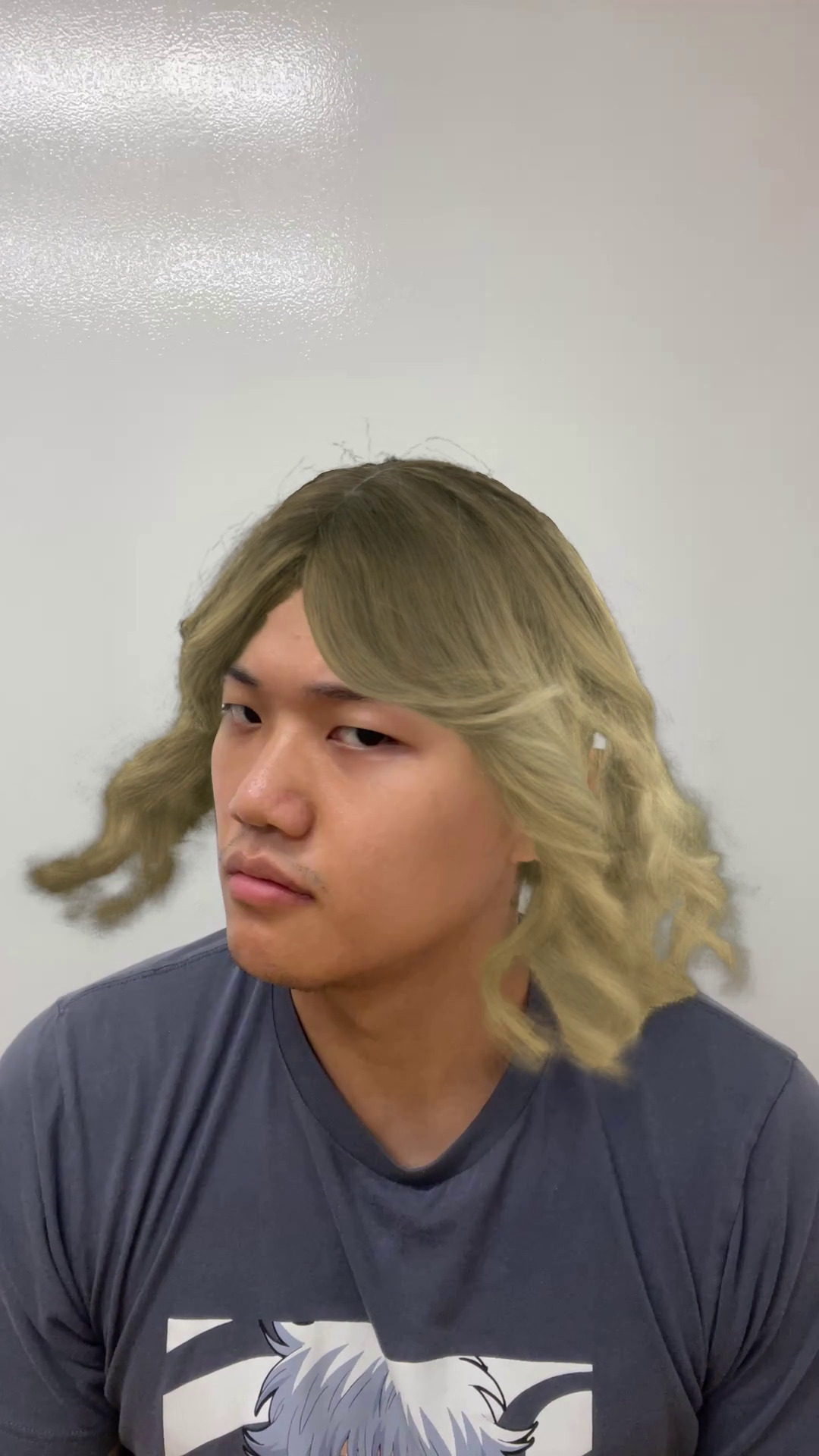}};
    \node[anchor=north west,inner sep=0] at (current bounding box.north west) {\setlength{\fboxsep}{0pt}\setlength{\fboxrule}{0.2pt}\fcolorbox{red}{white}{\adjincludegraphics[width=0.04\textwidth, trim={{0.2\width} {0.3\height} {0.2\width} {0.35\height}}, clip]{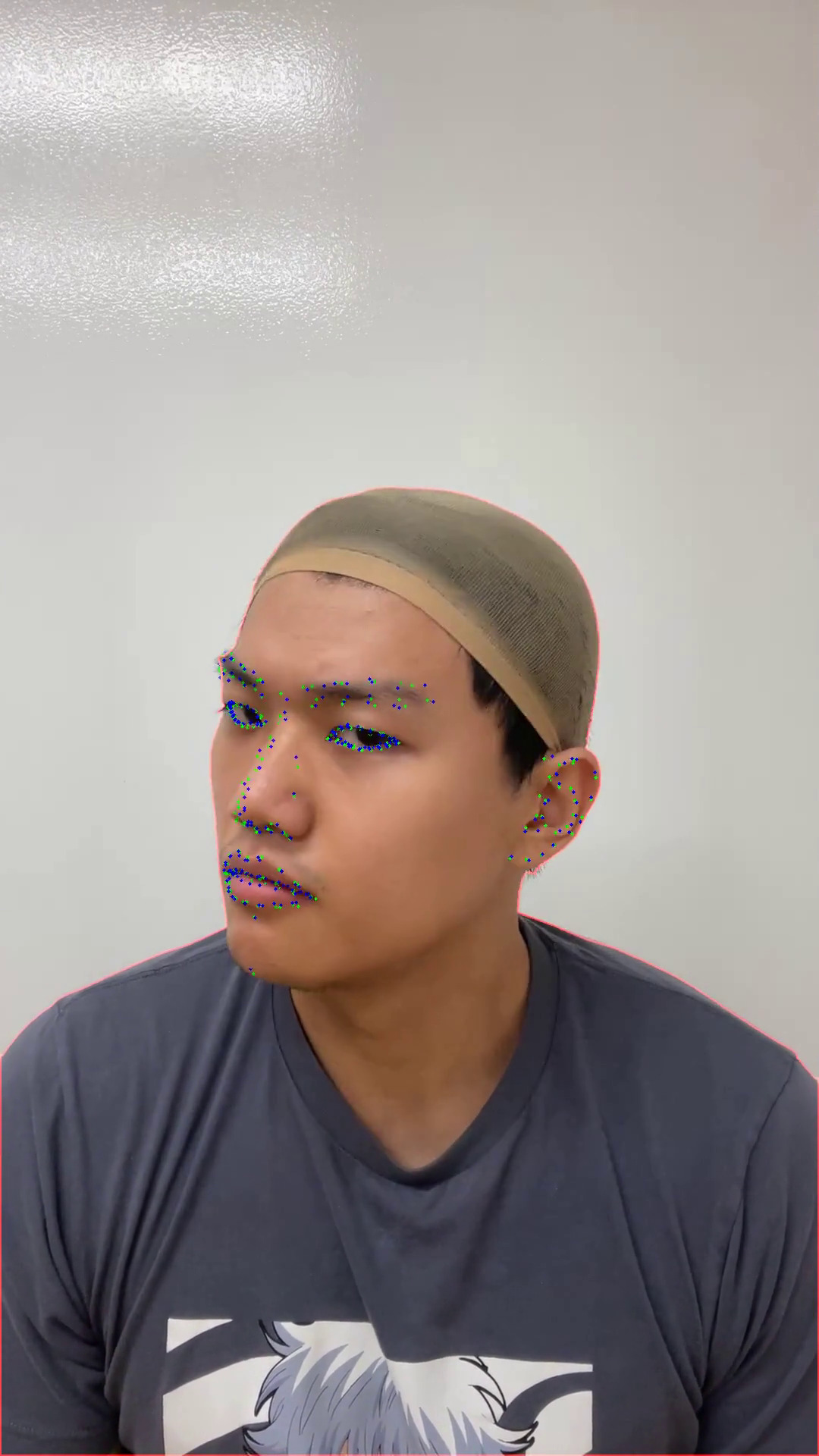}}};
 \end{tikzpicture} \\
 \begin{tikzpicture}
    \node[anchor=south west,inner sep=0] (image) at (0,0) {\adjincludegraphics[width=0.16\textwidth, trim={0 {0.25\height} 0 {0.2\height}}, clip]{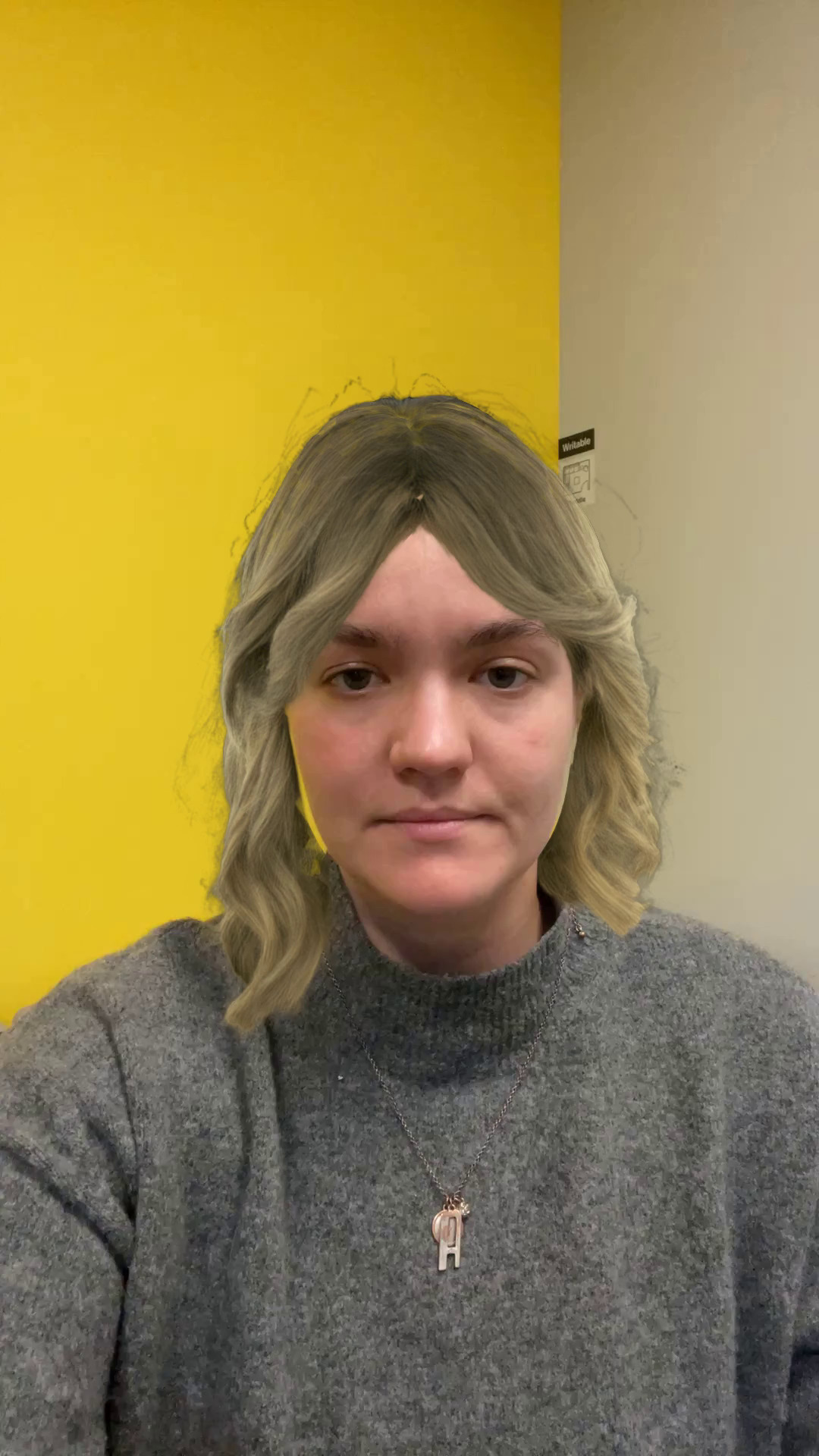}};
    \node[anchor=north west,inner sep=0] at (current bounding box.north west) {\setlength{\fboxsep}{0pt}\setlength{\fboxrule}{0.2pt}\fcolorbox{red}{white}{\adjincludegraphics[width=0.04\textwidth, trim={{0.2\width} {0.35\height} {0.2\width} {0.3\height}}, clip]{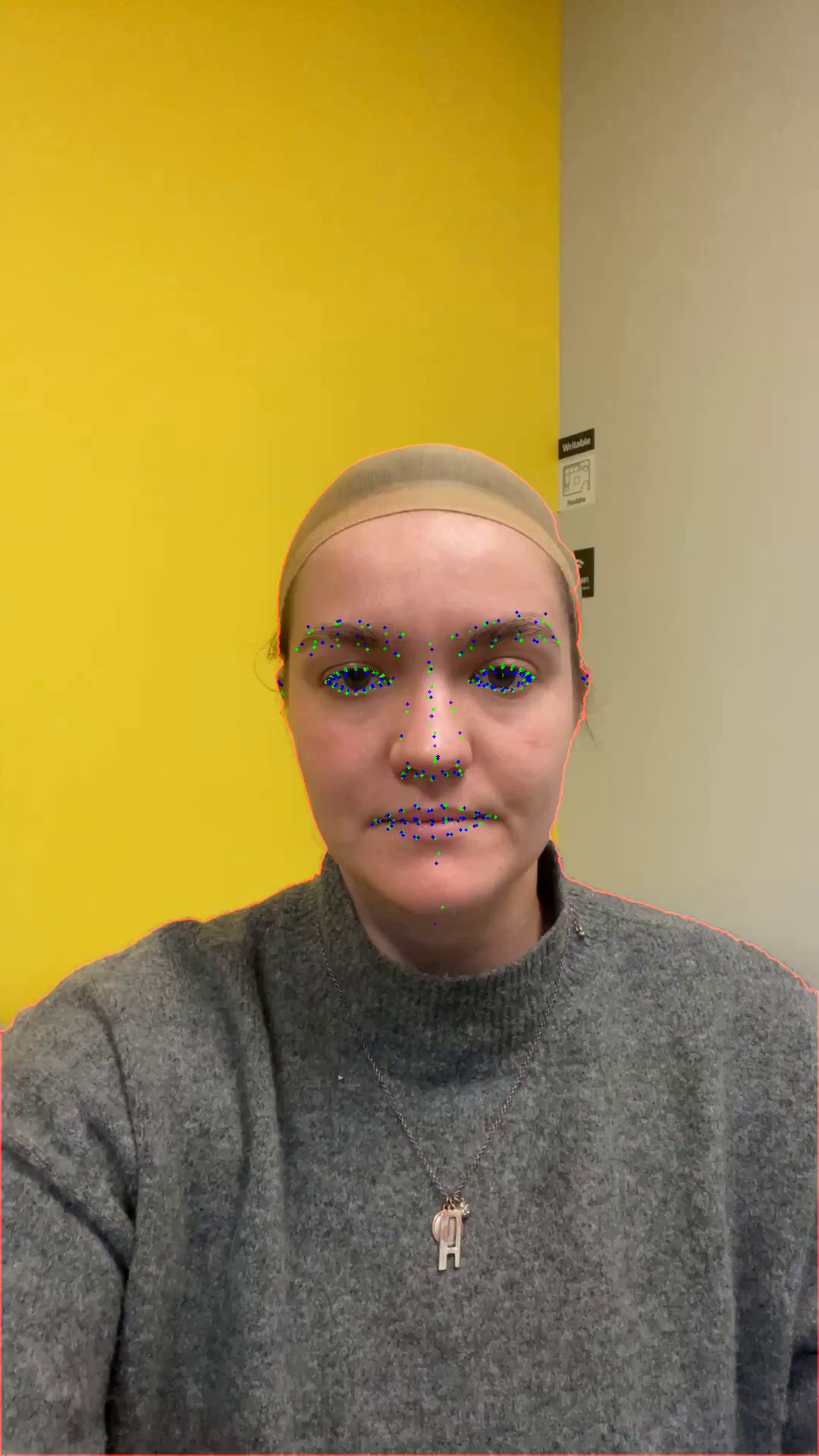}}};
 \end{tikzpicture} &
 \begin{tikzpicture}
    \node[anchor=south west,inner sep=0] (image) at (0,0) {\adjincludegraphics[width=0.16\textwidth, trim={0 {0.25\height} 0 {0.2\height}}, clip]{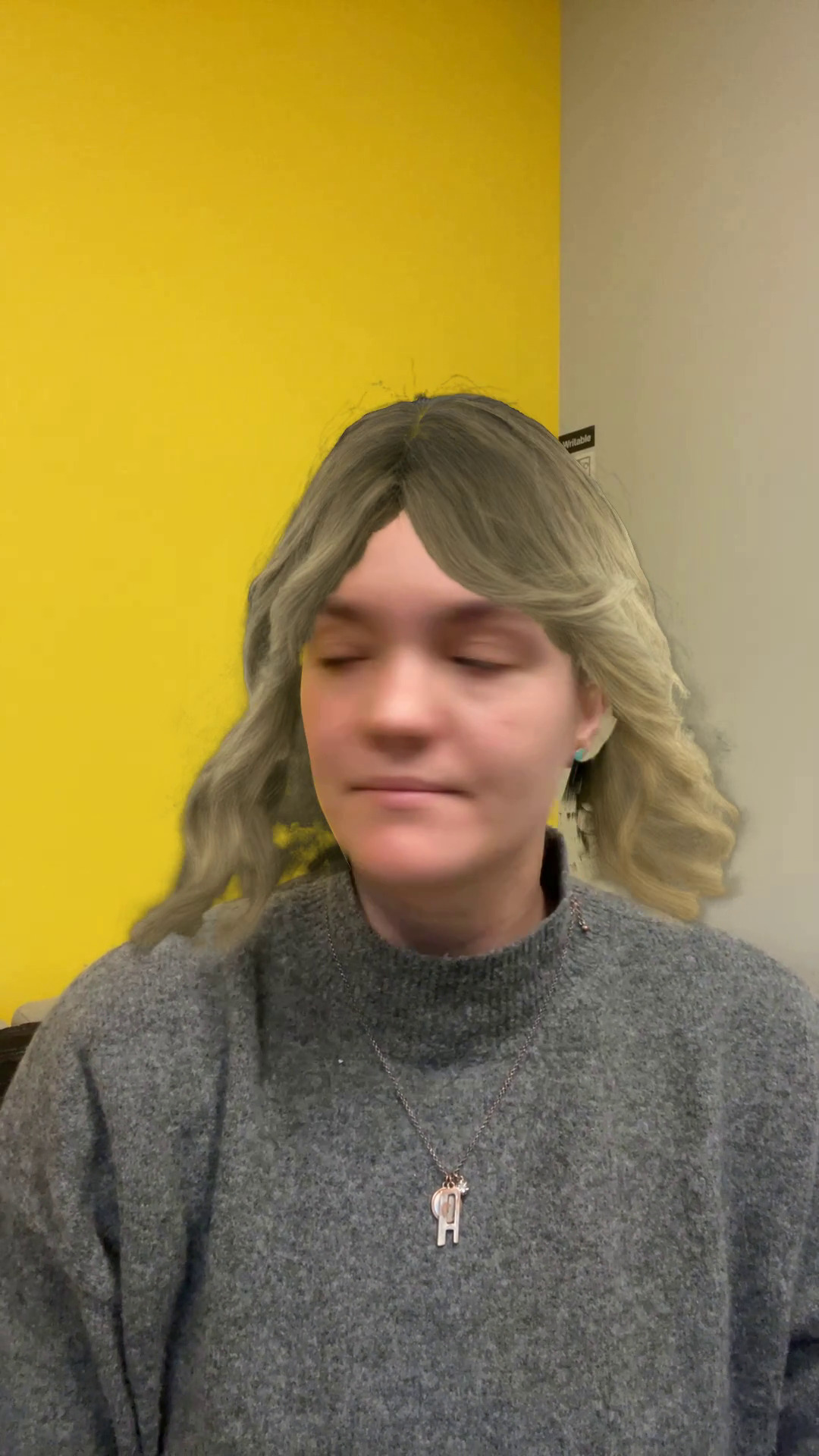}};
    \node[anchor=north west,inner sep=0] at (current bounding box.north west) {\setlength{\fboxsep}{0pt}\setlength{\fboxrule}{0.2pt}\fcolorbox{red}{white}{\adjincludegraphics[width=0.04\textwidth, trim={{0.2\width} {0.35\height} {0.2\width} {0.3\height}}, clip]{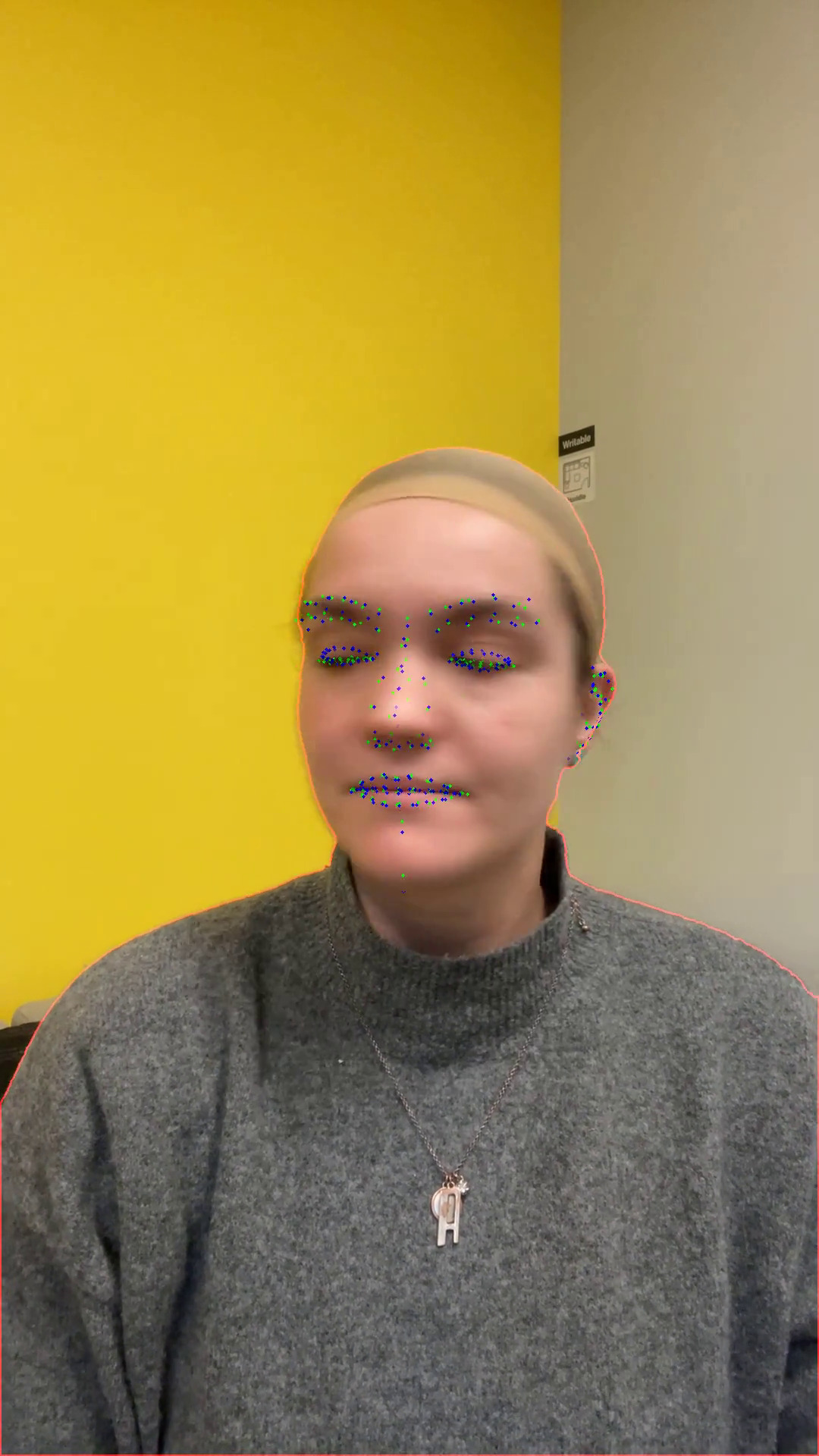}}};
 \end{tikzpicture} &
\begin{tikzpicture}
    \node[anchor=south west,inner sep=0] (image) at (0,0) {\adjincludegraphics[width=0.16\textwidth, trim={0 {0.25\height} 0 {0.2\height}}, clip]{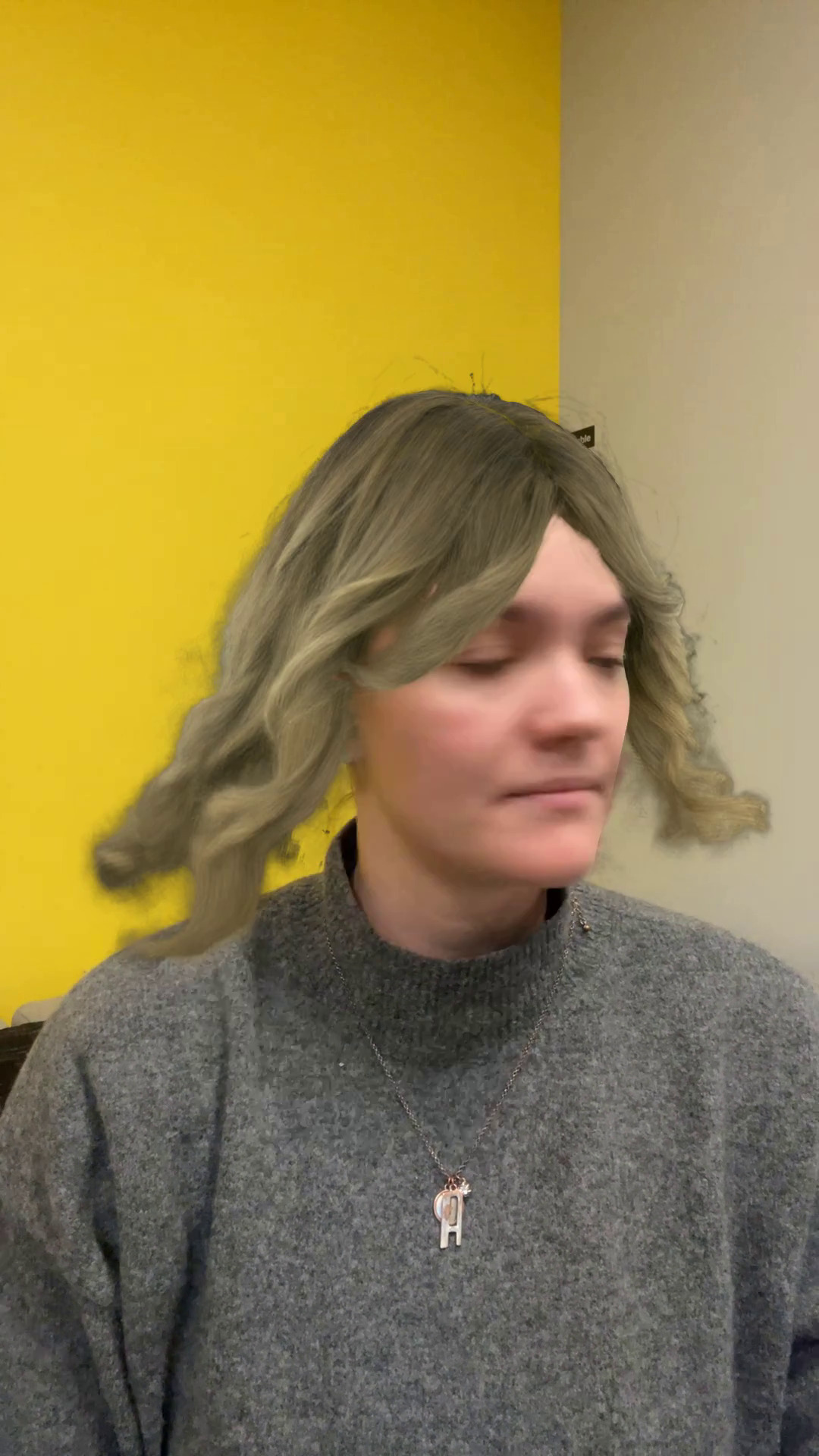}};
    \node[anchor=north west,inner sep=0] at (current bounding box.north west) {\setlength{\fboxsep}{0pt}\setlength{\fboxrule}{0.2pt}\fcolorbox{red}{white}{\adjincludegraphics[width=0.04\textwidth, trim={{0.2\width} {0.35\height} {0.2\width} {0.3\height}}, clip]{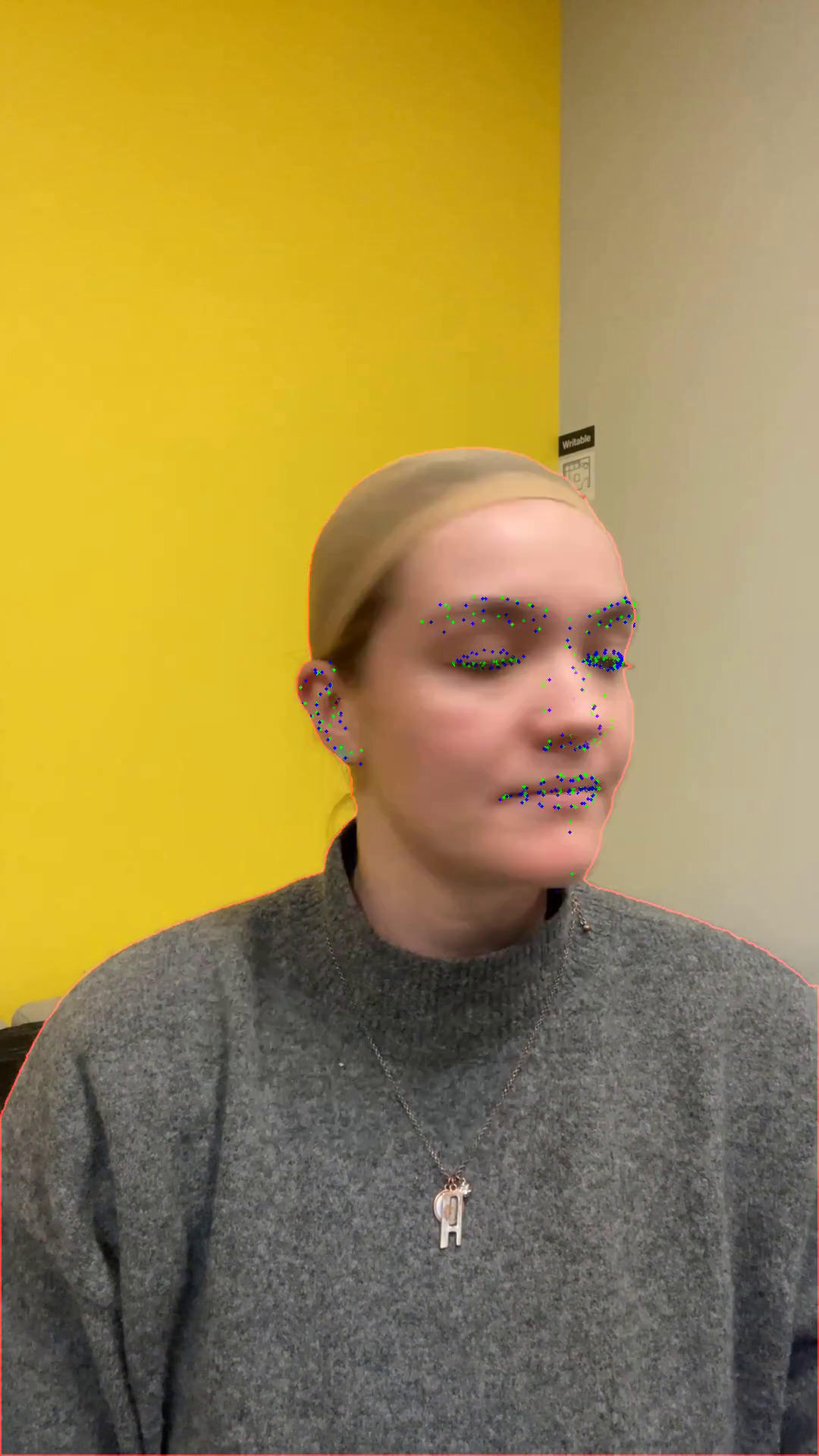}}};
 \end{tikzpicture}
\end{tabular}
\caption{\label{fig0:anim_iphone}\textbf{Animation from Single View Captures.} Our model can generate realistic hair animation from single view video based on head motion and gravity direction. Original captures of subjects wearing a wig cap are shown in red boxes.}
\end{figure}

\input{intro.tex}
\input{related_work.tex}
\input{method.tex}
\input{experiment.tex}

\input{discuss.tex}
\input{appendix}


{\small
\bibliographystyle{ieee_fullname}
\bibliography{main}
}

\end{document}

%% file: defs.tex
 \newcommand{\ZW}[1]{{\color{blue}{\bf ZW: #1}}}
 
 \newcommand{\SL}[1]{{\color{olive}{\bf SL: #1}}}
 \newcommand{\GN}[1]{{\color{orange}{\bf GN: #1}}}
 
 \definecolor{tscolor}{RGB}{153,51,255}
 \newcommand{\TS}[1]{{\color{tscolor}{\bf TS: #1}}}

 \definecolor{mzcolor}{RGB}{255,50,00}
 \newcommand\MZ[1] {\textbf{\textcolor{mzcolor}{MZ: #1}}}

 \newcommand{\CR}[1]{#1}
 
 \newcommand{\CL}[1]{{\color{red}{\bf{CL: #1}}}}
 
 \newcommand{\JKH}[1]{{\color{blue}{\bf JKH: #1}}}
 
 \definecolor{cccolor}{RGB}{0,50,255}
 \newcommand\CC[1] {\textbf{\textcolor{cccolor}{CC: #1}}}

 \newcommand{\bp}{\mathbf{p}}
 \newcommand{\bc}{\mathbf{c}}
 \newcommand{\bft}{\mathbf{f}}
 \newcommand{\bz}{\mathbf{z}}
 \newcommand{\bv}{\mathbf{v}}
 \newcommand{\br}{\mathbf{r}}
 \newcommand{\bV}{\mathbf{V}}
 \newcommand{\bR}{\mathbb{R}}

 \newcommand{\mL}{\mathcal{L}}
 \newcommand{\mR}{\mathcal{R}}
 \newcommand{\plh}{\mkern-1.5mu{\times}\mkern-2mu}
 
 \newcommand{\wt}[1]{\widetilde{#1}}
 
 \newcommand{\ttt}[1]{\small\texttt{#1}}

 \newcommand{\IGNORE}[1]{}

%% file: intro.tex
\section{Introduction}

%
The ability to model the details of human hair with high fidelity is key to achieving realism in human avatar creation because hair establishes part of our personal identity.
%
The realism of human hair involves not only geometry, appearance and interaction with light, but also motion.
The sheer number of hair strands leads to a very complex geometry, while the interactions between light and hair strands lead to non-trivial view-dependent appearance changes.
Capturing the dynamics of hair is difficult due to the complexity of the motion space as well as severe self-occlusions.
Generating realistic dynamics given only the head motion is similarly hard because the motion of hair is not solely controlled by the head position but also influenced by gravity and inertial forces. 
From a control perspective, hair does not respond linearly to the head position and a zero-order system is inadequate for modeling hair dynamics.

%
These challenges lead to two major problems for creating realistic hair avatars: 3D hair capture and dynamics modeling.
%
%
While modern capture systems reconstruct the hair geometry and appearance from a sparse and discrete set of real world observations with high fidelity, they do not directly solve the problem of novel motion generation.
%
To achieve that, we need to go beyond reconstructing to create a \emph{controllable} dynamic hair model using captured data.
%

%
In conventional animation techniques, hair geometry is created by an artist manually preparing 3D hair grooms. Motion of the 3D hair groom is created by a physics simulator where an artist selects the parameters for the simulation.
This process requires expert knowledge. 
%
%
In contrast, data-driven methods aim to achieve hair capture and animation in an automatic way while preserving metric photo-realism.
Most of the current data-driven hair capture and animation approaches learn to regress a dense 3D hair representation that is renderable directly from per-frame driving signals, without modeling dynamics.

However, there are several factors that limit the practical use of these data-driven methods for hair animation.
First of all, these methods mostly rely on sophisticated driving signals, like multi-view images~\cite{nvs_steve, park2021nerfies}, a tracked mesh of the hair~\cite{steve_mvp}, or tracked guide hair strands~\cite{wang2021hvh}, which are hard to acquire.
Furthermore, from an animation perspective, these models are limited to rendering hair based on hair observations and cannot be used to generate novel motion of hair .
%
Sometimes it is not possible to record the hair driving signals at all.
We might want to animate hair for a person wearing accessories or equipment that (partially) obstructs the view of their hair, for example VR glasses; or animate a novel hair style for a subject or animate hair for a bald person.
%

To address these limitations of existing data-driven hair capture and animation approaches, we present a neural dynamic model that is able to animate hair with high fidelity conditioned on head motion and relative gravity direction.
By building such a dynamic model, we are able to generate hair motions by evolving an initial hair state into a future one, without relying on per-frame hair observation as a driving signal.
We utilize a two-stage approach for creating this dynamic model: in the first stage, we perform state compression by learning a hair autoencoder from multi-view video captures with an evolving tracking algorithm.
Our method is capable of capturing a temporally consistent, fully renderable volumetric representation of hair from videos with both head and hair.
Hair states with different time-stamps are parameterized into a semantic embedding space via the autoencoder.
%
%
In the second stage, we sample temporally adjacent pairs from the semantic embedding space and learn a dynamic model that can perform the hair state transition between each state in the embedding space given the previous head motion and gravity direction.
With such a dynamic model, we can perform hair state evolution and hair animation in a recurrent manner which is not driven by existing hair observations. 
As shown in Fig~\ref{fig0:anim_iphone}, our method is capable of generating realistic hair animation with different hair styles on single view captures of a moving head with a bald cap.
%
%
In summary, the contributions of this work are
\begin{itemize}[leftmargin=*, noitemsep]
    \item We present NeuWigs, a novel end-to-end data-driven pipeline with a volumetric autoencoder as the backbone for real human hair capture and animation, learnt from multi-view RGB images.
    \item We learn the hair geometry, tracking and appearance end-to-end with a novel autoencoder-as-a-tracker strategy for hair state compression, where the hair is modeled separately from the head using multi-view hair segmentation.
    \item We train an animatable hair dynamic model that is robust to drift using a hair state denoiser realized by the 3D autoencoder from the compression stage.
\end{itemize}

%% file: related_work.tex
\section{Related Work}

We discuss related work in hair capture, data-driven hair animation, and volumetric avatars.

\noindent\textbf{Static Hair Capture.} Reconstructing static hair is challenging due to its complex geometry. Paris \textit{et al.}~\cite{paris2004capture} and Wei \textit{et al.}~\cite{wei2005modeling} reconstruct 3D hair from multi-view images and 2D orientation maps. PhotoBooth~\cite{paris2008hair_photobooth} further improves this technique by applying calibrated projectors and cameras with patterned lights to reconstruct finer geometry. Luo \textit{et al.}~\cite{luo2012multi} and Hu \textit{et al.}~
\cite{hu2014robust} leverage hair specific structure (strand geometry) and physically simulated strands for more robust reconstruction. The state-of-the-art work on static hair capture is Nam \textit{et al.}~\cite{nam2019lmvs}. They present a line-based patch method for reconstructing 3D line clouds. Sun \textit{et al.}~\cite{sun2021hairinverse} extends this approach to reconstruct both geometry and appearance with OLAT images. Rosu \textit{et al.}~\cite{rosu2022neuralstrands} presents a learning framework for hair reconstruction that utilizes a hair strand prior trained on synthetic data.

\noindent\textbf{Dynamic Hair Capture.} Hair dynamics is hard to capture as a result of the complex motion patterns and self-occlusions. Zhang \textit{et al.}~\cite{zhang2012simulation} refine hair dynamics by applying physical simulation techniques to per-frame hair reconstruction. Hu \textit{et al.}~\cite{hu2017simulation} invert the problem of hair tracking by directly solving for hair dynamic parameters through iterative grid search on thousands of simulation results under different settings. Xu \textit{et al.}~\cite{xu2014dynamic} perform hair strand tracking by extracting hair strand tracklets from spatio-temporal slices of a video volume. Liang \textit{et al.}\cite{liang2018video} and Yang \textit{et al.}~\cite{yang2019dynamic} design a learning framework fueled by synthetic hair data to regress 3D hair from video. Winberg \textit{et al.}~\cite{winberg2022facial} perform dense facial hair and underlying skin tracking under quasi-rigid motion with a multi-camera system. While achieving good results on the capture of the hair geometry from dynamic sequences, those methods either do not model hair appearance with photo-realism or do not solve the problem of drivable animation.

\noindent\textbf{Data-driven Hair Animation.} Using physics-based simulation for hair animation is a common practice in both, academia and the film/games industry~\cite{ward2007survey, bertails2008realistic}. However, generating hair animations with physics-based simulation can be computationally costly. To remedy this problem, reduced data-driven methods~\cite{chai2014reduced, chai2016adaptive, guan2012multi} simulate only a small portion of guide hair strands and interpolate the rest using skinning weights learned from full simulations. With the latest advances in deep learning, the efficiency of both dynamic generation~\cite{lyu2020real} and rendering~\cite{olszewski2020intuitive, chai2020neural} of hair has been improved using neural networks. Lyu \textit{et al.}~\cite{lyu2020real} uses deep neural networks for adaptive binding between normal hair and guide hair. Olszewski \textit{et al.}~\cite{olszewski2020intuitive} treats hair rendering as an image translation problem and generates realistic rendering of hair conditioned on 2D hair masks and strokes. Similarly, Chai \textit{et al.}~\cite{chai2020neural} achieves faster rendering with photorealistic results by substituting the rendering part in the animation pipeline with screen-space neural rendering techniques. Temporal consistency is enforced in this pipeline by conditioning on hair flow.
However, those methods still build on top of conventional hair simulation pipelines and use synthetic hair wigs, which require manual efforts by an artist to set up and are non-trivial to metrically evaluate. 
Wu \textit{et al.}~\cite{wu2016data} propose to use a secondary motion graph (SDG) for hair animation without relying on a conventional hair simulation pipeline at runtime.
However, this method is limited by artist's design of hair wigs and control of hair simulation parameters and is not able to capture or animate hair with realistic motion.  

\noindent\textbf{Volumetric Avatars.} With the recent advent of differentiable volumetric raymarching~\cite{nvs_steve, mildenhall2020nerf}, many works attempt to  directly build volumetric avatars from images or videos. %
To the best of our knowledge, Neural Volumes~\cite{steve_nvs} is the earliest work that creates a volumetric head avatar from multiview images using differentiable volumetric raymarching.
One of the many strengths of this work is that it directly optimizes a volume grid from multiview images while still producing high quality renders for semi-transparent objects like hair.
One followup work~\cite{wang2021learning} combines volumetric and coordinate-based representations into a hybrid form for better rendering quality and drivability.
However, the level of detail either methods can capture is limited by the resolution of the volume grid.

Another early work on differentiable volumetric rendering is NeRF~\cite{mildenhall2020nerf}, which parameterizes radiance fields implicitly with MLPs instead of using a volumetric grid. 
Due to the success of NeRF~\cite{mildenhall2020nerf} for modeling 3D scenes from multiple images, there are many works that build avatars with NeRF~\cite{Raj_2021_CVPR, park2021nerfies, Gafni_2021_CVPR, Zheng_2022_CVPR, Hong_2022_CVPR, Grassal_2022_CVPR, Kornilova_2022_CVPR, kania2022conerf, keypoint_nerf, park2021hypernerf,li2022tava}. 
%
%
%
PVA~\cite{Raj_2021_CVPR} and KeypointNeRF~\cite{keypoint_nerf} utilize pixel aligned information to extend NeRF's drivability and generalization over sequence data.
%
%
%
%
Nerfies~\cite{park2021nerfies}, HyperNeRF~\cite{park2021hypernerf} and TAVA~\cite{li2022tava} optimize a deformation field together with a NeRF in a canonical space given videos.
NeRFace~\cite{Gafni_2021_CVPR}, IM Avatar~\cite{Zheng_2022_CVPR} and HeadNeRF~\cite{Hong_2022_CVPR} substitute the deformation field with face models like 3DMM~\cite{blanz1999morphable} or FLAME~\cite{li2017flame} for better controllablity.
%
%
%
%
However, those methods mostly assume hair to be rigidly attached to the head without motion and most of the approaches suffer from prohibitively long rendering time. 

In contrast to NeRF-based avatars, a mixture of volumetric primitives (MVP)~\cite{steve_mvp} builds a volumetric representation that can generate high-quality \emph{real time} renderings that look realistic even for challenging materials like hair and clothing. 
Several follow up works extend it to model body dynamics~\cite{remelli2022drivable}, moderate hair dynamics~\cite{wang2021hvh} and even for in-the-wild captures~\cite{ica_chen}.
However, animating such volumetric representations with dynamics is still an unsolved problem. 
%

%% file: method.tex
\section{Method}
\begin{figure*}[htb]
    \centering
    \adjincludegraphics[width=\textwidth, trim={{0.005\width} {0.03\height} {0.004\width} 0}, clip]{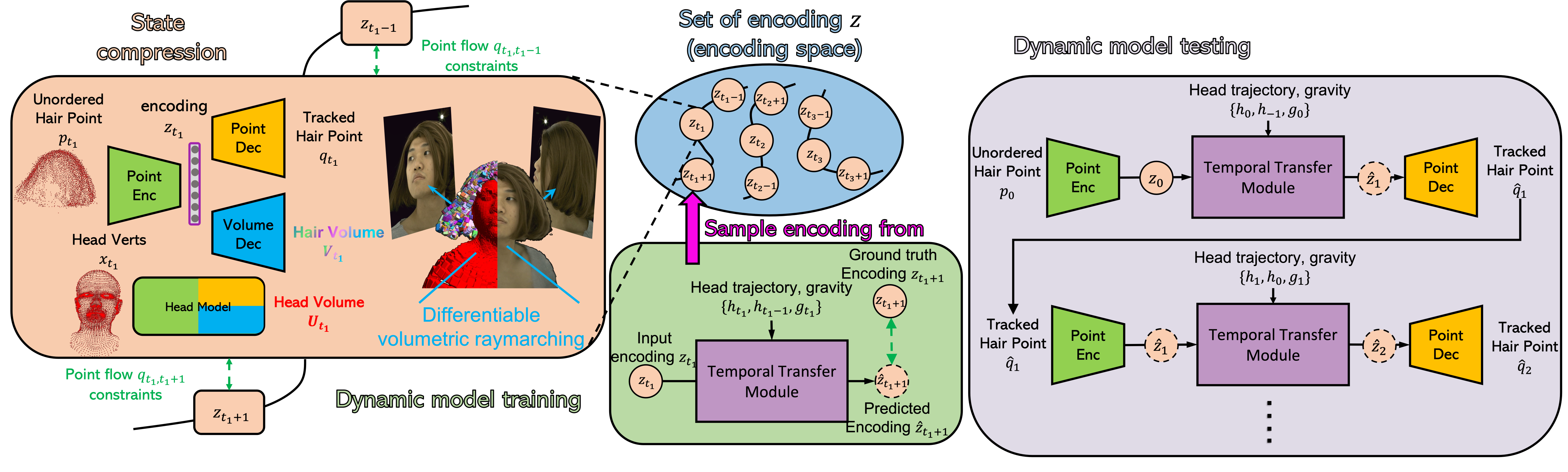}
    \caption{\label{fig3:pipeline}\textbf{Method Overview.} Our method is comprised of two stages: state compression and dynamic modeling. In the first stage, we train an autoencoder for hair and head appearance from multiview RGB images using differentiable volumetric raymarching; at the same time we create an encoding space of hair states. In the dynamic modeling stage, we sample temporally adjacent hair encodings to train a temporal transfer module ($\mathtt{T^2M}$) that performs the transfer between the two, based on head motion and head-relative gravity direction. \iffalse\CL{Please add edit link to latex source.}\fi \iffalse\href{https://fb-my.sharepoint.com/:p:/r/personal/ziyanw1_meta_com/Documents/pipeline_canva_cvpr2023_online.pptx?d=w0aaf8c0707044ee9a5aaf5cb25c3125a&csf=1&web=1&e=RNdre3}{link to figure}\fi}
\end{figure*}
Our method for hair performance capture and animation consists of two stages: state compression and dynamic modeling (see also Fig.~\ref{fig3:pipeline}).
The goal of the first stage is to perform dynamic hair capture from multi-view video of a head in motion. 
To be more specific, in this stage, we aim to distill a 3D renderable representation of hair from multi-view images at each frame into an embedding space.
To achieve that, we train a volumetric autoencoder in a self-supervised manner to model the hair geometry, tracking and appearance. 
The output of this model is a set of tracked hair point clouds $p_t$, their corresponding local radiance fields in the form of volumetric primitives $\bm{V}_{t}$ and a compact embedding space that is spanned by the 1D hair state encoding $\bm{z}_t$.
%
%
%
In the second stage, we perform modeling of hair dynamics based on the hair capture from the first stage.
The goal of this stage is to create a controllable, self-evolving representation of hair without relying on online observations of hair.
We achieve this by learning a neural network to regress the next possible hair state, which is conditioned on the previous hair state as well as the previous head motion and head-relative gravity direction.
Equipped with the hair encoding space acquired from the first stage, we can train that model in a supervised manner by simply sampling data pairs of temporally adjacent hair states from the encoding space.
Using both stages, we can perform dynamic hair animation at test time given an initialization using a recurrent strategy, without relying on direct observations of hair as a per-frame driving signal.
\subsection{State Compression}
We assume that multi-view image captures ${I_{cam_i}}$ with their corresponding calibrated cameras are given.
We denote the extrinsics of each camera $i$ as $\mathbf R_i$ and $\mathbf t_i$.
We then run l-MVS~\cite{nam2019lmvs} and non-rigid tracking~\cite{wu2018deep} to obtain the per-frame hair reconstructions $p_t\in \mathbb R^{N_{p_t}\times 3}$ and head tracked vertices $x_t\in \mathbb R^{N_{x_t}\times 3}$ at time frame $t$, where $N_{p_t}$ and $N_{x_t}$ denote the size of each.
The $p_t$ and $x_t$ together serves as a coarse representation for the hair and head.
Different from the head vertices, here the $p_t$ represent an unordered set of hair point clouds.
Due to the difference between hair and head dynamic patterns, we model them separately by training two different volumetric autoencoders. 
For the head model, we use an autoencoder to regress the volumetric texture in an unwrapped UV layout conditioned on the tracked head mesh, similar to~\cite{steve_mvp, wang2021hvh}.
For the hair model, we optimize the hair volumetric texture and its tracking simultaneously. 
To better enforce the disentanglement between hair and head, we attach head volumes only to the head mesh and hair volumes only to the hair point clouds.
Moreover, we use a segmentation loss to constrain each of them to only model the texture of their assigned category of hair or head.

\noindent\textbf{Autoencoder as a Tracker.}
Learning to track hair in a supervised manner with manual annotation is infeasible. 
%
%
We automatically discover the hair keypoints as well as their tracking information by optimizing a variational autoencoder (VAE)~\cite{kingma2013vae} in a semi-supervised manner.
By doing autoencoding on hair point clouds, we find that the VAE representing the hair point cloud can automatically align hair shapes along the temporal axis and is capable of tracking through both long and discontinuous hair video segments such as captures of different hair motions. 

The input to the point encoder $\mathcal{E}$ is the point coordinates of the unordered hair point cloud $p_t$.
Given its innate randomness in terms of point coverage and order, we use PointNet~\cite{qi2017pointnet} to extract the corresponding encoding $\bm{z_t} \in \mathbb R^{256}$.
Besides being agnostic to the order of $p_t$, it also can process varying numbers of points and aggregate global information from the input point cloud.
The point decoder $\mathcal{D}$ is a simple MLP that regresses the coordinate and point tangential direction of the tracked point cloud $q_t\in \mathbb R^{N_{prim}\times 3}$, $\bm{dir}(q_t)\in \mathbb R^{N_{prim}\times 3}$ and $s_t \in \mathbb R^{N_{prim}}$ from $\bm{z_t}$, where $N_{prim}$ is the number of tracked hair points.
We denote $\bm{dir}(x)$ as the tangential direction of $x$.
$s_t$ is a per-point scale factor which will be used later. We optimize the following loss to train the point autoencoder:
\begin{align*}
    \mathcal{L}_{geo} = \mathcal{L}_{cham} + \omega_{temp}\mathcal{L}_{temp} + \omega_{KL}\mathcal{L}_{KL}.
\end{align*}
The first term is the Chamfer distance loss which aims to align the shape of tracked point cloud $q_t$ to $p_t$:
\begin{align*}
    \mathcal{L}_{cham} &= ||q_t - \mathtt{N}_{q_t,p_t}||_2 - \boldsymbol{cos}(\boldsymbol{dir}(q_t), \boldsymbol{dir}(\mathtt{N}_{q_t,p_t}))\\
    +& ||p_t - \mathtt{N}_{q_t,p_t}||_2 - \boldsymbol{cos}(\boldsymbol{dir}(p_t), \boldsymbol{dir}(\mathtt{N}_{q_t, p_t})),
\end{align*}
%
\noindent where $\bm{cos}(\cdot,\cdot)$ is the cosine similarity and $\mathtt{N}_{x,y}\in \mathbb R^{N_x\times 3}$ are the coordinates of the nearest neighbor of each point of $x$ in $y$.
To further enforce temporal smoothness, we use point flow $\overrightarrow{fl}(p_t)$ and $\overleftarrow{fl}(p_t)$ denoting forward and backward flow from $p_t$ to $p_{t+1}$ and $p_{t-1}$ as additional supervision and formulate $\mathcal{L}_{temp}$ as follows:
\begin{align*}
    \mathcal{L}_{temp} &= ||\overleftarrow{fl}(\hat{p_t}) - \overleftarrow{fl}(\mathtt{N}_{q_t, p_t})||_2 + ||\overleftarrow{fl}(p_t) - \overleftarrow{fl}(\mathtt{N}_{p_t,q_t})||_2 \\
    +& ||\overrightarrow{fl}(\hat{p_t}) - \overrightarrow{fl}(\mathtt{N}_{q_t, p_t})||_2 + ||\overrightarrow{fl}(p_t) - \overrightarrow{fl}(\mathtt{N}_{p_t,q_t})||_2,
\end{align*}
\noindent where as $q_t$ is the tracked point, we can simply have $\overleftarrow{fl}(q_t)=q_t-q_{t-1}$ and $\overrightarrow{fl}(q_t)=q_t-q_{t+1}$.
Please see the supplemental materials for how we estimate $\overleftarrow{fl}(\mathtt{N}_{q_t, p_t})$. 
%
The last term $\mathcal{L}_{KL}$ is the KL-divergence loss~\cite{kingma2013vae} on the encoding $\bm{z_t}$ to enforce similarity with a normal distribution $\mathcal{N}(0, 1)$.

\noindent\textbf{Hair Volumetric Decoder.}
In parallel to the point decoder $\mathcal{D}$, we optimize a hair volumetric decoder that regresses a volumetric radiance field around each of the hair points.
The hair volumetric primitives $\bm{V}_{t}\in \mathbb R^{N_{prim}\times4\times m^3}$ store RGB and alpha in resolution of $m^3$.
We use a decoder similar to HVH~\cite{wang2021hvh} to regress the volume payload.
The pose of each volumetric primitive is directly determined by the output of the point decoder: $q_t$ and $\bm{dir}(q_t)$.
We denote $\bm{R}_t^{p,n}\in SO(3)$ and $\bm d_t^{p,n}\in \mathbb R^3$ as the $\bm{n}$th volume-to-world rotation and translation of hair volume and per-hair-volume scale $s_t^{p,n}$ as the $\bm{n}$th element in scale $\bm{s_t}$.
Similarly, $\bm d_t^n=q_t^n$ is the $\bm{n}$th element of $q_t$.
Given the head center $\bm{x}_t^c$ extracted from the head vertices $\bm{x}_t$ and hair head direction as $\bm{\bar{h}}_t^n=q_t^n-\bm{x}_t^c$, we formulate the rotation $\bm{R}_t^{p,n}$ as $\bm{R}_t^{p,n} = [\boldsymbol{l}(q_t^n), \boldsymbol{l}(q_t^n \times \bm{\bar{h}}_t^n), \boldsymbol{l}(q_t^n \times (q_t^n \times \bm{\bar{h}}_t^n))]^T$, where $\bm{l}(x) = x / ||x||_2$ is the normalization function.
The output of the head model is similar to the hair part except that it is modeling head related (non-hair) regions.
We denote the head volume payload as $\bm{U}_t\in \mathbb R^{N_{prim}\times3\times m^3}$ and head related rotation $\bm{R}_t^{x,n}\in SO(3)$, translation $\bm{d}_t^{x,n}\in \mathbb{R}^3$ and scale $\bm{s}_t^{x, n}\in \mathbb{R}$. 

\noindent\textbf{Differentiable Volumetric Raymarching.}
Given all volume rotations $\bm{R}_t^{all}=[\bm{R}_t^{x,n}, \bm{R}_t^{p,n}]$, translations $\bm{d}_t^{all}=[\bm{d}_t^{x,n}, \bm{d}_t^{p,n}]$, scales $\bm{s}_t^{all}=[\bm{s}_t^{x, n}, \bm{s}_t^{p, n}]$ and local radiance fields $\bm{V}_t^{all}=[\bm{U}_t, \bm{V}_{t}]$, we can render them into image $\mathcal{I}_{cam_i}$ and compare it with $I_{cam_i}$ to optimize all volumes.
Using an optimized BVH implementation similar to MVP~\cite{steve_mvp}, we can efficiently determine how each ray intersects with each volume.
We define a ray as $\bm{r}(\mathfrak{p}, l)=\bm{o}(\mathfrak{p})+l\bm{v}(\mathfrak{p})$ shooting from pixel $\mathfrak{p}$ in direction of $\bm{v}(\mathfrak{p})$ with a depth $l$ in range of $(l_{min}, l_{max})$.
The differentiable formation of an image given the volumes can then be formulated as below:
\begin{align*}
    \mathcal{I}_\mathfrak{p} &= \int_{l_{min}}^{l_{max}} \mathbf{V}_{t,rgb}^{all}(\mathbf{r}_\mathfrak{p}(l))\frac{dT(l)}{dl}dl, \\
    T(l) &= min(\int_{l_{min}}^{l} \mathbf{V}_{t,\alpha}^{all}(\mathbf{r}_\mathfrak{p}(l))dl , 1),
\end{align*}
\noindent where $\bm{V}_{t,rgb}^{all}$ is the RGB part of $\bm{V}_t^{all}$ and $\bm{V}_{t,\alpha}^{all}$ is the alpha part of $\bm{V}_t^{all}$.
To get the full rendering, we composite the rendered image as $\mathcal{\tilde{I}}_\mathfrak{p}=\mathcal{I}_\mathfrak{p} + (1-\mathcal{A}_\mathfrak{p})I_{\mathfrak{p},bg}$ where $\mathcal{A}_\mathfrak{p}=T(l_{max})$ and $I_{\mathfrak{p},bg}$ is the background image.
We optimize the following loss to train the volume decoder:
\begin{align*}
    \mathcal{L}_{pho} = ||\mathcal{\tilde{I}}_\mathfrak{p} - I_{\mathfrak{p},gt}||_1 + \omega_{VGG}\mathcal{L}_{VGG}(\mathcal{\tilde{I}}_\mathfrak{p}, I_{\mathfrak{p},gt}), 
\end{align*}
\noindent where $\mathcal{L}_{VGG}$ is the perceptual loss in~\cite{johnson2016perceptual} and $I_{\mathfrak{p},gt}$ is the ground truth pixel value of $\mathfrak{p}$.
We find that the usage of a perceptual loss yields more salient rendering results.

However, as we optimize both $\bm{U}_t$ and $\bm{V}_{t}$ from the images, texture bleeding between the hair volume $\bm{V}_{t}$ and the head volume $\bm{U}_t$ becomes a problem.
The texture bleeding issue is especially undesirable when we want to treat the hair and head separately, for example, when we want to animate just the hair.
To prevent this, we additionally render a hair mask map to regularize both $\bm{V}_{t,\alpha}$ and $\bm{U}_{t,\alpha}$.
We denote the ground truth hair mask as $M_{\mathfrak{p},gt}$ and the rendered hair mask as $\mathcal{M}_\mathfrak{p}$:
%
\begin{align*}
    \mathcal{M}_\mathfrak{p} &= \int_{l_{min}}^{l_{max}} \mathbf{V}_{t,\mathbbm{1}}^{all}(\mathbf{r}_\mathfrak{p}(l))\frac{dT(l)}{dl}dl, \\
    T(l) &= min(\int_{l_{min}}^{l} \mathbf{V}_{t,\alpha}^{all}(\mathbf{r}_\mathfrak{p}(l))dl , 1),
\end{align*}
\noindent where $\mathbf{V}_{t,\mathbbm{1}}^{all}$ is all one volume if it belongs to $\bm{V}_{t}$ otherwise zero. We formulate the segmentation loss as $\mathcal{L}_{mask}=||\mathcal{M}_\mathfrak{p} - M_{\mathfrak{p},gt}||_1$. The final objective for training the whole autoencoder is $\mathcal{L} = \mathcal{L}_{geo} + \mathcal{L}_{pho} + \omega_{mask}\mathcal{L}_{mask}$.
\subsection{Dynamic Model}
In the second stage, we aim to build a dynamic model that can evolve hair states over time without relying on per-frame hair observation as a driving signal.
To achieve that, we leverage the embedding space of hair states built at the state compression stage and train a model that performs the hair state transfer in a supervised manner.
%
%
%
To this end, we build a temporal transfer module ($\mathtt{T^2M}$) of hair dynamic priors, that evolves the hair state and can produce hair animation based on the indirect driving signals of head motion and head-relative gravity direction in a self-evolving manner.
The design of $\mathtt{T^2M}$ is similar to a hair simulator except that it is fully data-driven.
One of the inputs to $\mathtt{T^2M}$ is the encoding $\bm{z}_{t-1}$ of the previous time step.
At the same time, $\mathtt{T^2M}$ is also conditioned on the head per-vertex displacement $\bm{h}_t=\bm{x}_t-\bm{x}_{t-1}$ and $\bm{h}_{t-1}=\bm{x}_{t-1}-\bm{x}_{t-2}$ from the previous two time steps as well as head relative gravity direction $\bm{g}_t\in \mathbb{R}^3$ at the current time step.
$\mathtt{T^2M}$ will then predict the next possible state $\bm{\hat{z}}_{t}$ from $\mathtt{T^2M}$ based on those inputs.
Similar to the design of the VAE, the output of $\mathtt{T^2M}$ is a distribution instead of a single vector.
To be more specific, the mean and standard deviation of $\bm{\hat{z}}_{t}$ are $\mu(\bm{\hat{z}}_{t}),\delta(\bm{\hat{z}}_{t})=\mathtt{T^2M}(\bm{z}_{t-1}|\bm{h}_t, \bm{h}_{t-1}, \bm{g}_t)$.
During training, we take $\bm{\hat{z}}_{t}=\mu(\bm{\hat{z}}_{t})+\mathfrak{n}\odot\delta(\bm{\hat{z}}_{t})$ and during testing $\bm{\hat{z}}_{t}=\mu(\bm{\hat{z}}_{t})$.
The per point normal distribution vector, $\mathfrak{n}$, is the same shape as $\delta(\bm{\hat{z}}_{t})$ and $\odot$ is the element-wise multiplication. 

\noindent\textbf{Training Objectives.} %
We denote the point encoder as $\mathcal{E}(\cdot)$ and the point decoder as $\mathcal{D}(\cdot)$.
Across the training of $\mathtt{T^2M}$, we freeze the parameters of both, $\mathcal{E}$ and $\mathcal{D}$.
We denote the unordered point cloud at frame $t$ as $p_t$, its corresponding encoding as $\mu(\bm{z}_t), \delta(\bm{z}_t) = \mathcal{E}(p_t)$ and the tracked point cloud as $q_t=\mathcal{D}(\bm{z}_t)$.
The following loss enforces the prediction of $\mathtt{T^2M}$ to be similar to its ground truth:
\begin{align*}
    &\mathcal{L}_{mse} = ||\mu(\bm{\hat{z}}_{t+1}) - \mu(\bm{z}_{t+1})||_2 + ||\delta(\bm{\hat{z}}_{t+1}) - \delta(\bm{z}_{t+1})||_2 \\
    &\mathcal{L}_{cos} = -\bm{cos}(\mu{\bm{\hat{z}}_{t+1}}, \mu{\bm{z}}_{t+1}) - \bm{cos}(\delta{\bm{\hat{z}}_{t+1}}, \delta{\bm{z}}_{t+1}) \\
    &\mathcal{L}_{ptsmse} = ||\mathcal{D}(\bm{\hat{z}}_{t+1}) - \mathcal{D}(\bm{z}_{t+1})||_2,
\end{align*}
\noindent where we not only minimize the $\ell_2$ distance between $\bm{\hat{z}}_{t+1}$ and $\bm{z}_{t+1}$, but also enforce the cosine similarity and the corresponding tracked point cloud to be equivalent.
To adapt $\mathtt{T^2M}$ to the tracked point cloud $q_t$, we compute the above loss again but using $q_t$ as input, where we generate the corresponding encoding as $\bm{z}'_t$ from $\mathcal{E}(q_t)$ and its prediction $\bm{\hat{z}}'_{t+1}$ from $\mathtt{T^2M}(\bm{z}'_t|\bm{h}_t, \bm{h}_{t-1}, \bm{g}_t)$:
\begin{align*}
    &\mathcal{L}_{mse,cyc} = ||\mu(\bm{\hat{z}'}_{t+1}) - \mu(\bm{z}_{t+1})||_2 + ||\delta(\bm{\hat{z}'}_{t+1}) - \delta(\bm{z}_{t+1})||_2 \\
    &\mathcal{L}_{cos,cyc} = -\bm{cos}(\mu{\bm{\hat{z}'}_{t+1}}, \mu{\bm{z}}_{t+1}) - \bm{cos}(\delta{\bm{\hat{z}'}_{t+1}}, \delta{\bm{z}}_{t+1}) \\
    &\mathcal{L}_{ptsmse,cyc} = ||\mathcal{D}(\bm{\hat{z}'}_{t+1}) - \mathcal{D}(\bm{z}_{t+1})||_2.
\end{align*}
Similar to how we train our autoencoder, we also enforce two KL divergence losses on both the predicted $\bm{\hat{z}}_{t+1}$ and $\bm{\hat{z}'}_{t+1}$ with a normal distribution $\mathcal{N}$.
The final objective for training the $\mathtt{T^2M}$ is a weighted sum of the above eight terms.

\noindent\textbf{Animation.} 
%
Given an initialized hair state, our dynamic model $\mathtt{T^2M}$ can evolve the hair state into future states conditioned on head motion and head-relative gravity direction. 
%
%
One straightforward implementation of the  would be to directly propagate the hair state encoding $\bm{z}_t$.
However, in practice, we find this leads to severe drift in the semantic space.
As a simple feed forward neural network, $\mathtt{T^2M}$ can not guarantee that its output is noise free. 
The noise in the output becomes even more problematic when we use $\mathtt{T^2M}$ in a recurrent manner, where the output noise will aggregate and lead to drift. 
%
To remedy this, instead of propagating the encoding $\bm{z}_t$ directly, we reproject the predicted encoding $\bm{z}_t$ by the point autoencoder $\mathcal{E}$ and $\mathcal{D}$ every time for denoising. 
To be more specific, we acquire the de-noised predicted hair encoding $\bm{\hat{z}}_{t+1}=\mathcal{E}(\mathcal{D}(\bm{\hat{z}}_{t+1}))$ from the raw prediction $\bm{\hat{z}}_{t+1}$ of $\mathtt{T^2M}$.
The use of the point autoencoder $\mathcal{E}$ and $\mathcal{D}$ can help us remove the noise in $\bm{z}_{t+1}$, as the point cloud encoder $\mathcal{E}$ can regress the mean $\mu_{t+1}$ and standard deviation $\delta_{t+1}$ of $\bm{z}_{t+1}$ separately by using $q_{t+1}$ as an intermediate variable.
Thus, we can extract the noise free part of $\bm{z}_{t+1}$ by taking the mean $\mu_{t+1}$ regressed from $\mathcal{E}$. 
%
Please see our experiments for further details of this approach.

%% file: experiment.tex
\section{Experiments}


In order to test our proposed model, 
we conduct experiments on both the hair motion data set presented in HVH~\cite{wang2021hvh} and our own dataset with longer sequences following a similar capture protocol as HVH~\cite{wang2021hvh}.
%
%
We collect a total of four different hair wig styles with scripted head motions like nodding, swinging and tilting.
%
%
We also collect an animation test set with the same scripted head motions performed by different actors wearing a wig cap, which we will refer to as ``bald head motion sequence''.
The animation test set contains both single view captures from a smart phone and multiview captures.
The total length of each hair wig capture is around 1-1.5 minutes with a frame rate of 30Hz.
100 cameras are used during the capture where 93 of them are used to obtain training views and the rest are providing held-out test views.
We split each sequence into two folds with similar amounts of frames and train our model exclusively on the training portion of each sequence.

\subsection{Evaluation of the State Compression Model}

We first test our state compression model to evaluate its ability to reconstruct the appearance of hair and head.

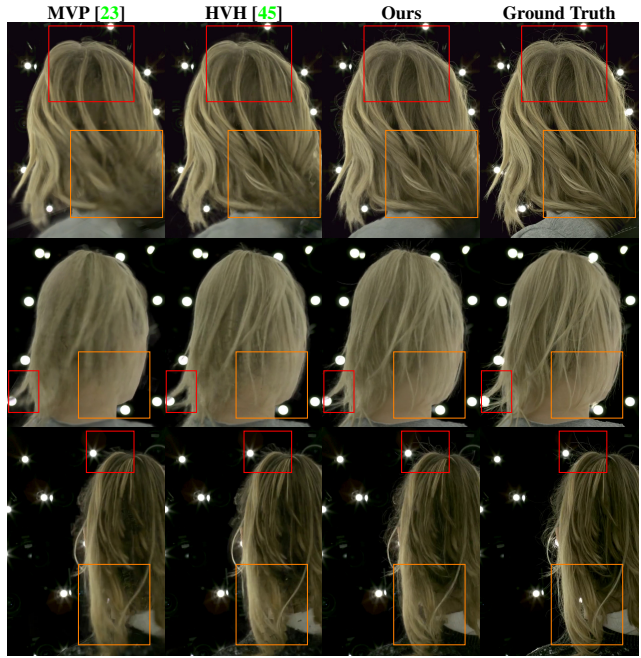
\begin{figure}[tb]
\setlength\tabcolsep{0pt}
\renewcommand{\arraystretch}{0}
\centering
\begin{tabular}{cccc}
 \textbf{\scriptsize MVP~\cite{steve_mvp}} &
 \textbf{\scriptsize HVH~\cite{wang2021hvh}} &
 \textbf{\scriptsize Ours} & 
 \textbf{\scriptsize Ground Truth} \\
 \begin{tikzpicture}
    \node[anchor=south west,inner sep=0] (image) at (0,0) {\adjincludegraphics[width=0.12\textwidth, trim={{0.05\width} {0.1\height} 0 {0.05\height}}, clip]{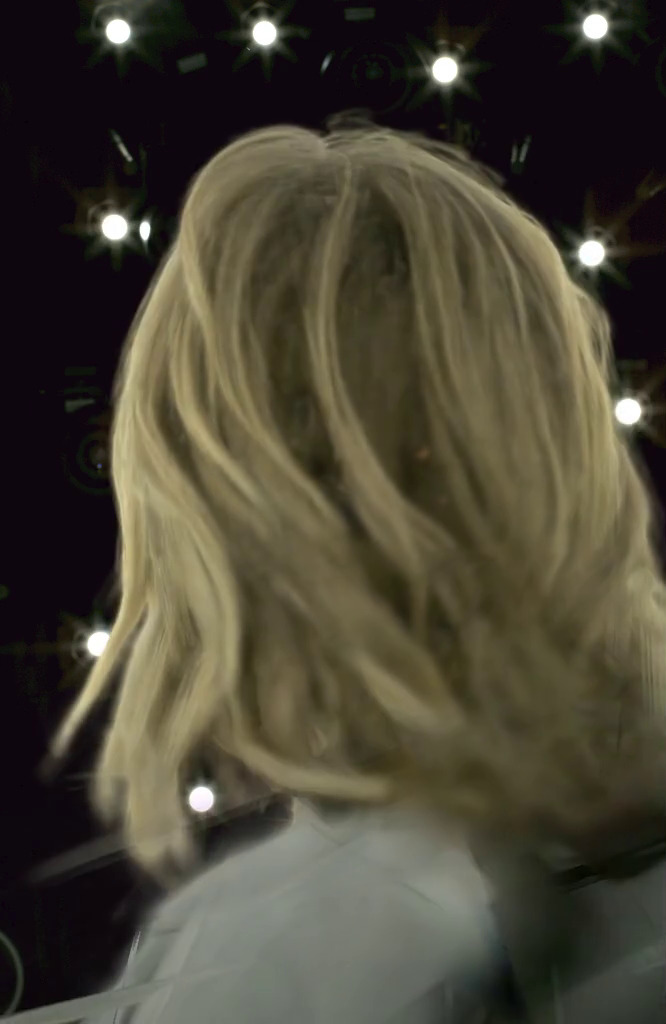}};
    \begin{scope}[x={(image.south east)},y={(image.north west)}]
        \draw[red] (0.263,0.632) rectangle (0.8,0.98);
    \end{scope}
    \begin{scope}[x={(image.south east)},y={(image.north west)}]
        \draw[orange] (0.4,0.1) rectangle (0.98,0.5);
    \end{scope}
 \end{tikzpicture}
&
\begin{tikzpicture}
    \node[anchor=south west,inner sep=0] (image) at (0,0) {\adjincludegraphics[width=0.12\textwidth, trim={{0.05\width} {0.1\height} 0 {0.05\height}}, clip]{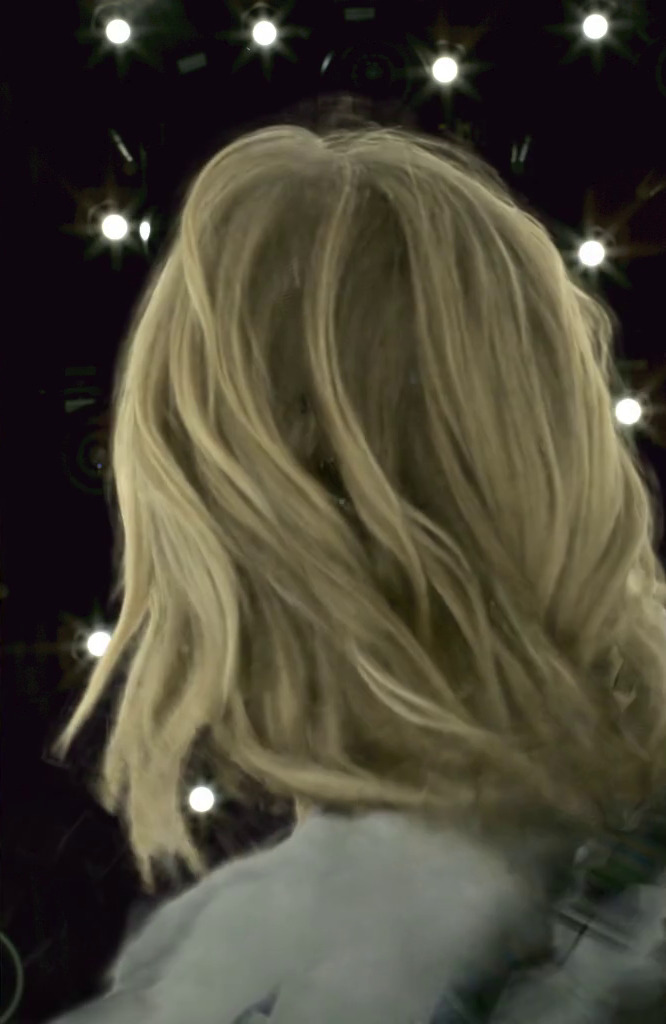}};
    \begin{scope}[x={(image.south east)},y={(image.north west)}]
        \draw[red] (0.263,0.632) rectangle (0.8,0.98);
    \end{scope}
    \begin{scope}[x={(image.south east)},y={(image.north west)}]
        \draw[orange] (0.4,0.1) rectangle (0.98,0.5);
    \end{scope}
 \end{tikzpicture}
&
\begin{tikzpicture}
    \node[anchor=south west,inner sep=0] (image) at (0,0) {\adjincludegraphics[width=0.12\textwidth, trim={{0.05\width} {0.1\height} 0 {0.05\height}}, clip]{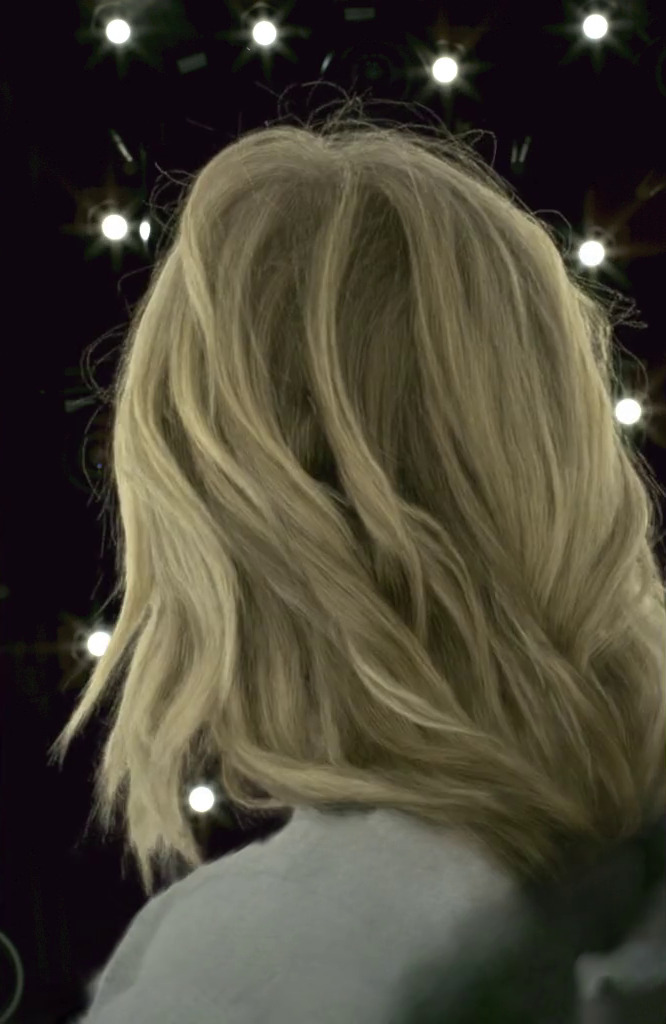}};
    \begin{scope}[x={(image.south east)},y={(image.north west)}]
        \draw[red] (0.263,0.632) rectangle (0.8,0.98);
    \end{scope}
    \begin{scope}[x={(image.south east)},y={(image.north west)}]
        \draw[orange] (0.4,0.1) rectangle (0.98,0.5);
    \end{scope}
 \end{tikzpicture}
&
\begin{tikzpicture}
    \node[anchor=south west,inner sep=0] (image) at (0,0) {\adjincludegraphics[width=0.12\textwidth, trim={{0.05\width} {0.1\height} 0 {0.05\height}}, clip]{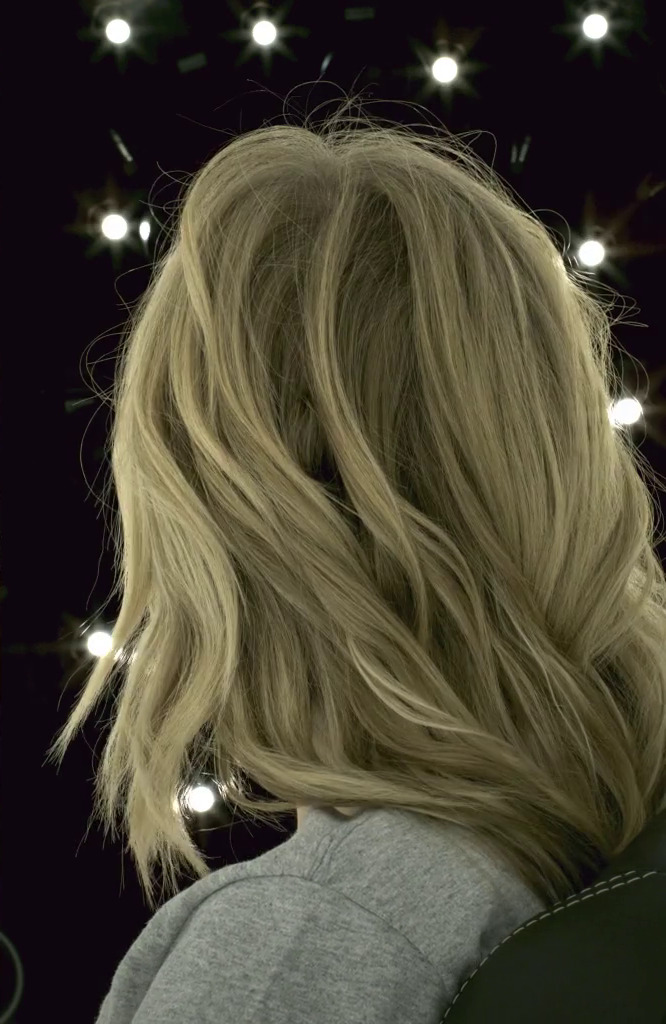}};
    \begin{scope}[x={(image.south east)},y={(image.north west)}]
        \draw[red] (0.263,0.632) rectangle (0.8,0.98);
    \end{scope}
    \begin{scope}[x={(image.south east)},y={(image.north west)}]
        \draw[orange] (0.4,0.1) rectangle (0.98,0.5);
    \end{scope}
 \end{tikzpicture}
\\ 
\begin{tikzpicture}
    \node[anchor=south west,inner sep=0] (image) at (0,0) {\adjincludegraphics[width=0.12\textwidth, trim={{0 {0.3\height} {0.1\width} 0}}, clip]{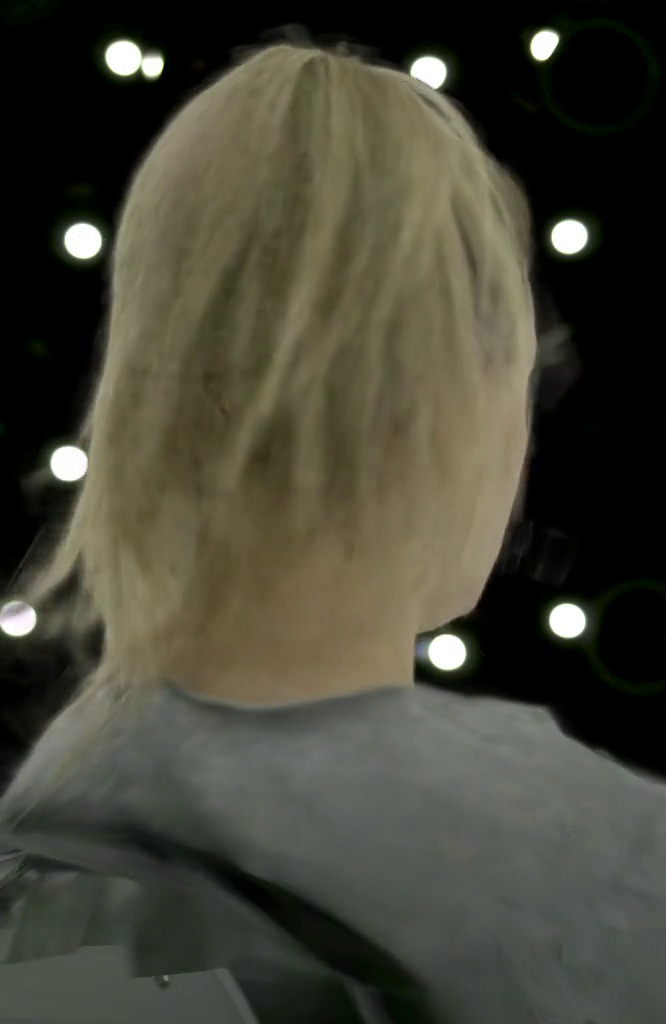}};
    \begin{scope}[x={(image.south east)},y={(image.north west)}]
        \draw[red] (0.01,0.08) rectangle (0.2,0.3);
    \end{scope}
    \begin{scope}[x={(image.south east)},y={(image.north west)}]
        \draw[orange] (0.45,0.05) rectangle (0.9,0.4);
    \end{scope}
 \end{tikzpicture}
&
\begin{tikzpicture}
    \node[anchor=south west,inner sep=0] (image) at (0,0) {\adjincludegraphics[width=0.12\textwidth, trim={{0 {0.3\height} {0.1\width} 0}}, clip]{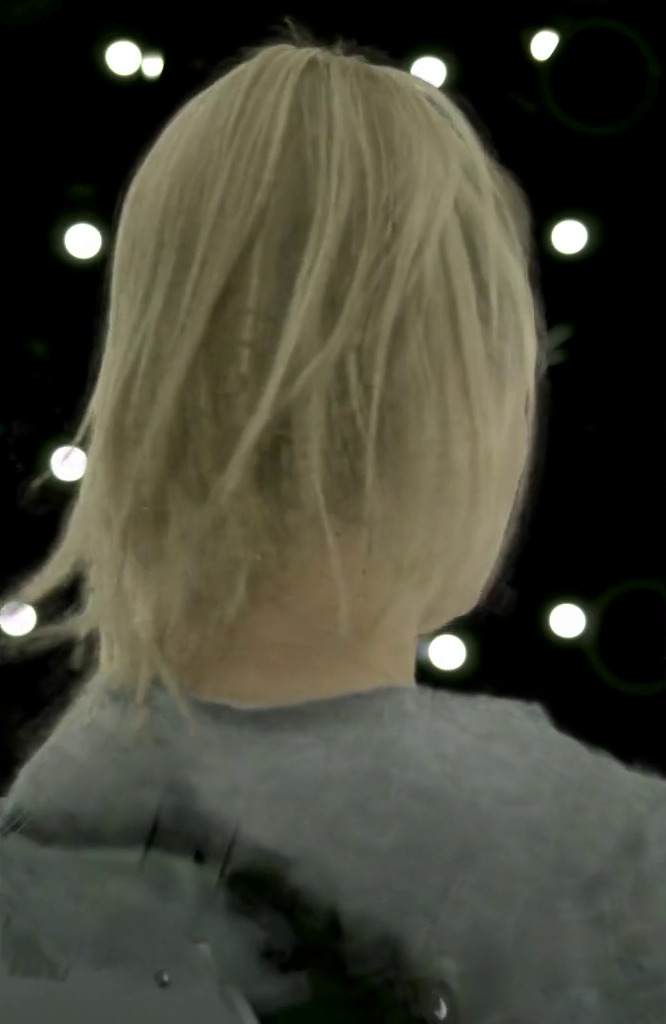}};
    \begin{scope}[x={(image.south east)},y={(image.north west)}]
        \draw[red] (0.01,0.08) rectangle (0.2,0.3);
    \end{scope}
    \begin{scope}[x={(image.south east)},y={(image.north west)}]
        \draw[orange] (0.45,0.05) rectangle (0.9,0.4);
    \end{scope}
 \end{tikzpicture}
 &
\begin{tikzpicture}
    \node[anchor=south west,inner sep=0] (image) at (0,0) {\adjincludegraphics[width=0.12\textwidth, trim={{0 {0.3\height} {0.1\width} 0}}, clip]{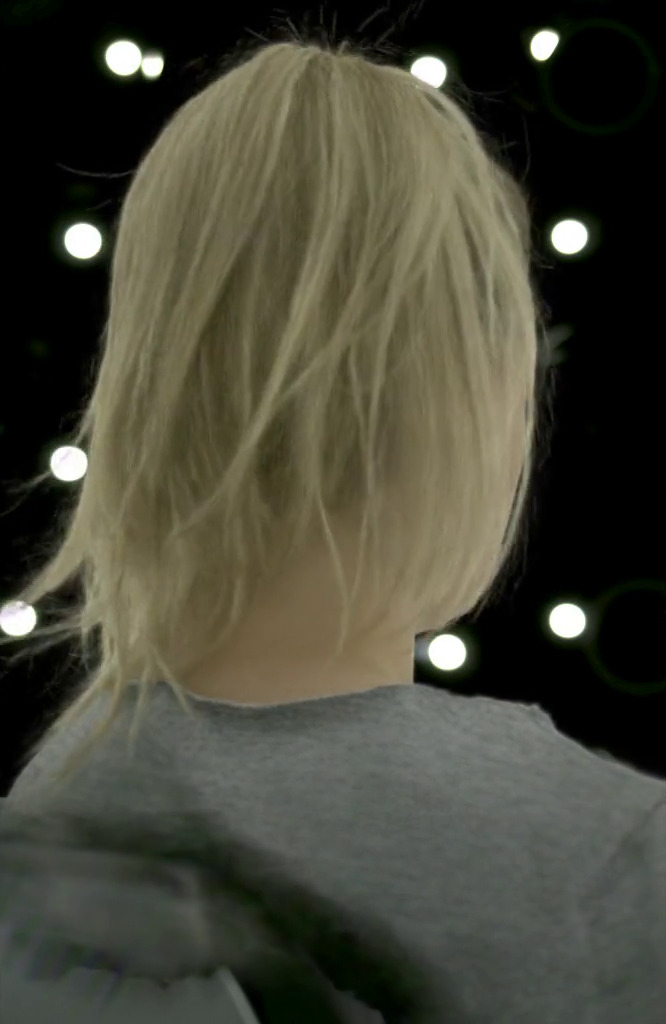}};
    \begin{scope}[x={(image.south east)},y={(image.north west)}]
        \draw[red] (0.01,0.08) rectangle (0.2,0.3);
    \end{scope}
    \begin{scope}[x={(image.south east)},y={(image.north west)}]
        \draw[orange] (0.45,0.05) rectangle (0.9,0.4);
    \end{scope}
 \end{tikzpicture} 
 &
\begin{tikzpicture}
    \node[anchor=south west,inner sep=0] (image) at (0,0) {\adjincludegraphics[width=0.12\textwidth, trim={{0 {0.3\height} {0.1\width} 0}}, clip]{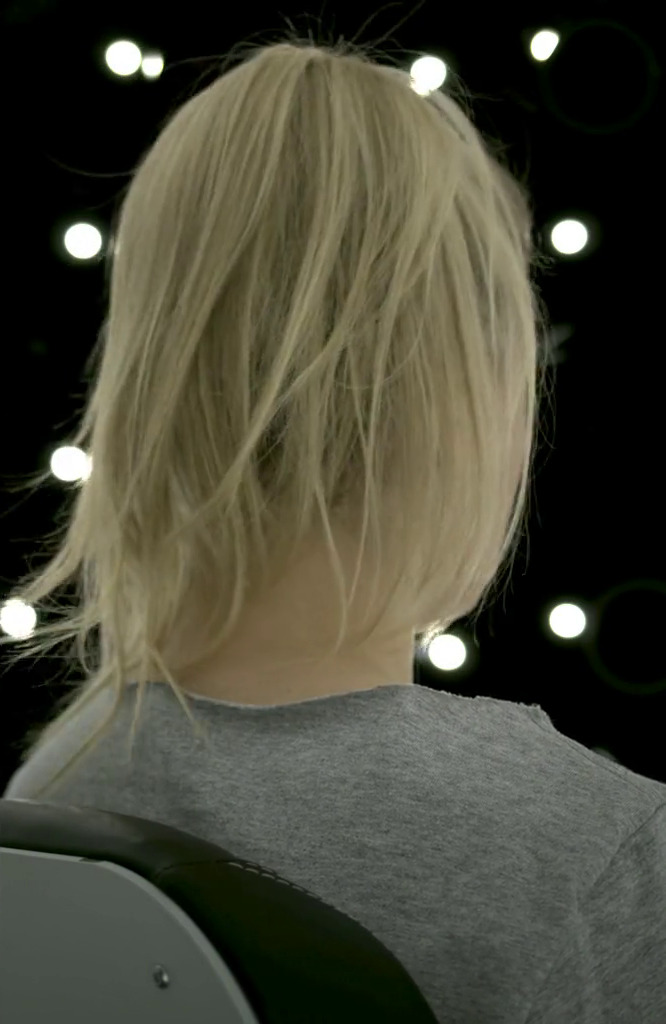}};
    \begin{scope}[x={(image.south east)},y={(image.north west)}]
        \draw[red] (0.01,0.08) rectangle (0.2,0.3);
    \end{scope}
    \begin{scope}[x={(image.south east)},y={(image.north west)}]
        \draw[orange] (0.45,0.05) rectangle (0.9,0.4);
    \end{scope}
 \end{tikzpicture} 
 \\
 \begin{tikzpicture}
    \node[anchor=south west,inner sep=0] (image) at (0,0) {\adjincludegraphics[width=0.12\textwidth, trim={{0.1\width} {0.1\height} 0 {0.05\height}}, clip]{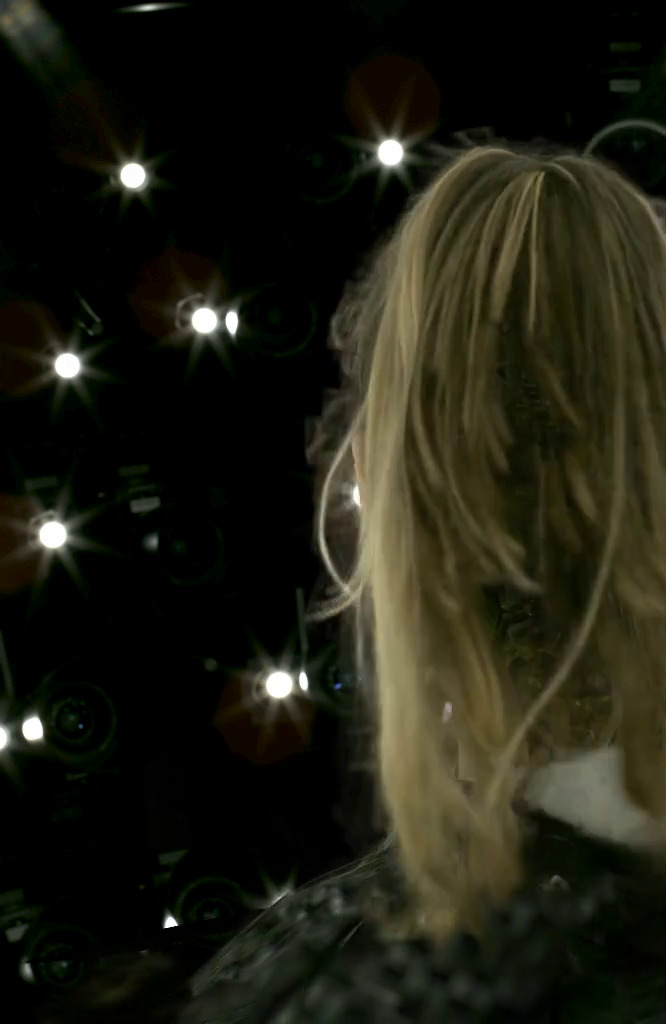}};
    \begin{scope}[x={(image.south east)},y={(image.north west)}]
        \draw[red] (0.5,0.8) rectangle (0.8,0.98);
    \end{scope}
    \begin{scope}[x={(image.south east)},y={(image.north west)}]
        \draw[orange] (0.45,0.05) rectangle (0.9,0.4);
    \end{scope}
 \end{tikzpicture}
 &
\begin{tikzpicture}
    \node[anchor=south west,inner sep=0] (image) at (0,0) {\adjincludegraphics[width=0.12\textwidth, trim={{0.1\width} {0.1\height} 0 {0.05\height}}, clip]{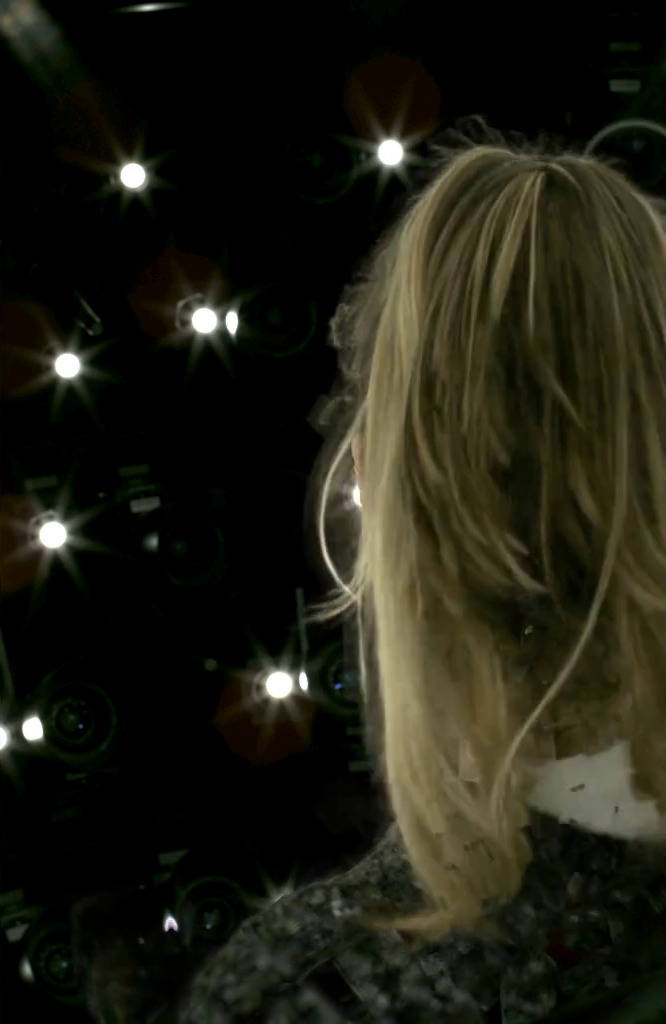}};
    \begin{scope}[x={(image.south east)},y={(image.north west)}]
        \draw[red] (0.5,0.8) rectangle (0.8,0.98);
    \end{scope}
    \begin{scope}[x={(image.south east)},y={(image.north west)}]
        \draw[orange] (0.45,0.05) rectangle (0.9,0.4);
    \end{scope}
 \end{tikzpicture} 
 &
\begin{tikzpicture}
    \node[anchor=south west,inner sep=0] (image) at (0,0) {\adjincludegraphics[width=0.12\textwidth, trim={{0.1\width} {0.1\height} 0 {0.05\height}}, clip]{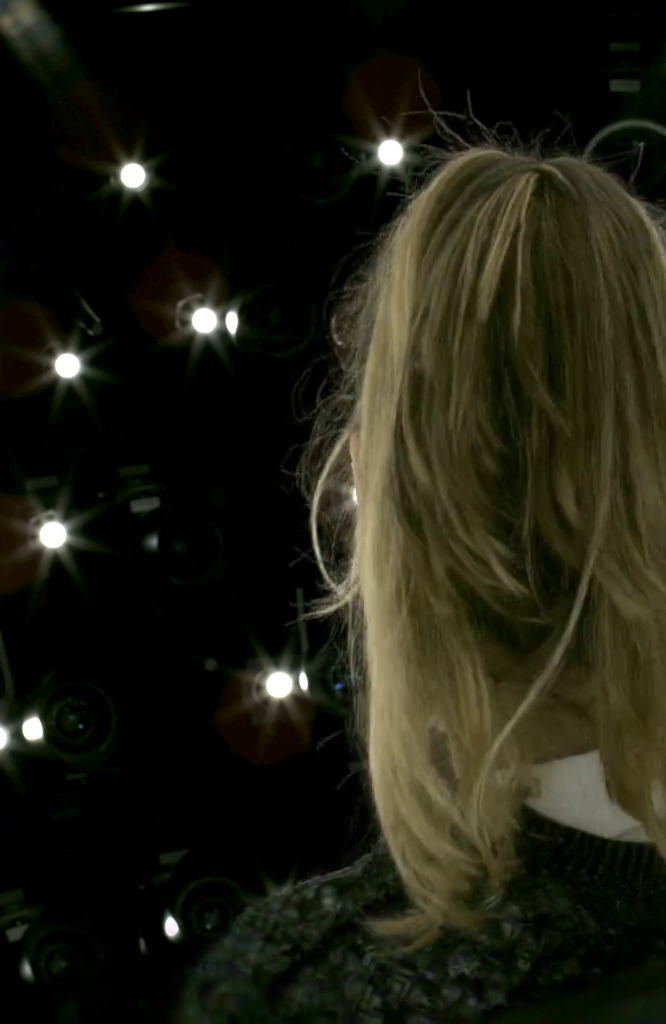}};
    \begin{scope}[x={(image.south east)},y={(image.north west)}]
        \draw[red] (0.5,0.8) rectangle (0.8,0.98);
    \end{scope}
    \begin{scope}[x={(image.south east)},y={(image.north west)}]
        \draw[orange] (0.45,0.05) rectangle (0.9,0.4);
    \end{scope}
 \end{tikzpicture} 
 &
\begin{tikzpicture}
    \node[anchor=south west,inner sep=0] (image) at (0,0) {\adjincludegraphics[width=0.12\textwidth, trim={{0.1\width} {0.1\height} 0 {0.05\height}}, clip]{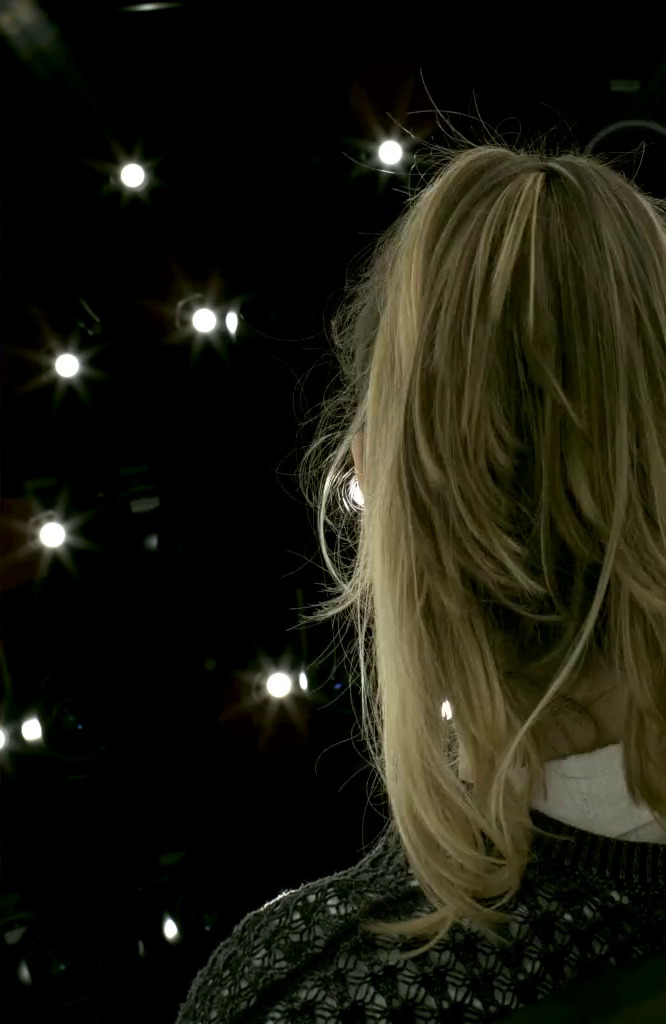}};
    \begin{scope}[x={(image.south east)},y={(image.north west)}]
        \draw[red] (0.5,0.8) rectangle (0.8,0.98);
    \end{scope}
    \begin{scope}[x={(image.south east)},y={(image.north west)}]
        \draw[orange] (0.45,0.05) rectangle (0.9,0.4);
    \end{scope}
 \end{tikzpicture}
\end{tabular}
\caption{\label{fig1:nvs_comp}\textbf{Novel View Synthesis.} Compared with previous methods, our method captures hair with more details, including fly-away hair strands and creates an overall more accurate hair reconstruction with perceptually better rendering results.}
\end{figure}

\noindent\textbf{Novel View Synthesis.}
We compare with volumetric methods like NeRF based methods~\cite{li2021nsff, tretschk2021nrnerf} and  volumetric primitives based methods~\cite{steve_mvp,wang2021hvh} on the data set from HVH~\cite{wang2021hvh}.
In Tab~\ref{tab1:nvs}, we show the reconstruction related metrics MSE, SSIM, PSNR and LPIPS~\cite{zhang2018perceptual} between predicted images and the ground truth image on hold out views.
%
Our method yields a good balance between perceptual loss and reconstruction loss while keeping both of them relatively low.
Furthermore, our method achieves a much higher perceptual similarity with ground truth images.
In Fig.~\ref{fig1:nvs_comp} we show that our method can capture high frequency details and even preserve some fly-away hair strands.


\definecolor{gold}{rgb}{1.0, 0.84, 0.0}
\newcommand{\gB}[1]{\fcolorbox{white}{gold}{#1}}
\definecolor{silver}{rgb}{0.75, 0.75, 0.75}
\newcommand{\sB}[1]{\fcolorbox{white}{silver}{#1}}
\definecolor{bronze}{rgb}{0.8, 0.5, 0.2}
\newcommand{\bB}[1]{\fcolorbox{white}{bronze}{#1}}

\begin{table}[tb]
\centering
\resizebox{\columnwidth}{!}{
\begin{tabular}{c|cccc|cccc|cccc|}
\multirow{2}{*}{} & \multicolumn{4}{c|}{seq01}                                                                         & \multicolumn{4}{c|}{seq02}                                                                         & \multicolumn{4}{c|}{seq03}                                                                         \\ \cline{2-13} 
                  & \multicolumn{1}{c|}{MSE$\downarrow$} & \multicolumn{1}{c|}{PSNR$\uparrow$} & \multicolumn{1}{c|}{SSIM$\uparrow$} & LPIPS$\downarrow$           & \multicolumn{1}{c|}{MSE$\downarrow$} & \multicolumn{1}{c|}{PSNR$\uparrow$} & \multicolumn{1}{c|}{SSIM$\uparrow$} & LPIPS$\downarrow$           & \multicolumn{1}{c|}{MSE$\downarrow$} & \multicolumn{1}{c|}{PSNR$\uparrow$} & \multicolumn{1}{c|}{SSIM$\uparrow$} & LPIPS$\downarrow$          \\ \hline
PFNeRF & 51.25 & 31.16 & 0.9269 & 0.3717 & 103.41 & 28.15 & 0.8659 & 0.5067 & 76.59 & 29.50 & 0.9000 &  0.2949 \\
NSFF   & 50.13 & 31.21 & 0.9346 & 0.3672 & 90.06 & 28.75 & 0.8885 & 0.4728 & 83.18 & 29.10  & 0.8936 & 0.3292 \\
NRNeRF & 56.78 & 30.78 & 0.9231 & 0.3554 & 132.16 & 27.13 & 0.8549 & 0.5241 & 79.83 & 29.33 & 0.8987 & 0.3067 \\ \hline
MVP    & 47.54 & 31.60 & 0.9476 & 0.2587 & 77.23 & 29.62 & 0.9088 & 0.3051 & 73.78 & 29.66 & 0.9224 & 0.2455          \\
HVH    & \sB{41.89} & \sB{32.17} & \gB{0.9543} & \sB{0.2019} & \sB{59.84} & \sB{30.69} & \sB{0.9275} & \sB{0.2353} & \gB{71.58} & \gB{29.81} & \gB{0.9314} & \sB{0.2021}\\ \hline
Ours   & \gB{40.34}& \gB{32.28}  & \sB{0.9558} & \gB{0.1299} & \gB{56.47} & \gB{30.94} & \gB{0.9329}& \gB{0.1254} & \sB{73.65} & \sB{29.69} & \sB{0.9247} & \gB{0.1496}
\end{tabular}
}
\caption{\label{tab1:nvs}\textbf{Novel view synthesis.} We compute MSE$\downarrow$, PSNR$\uparrow$, SSIM$\uparrow$ and LPIPS$\downarrow$ comparing rendered and ground truth images on hold-out views. \gB{First} and \sB{second} best results are highlighted.}
\end{table}

\begin{figure}[tb]
\setlength\tabcolsep{0pt}
\renewcommand{\arraystretch}{0}
\centering
\begin{tabular}{cccc}
 \textbf{\scriptsize HVH~\cite{wang2021hvh} Seq01} &
 \textbf{\scriptsize Ours Seq01} &
 \textbf{\scriptsize HVH~\cite{wang2021hvh} Seq02} & 
 \textbf{\scriptsize Ours Seq02} \\
 \adjincludegraphics[width=0.12\textwidth, trim={{0.2\width} {0.12\height} 0 0}, clip]{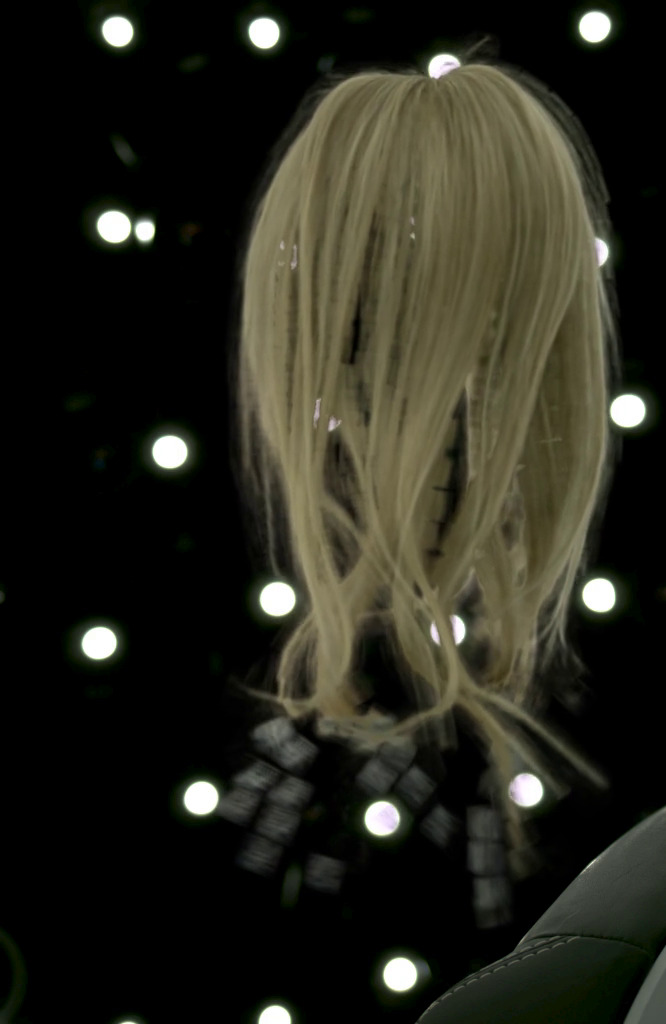} &
 \adjincludegraphics[width=0.12\textwidth, trim={{0.2\width} {0.12\height} 0 0}, clip]{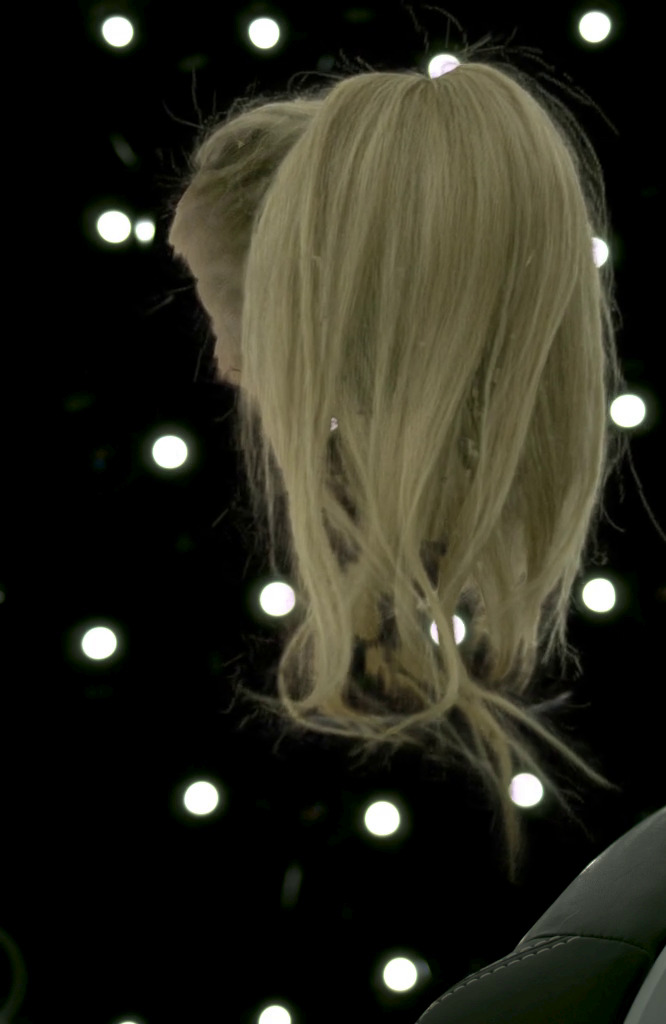} &
 \adjincludegraphics[width=0.12\textwidth, trim={{0.2\width} {0.12\height} 0 0}, clip]{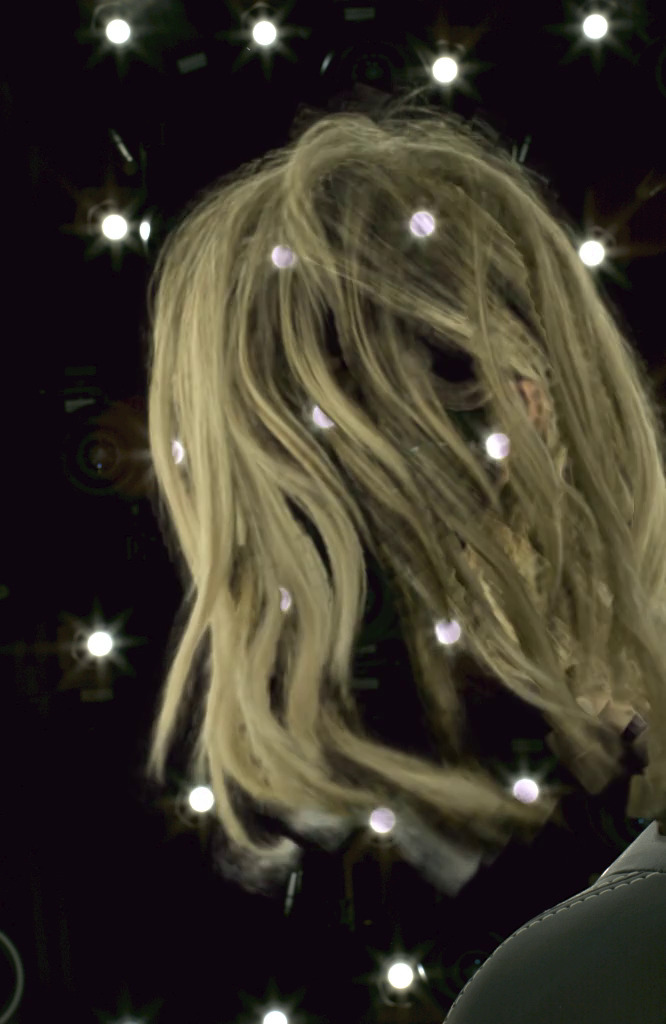} &
 \adjincludegraphics[width=0.12\textwidth, trim={{0.2\width} {0.12\height} 0 0}, clip]{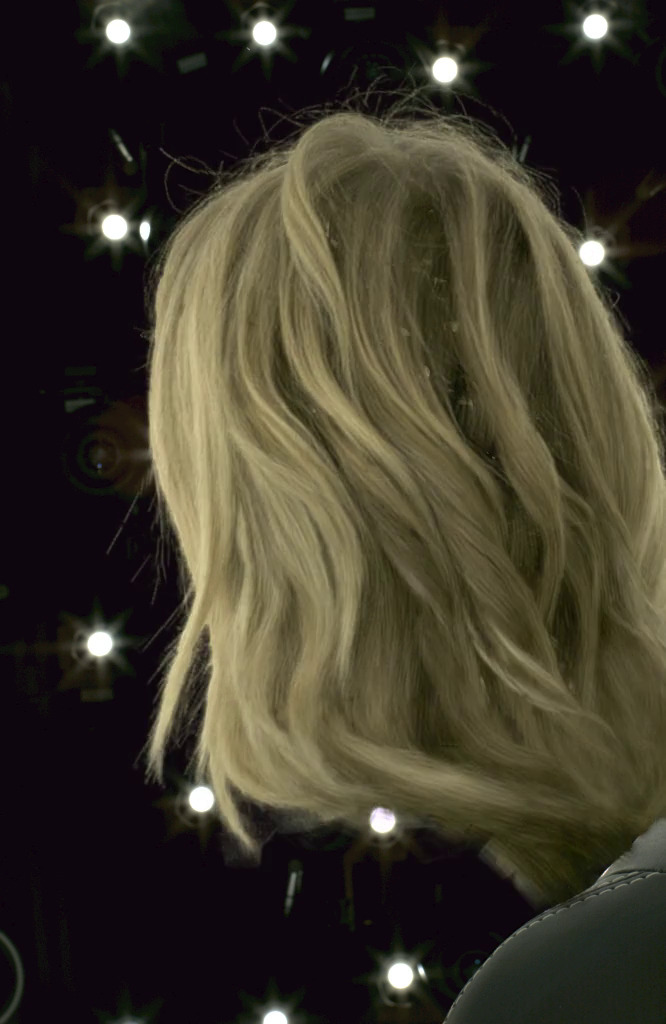} 
\end{tabular}
\caption{\label{fig1:hair_silh}\textbf{Hair/Head Disentanglement.} By explicitly enforcing the semantic segmentation of head and hair through additional supervision, we learn a more opaque hair texture while the result suffers less from texture bleeding.}
\end{figure}

\noindent\textbf{Ablation on Hair/Head Disentanglement.}
We test how well our model handles hair/head disentanglement compared to previous work.
As hair and head are exhibiting different dynamic patterns, disentanglement is usually required, especially to achieve independent controllability for both.
%
%
%
We compare with HVH, which implicitly separates the hair and head using the dynamic discrepancy between them and optical flow.
In our model, we further facilitate the disentanglement by using semantic segmentation.
In Tab~\ref{tab1:iou}, we show the IoU between the rendered silhouette of hair volumes and ground truth hair segmentatation of the different methods.
We visualize the difference in Fig.~\ref{fig1:hair_silh}.
Our method generates a more opaque hair texture with less texture bleeding between hair volumes and non-hair volumes.
%
Moreover, our model creates the hair shape in an entirely data-driven fashion, which yields higher fidelity results than the artist prepared hair in HVH~\cite{wang2021hvh}.

\begin{table}[tb]
\centering
\begin{tabular}{c|ccc}
      & Seq01    & Seq02 & Seq03   \\ \hline
HVH~\cite{wang2021hvh} & 0.6685 & 0.4121 & 0.3766 \\ 
Ours &  \textbf{0.8289} & \textbf{0.9243} & \textbf{0.8571}
\end{tabular}
\caption{\label{tab1:iou}\textbf{IoU}($\uparrow$) between rendered hair silhouette and ground truth hair segmentation. \iffalse\MZ{Do we want to highlight our numbers in bold text, since we are better?}\fi}
\end{table}

\begin{figure}[tb]
\setlength\tabcolsep{0pt}
\renewcommand{\arraystretch}{0}
\centering
\begin{tabular}{cccc}
 \textbf{\scriptsize VGG 0.0} &
 \textbf{\scriptsize VGG 1.0} &
 \textbf{\scriptsize VGG 10.0} & 
 \textbf{\scriptsize GT} \\
 {\setlength{\fboxsep}{0pt}\setlength{\fboxrule}{1pt}\fcolorbox{red}{white}{\adjincludegraphics[width=0.12\textwidth, trim={{0.31\width} {0.4\height} {0.21\width} {0.26\height}}, clip]{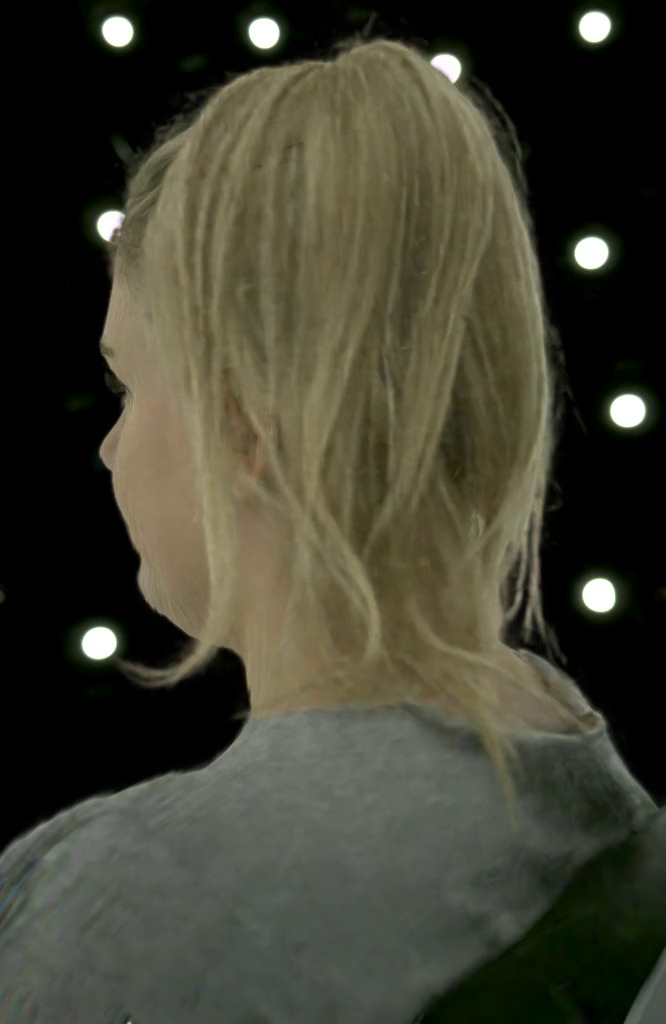}}} &
 {\setlength{\fboxsep}{0pt}\setlength{\fboxrule}{1pt}\fcolorbox{red}{white}{\adjincludegraphics[width=0.12\textwidth, trim={{0.31\width} {0.4\height} {0.21\width} {0.26\height}}, clip]{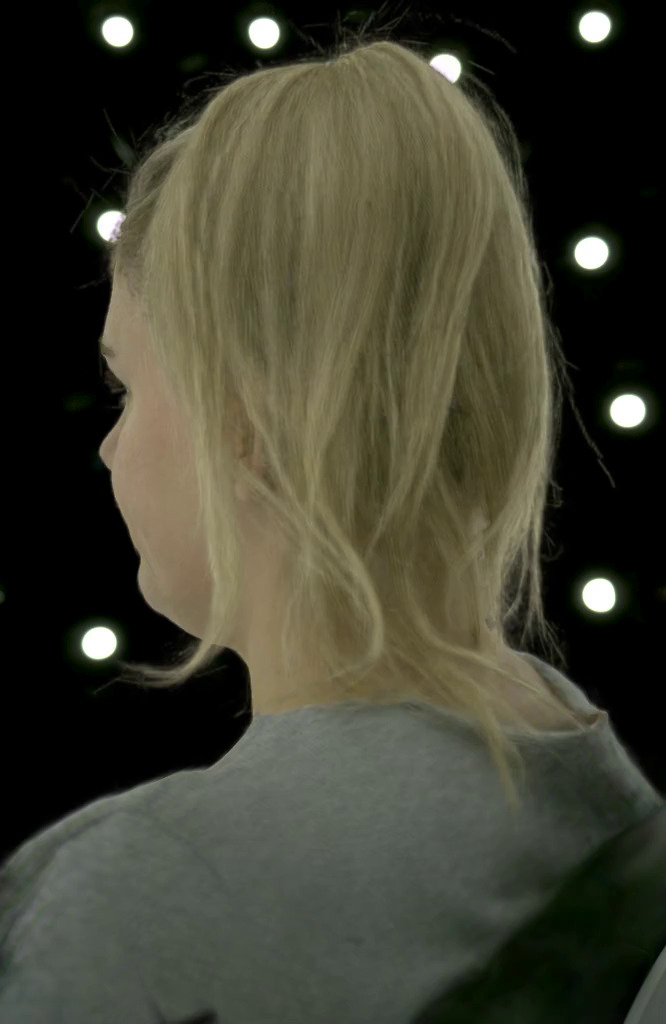}}} &
 {\setlength{\fboxsep}{0pt}\setlength{\fboxrule}{1pt}\fcolorbox{red}{white}{\adjincludegraphics[width=0.12\textwidth, trim={{0.31\width} {0.4\height} {0.21\width} {0.26\height}}, clip]{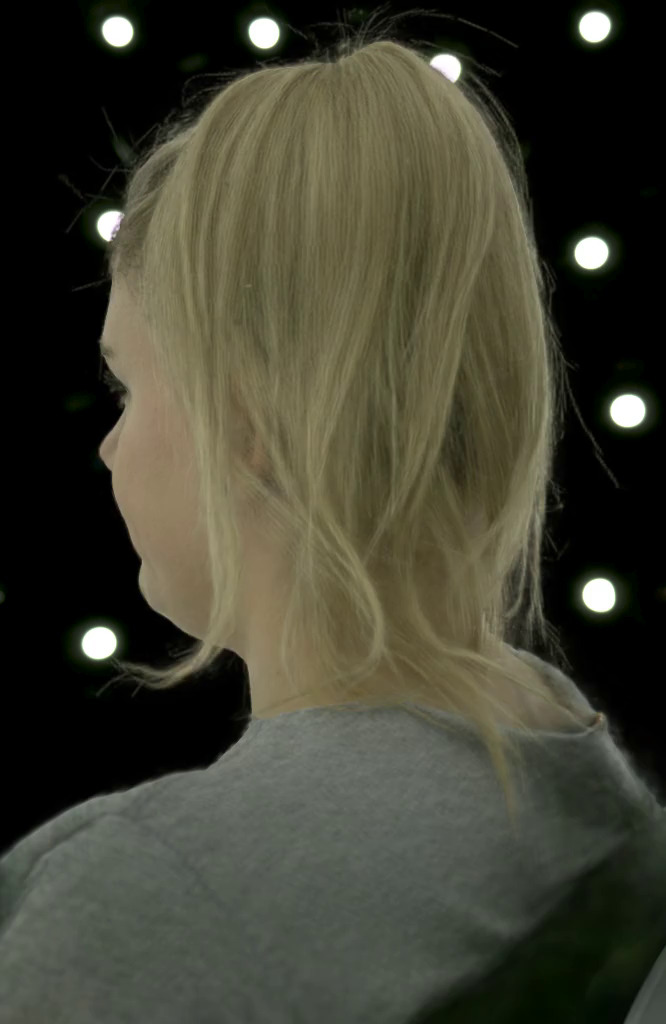}}} &
 {\setlength{\fboxsep}{0pt}\setlength{\fboxrule}{1pt}\fcolorbox{red}{white}{\adjincludegraphics[width=0.12\textwidth, trim={{0.31\width} {0.4\height} {0.21\width} {0.26\height}}, clip]{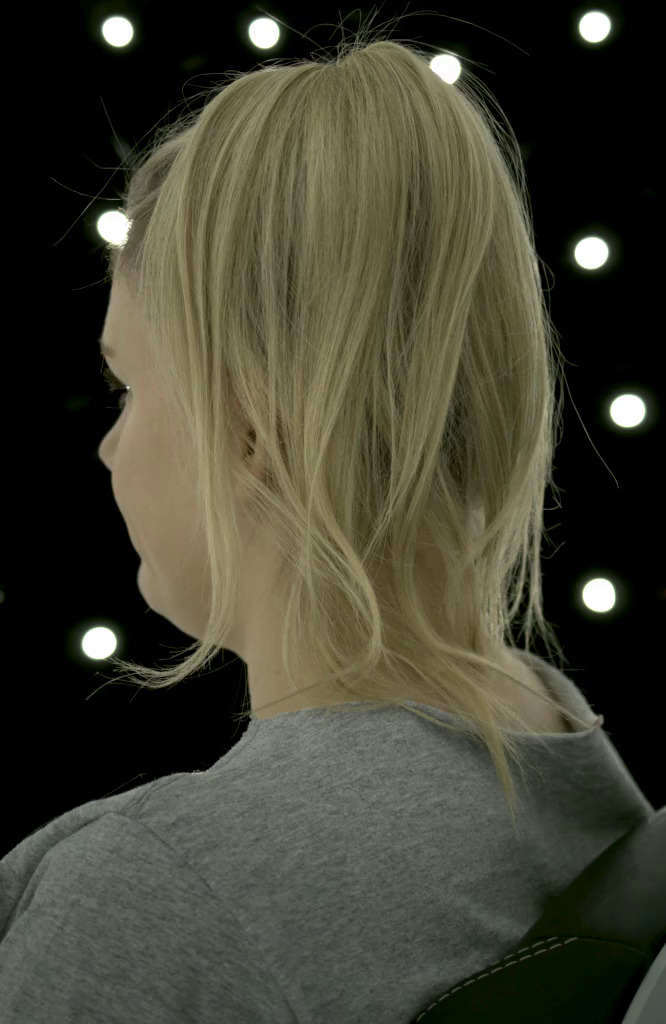}}} \\
 {\setlength{\fboxsep}{0pt}\setlength{\fboxrule}{1pt}\fcolorbox{red}{white}{\adjincludegraphics[width=0.12\textwidth, trim={{0.4\width} {0.52\height} {0.16\width} {0.17\height}}, clip]{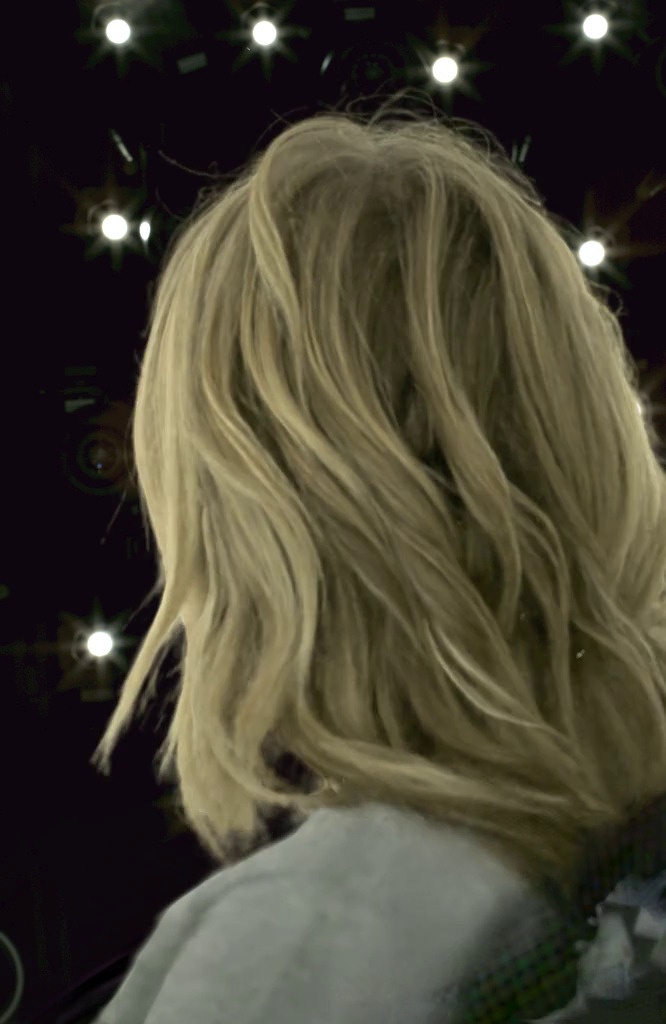}}} &
 {\setlength{\fboxsep}{0pt}\setlength{\fboxrule}{1pt}\fcolorbox{red}{white}{\adjincludegraphics[width=0.12\textwidth, trim={{0.4\width} {0.52\height} {0.16\width} {0.17\height}}, clip]{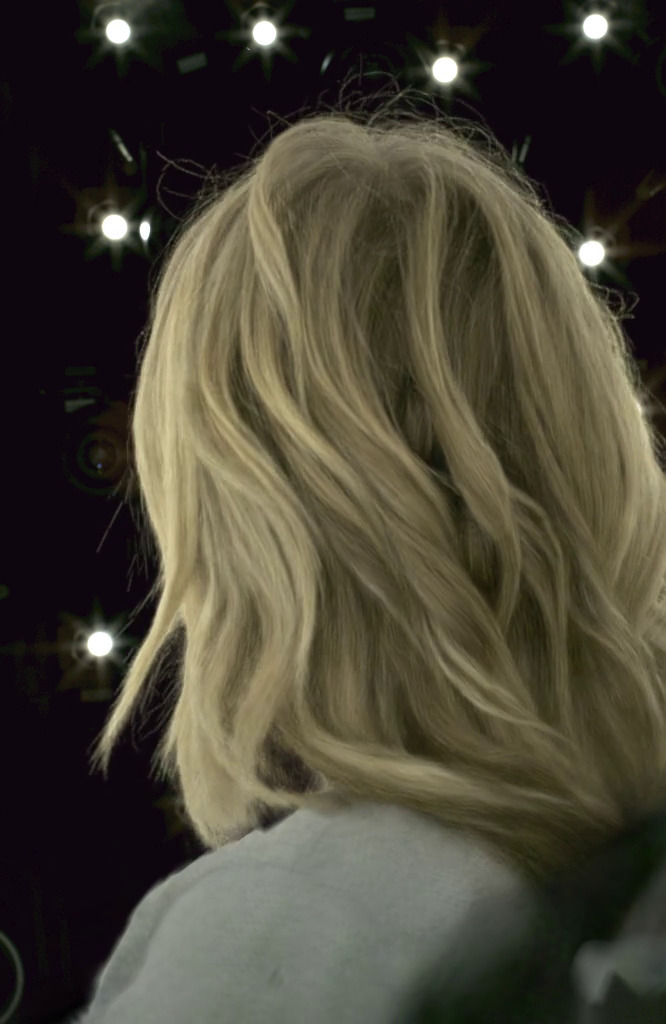}}} &
 {\setlength{\fboxsep}{0pt}\setlength{\fboxrule}{1pt}\fcolorbox{red}{white}{\adjincludegraphics[width=0.12\textwidth, trim={{0.4\width} {0.52\height} {0.16\width} {0.17\height}}, clip]{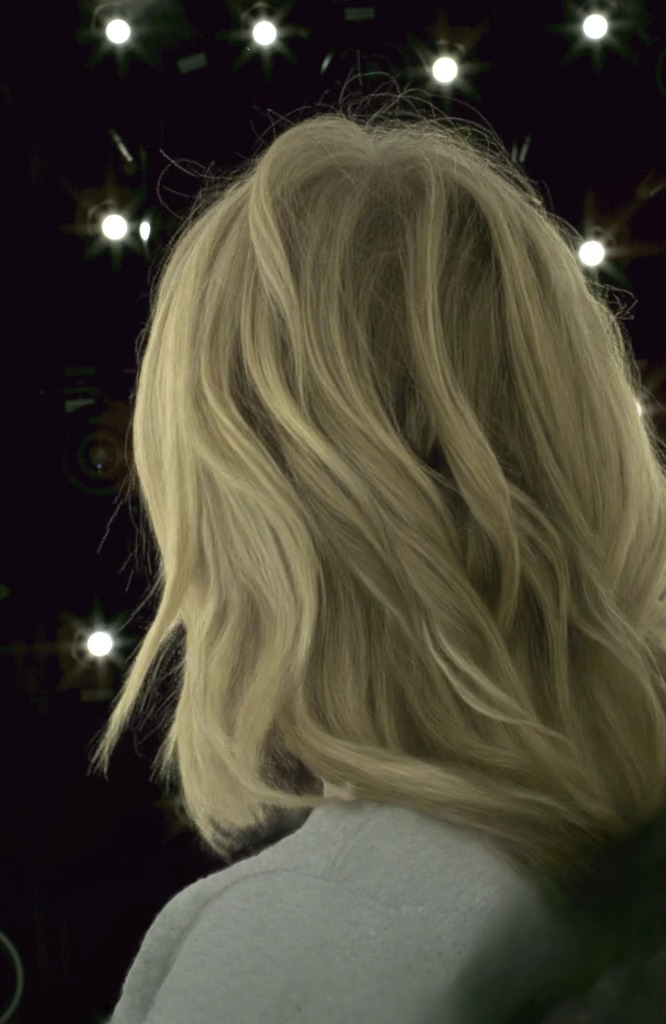}}} &
 {\setlength{\fboxsep}{0pt}\setlength{\fboxrule}{1pt}\fcolorbox{red}{white}{\adjincludegraphics[width=0.12\textwidth, trim={{0.4\width} {0.52\height} {0.16\width} {0.17\height}}, clip]{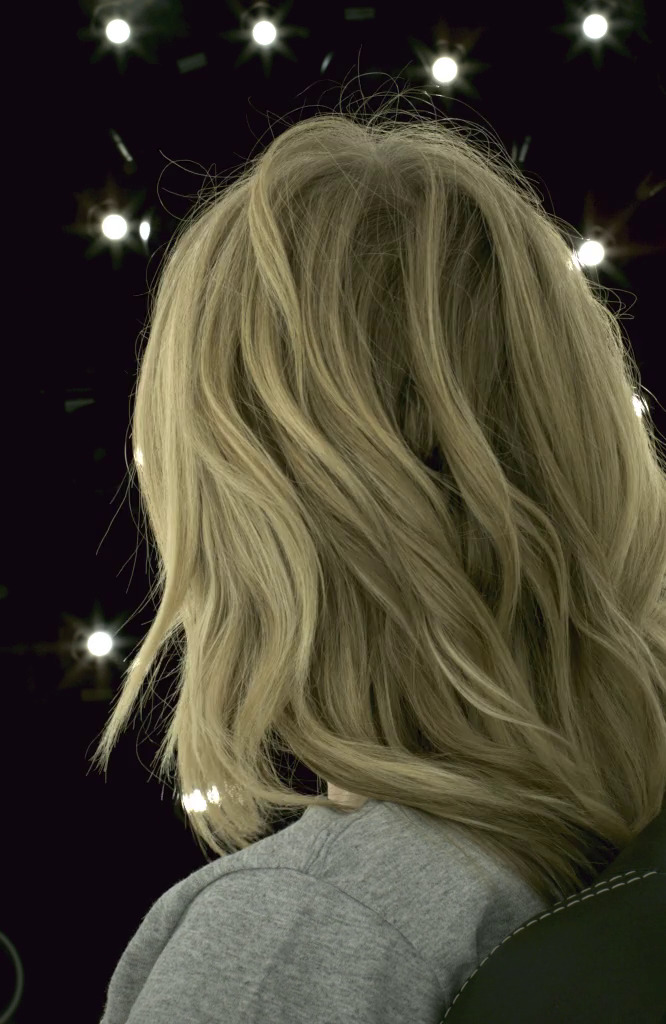}}}
\end{tabular}
\caption{\label{fig1:hair_vgg}\textbf{Ablation on $\mathcal{L}_{VGG}$.} Adding a perceptual loss leads to sharper reconstruction results.
}
\end{figure}

\noindent\textbf{Ablation on $\mathcal{L}_{VGG}$.}
We examine the synergy between $\mathcal{L}_{VGG}$ and the $\ell_1$ loss for improving the rendering quality.
As shown in Tab.~\ref{tab1:vgg}, we find that the perceptual loss has positive effects on the reconstruction performance while the improvements are negated when the weight is too large.
In Fig.~\ref{fig1:hair_vgg}, we compare the rendered images using different $\mathcal{L}_{VGG}$ weights.
The results are more blurry when not using $\mathcal{L}_{VGG}$ and fewer details are reconstructed, such as fly-away strands.

\begin{table}[tb]
\centering
\resizebox{\columnwidth}{!}{
\begin{tabular}{c|c|c|c|c|c|c}
      & vgg=0.0 & vgg=0.1 & vgg=0.3 & vgg=1.0 & vgg=3.0 & vgg=10.0 \\ \hline
MSE   & 42.84   & 42.40   & 39.94   & 40.34   & 40.98   & 42.82    \\ \hline
PSNR  & 32.04   & 32.09   & 32.34   & 32.28   & 32.25   & 32.07    \\ \hline
SSIM  & 0.9544  & 0.9564  & 0.9576  & 0.9558  & 0.9541  & 0.9518   \\ \hline
LPIPS & 0.2021  & 0.1765  & 0.1511  & 0.1299  & 0.1238   & 0.1257   
\end{tabular}
}
\caption{\label{tab1:vgg}\textbf{Ablation on $\mathcal{L}_{VGG}$.} We find using an additional complementary perceptual loss leads to better appearance reconstruction.}
\end{table}

\begin{figure}[tb]
\setlength\tabcolsep{0pt}
\renewcommand{\arraystretch}{0}
\centering
\begin{tabular}{cccc}
 \textbf{\scriptsize w/o flow (frame T)} &
 \textbf{\scriptsize w/o flow (frame T+1)} &
 \textbf{\scriptsize w/ flow (frame T)} & 
 \textbf{\scriptsize w/ flow (frame T+1)} \\
 {\setlength{\fboxsep}{0pt}\setlength{\fboxrule}{1pt}\fcolorbox{blue}{white}{\adjincludegraphics[width=0.12\textwidth, trim={{0.32\width} {0.55\height} {0.24\width} {0.17\height}}, clip]{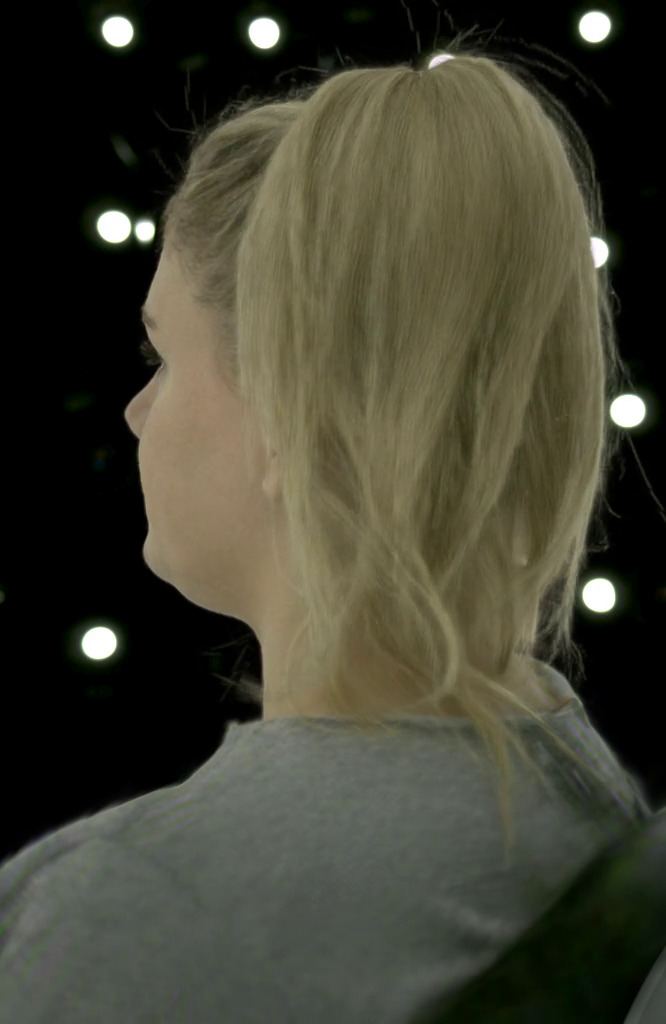}}} &
 {\setlength{\fboxsep}{0pt}\setlength{\fboxrule}{1pt}\fcolorbox{green}{white}{\adjincludegraphics[width=0.12\textwidth, trim={{0.32\width} {0.55\height} {0.24\width} {0.17\height}}, clip]{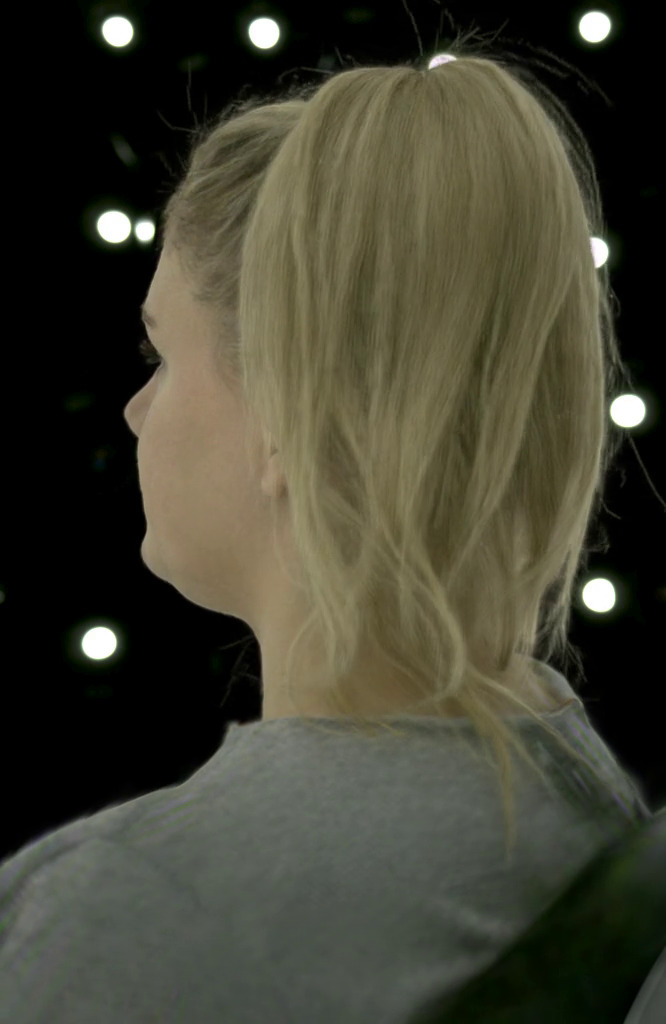}}} &
 {\setlength{\fboxsep}{0pt}\setlength{\fboxrule}{1pt}\fcolorbox{blue}{white}{\adjincludegraphics[width=0.12\textwidth, trim={{0.32\width} {0.55\height} {0.24\width} {0.17\height}}, clip]{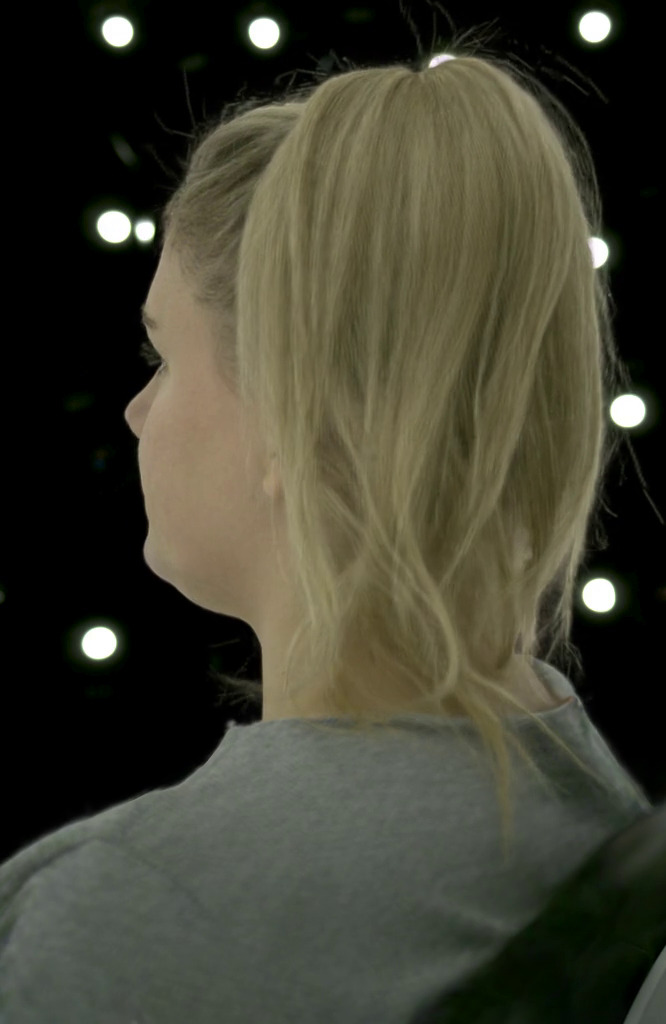}}} &
 {\setlength{\fboxsep}{0pt}\setlength{\fboxrule}{1pt}\fcolorbox{green}{white}{\adjincludegraphics[width=0.12\textwidth, trim={{0.32\width} {0.55\height} {0.24\width} {0.17\height}}, clip]{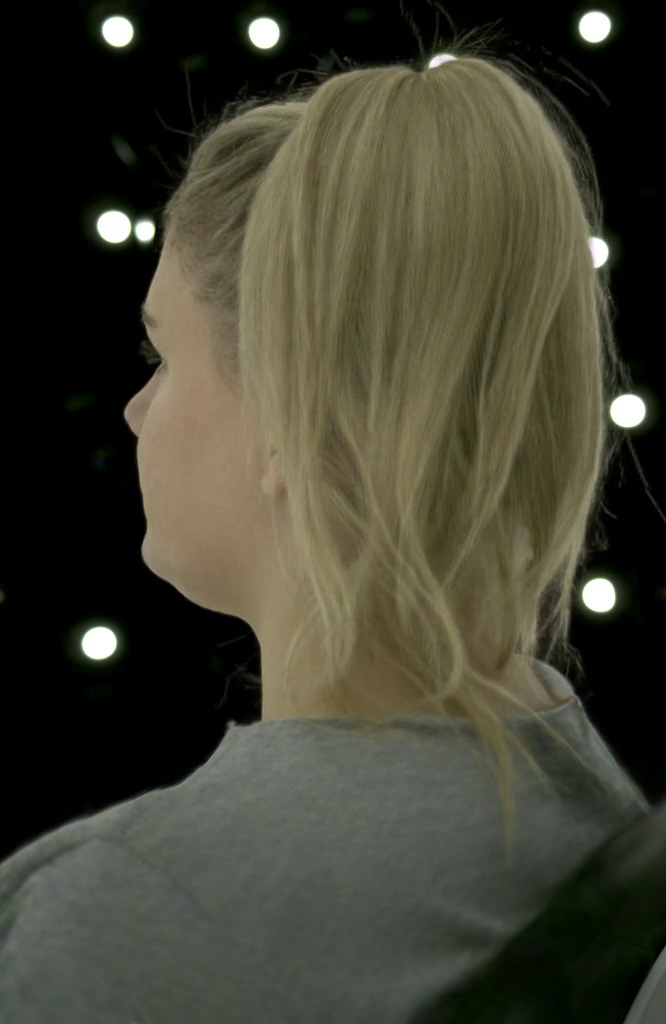}}}
\end{tabular}
\caption{\label{fig1:hair_ptsflow}\textbf{Ablation on Point Flow.} We find that adding point flow to regularize the offsets between temporally adjacent tracked points prevents jittering.}
\end{figure}

\noindent\textbf{Ablation on Point Flow Supervision.} Although with $\mathcal{L}_{cham}$ we can can already optimize a reasonably tracked point cloud $p_t$, we find that point flow can help remove the jittering in appearance.
We show the temporal smoothness enforced by the point flow supervision in Fig.~\ref{fig1:hair_ptsflow}.
Our model learns a more consistent hair texture with less jittering when trained with point flow.
Please see the videos in the supplemental material for a visualization over time.

\subsection{Evaluation of the Dynamic Model}


Lastly, we perform tests of our animation model.
Compared to a per-frame model that takes hair observations as input, the input to our dynamic model and any of its variations is a subset of the head motion trajectory and hair point cloud at the initialization frame.
As a quantitative evaluation, we compare our model with per-frame driven models using either hair observations or head observations as driving signals. 
For qualitative evaluation, we render new hair animations for the bald head motion sequences.

\noindent\textbf{Quantitative Test of the Dynamic Model.}
We evaluate our dynamic model on the test sequences of scripted hair motion capture.
The goal is to test whether our dynamic model generates reasonable novel content rather than only testing how well it reconstructs the test sequence.
To test the performance of our dynamic model, we treat the model driven by per-frame (\textbf{pf}) hair observations as an oracle paradigm to compare with, while our dynamic model does not use any per-frame hair observation as a driving signal.
%
%
In Tab.~\ref{tab2:drive_test}, we compare the rendering quality of our dynamic model with \textbf{pf} models and ablate several designs.
%
%
%
The best performing dynamic model~\textbf{(dyn)} has similar performance with the \textbf{pf} model even without the per-frame hair observation as driving signal.
%
%
%
%
We find that both adding a cosine similarity loss as an additional objective to MSE and adding a cycle consistency loss helps improve the stability of the dynamic model.
Meanwhile, we find adding gravity as an auxiliary input stabilizes our model on slow motions.
This improvement might be related to the fact that during slow motions of the head, the hair motion is primarily driven by gravity.

\begin{table}[tb]
\resizebox{\columnwidth}{!}
{
\begin{tabular}{c|cccc|c}
               & MSE($\downarrow$) & PSNR($\uparrow$) & SSIM($\uparrow$) & LPIPS($\downarrow$) & ChamDis($\downarrow$) \\ \hline
pf w/ hair img & \textbf{37.36} & \textbf{32.94} & \textbf{0.9559} & \textbf{0.1333} & \textbf{10.47} \\ \hline
dyn w/o cos    & 44.96 & 32.26 & 0.9458 & 0.1327 & 25.49 \\
dyn w/o cyc    & 45.22 & 32.23 & 0.9453 & 0.1335 & 26.79 \\
dyn w/o grav   & 40.12 & 32.64 & 0.9504 & 0.1268 & 13.76 \\
dyn            & \textbf{38.49} & \textbf{32.80} & \textbf{0.953}2 & \textbf{0.1211} & \textbf{11.12} \\ \hline
\end{tabular}
}
\caption{\label{tab2:drive_test} \textbf{Ablation of Different Dynamic Models.} We compare different models in terms of rendering quality and tracking accuracy. \iffalse\MZ{highlight best results in bold}\fi}
\end{table}

\begin{figure}[tb]
\setlength\tabcolsep{0pt}
\renewcommand{\arraystretch}{0}
\centering
\begin{tabular}{ccc}
 \textbf{\scriptsize enc vis} &
 \textbf{\scriptsize \textcolor{red}{encprop}} &
 \textbf{\scriptsize \textcolor{TealBlue}{ptsprop}} \\
 \adjincludegraphics[width=0.145\textwidth, trim={0 0 0 {0.15\height}}, clip]{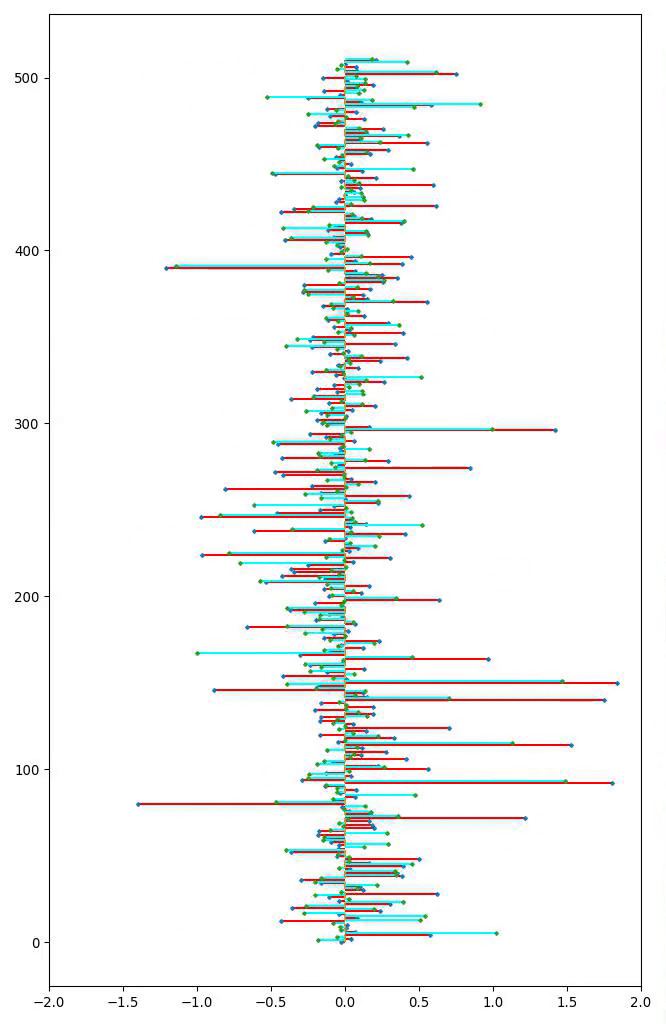} &
 \adjincludegraphics[width=0.145\textwidth, trim={{0.2\width} {0.12\height} 0 {0.2\height}}, clip]{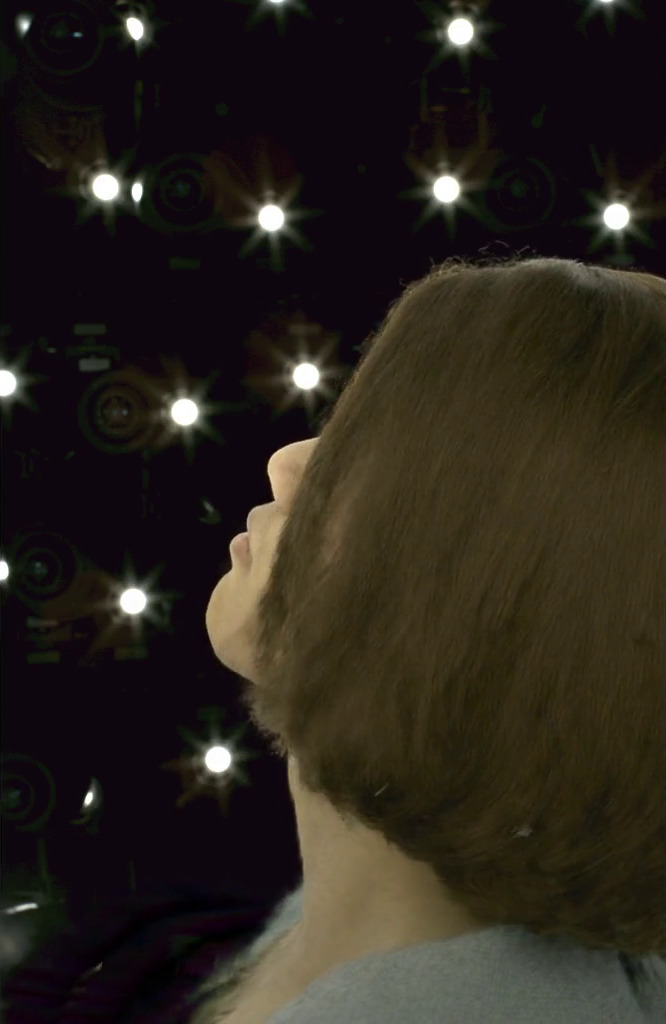} &
 \adjincludegraphics[width=0.145\textwidth, trim={{0.2\width} {0.12\height} 0 {0.2\height}}, clip]{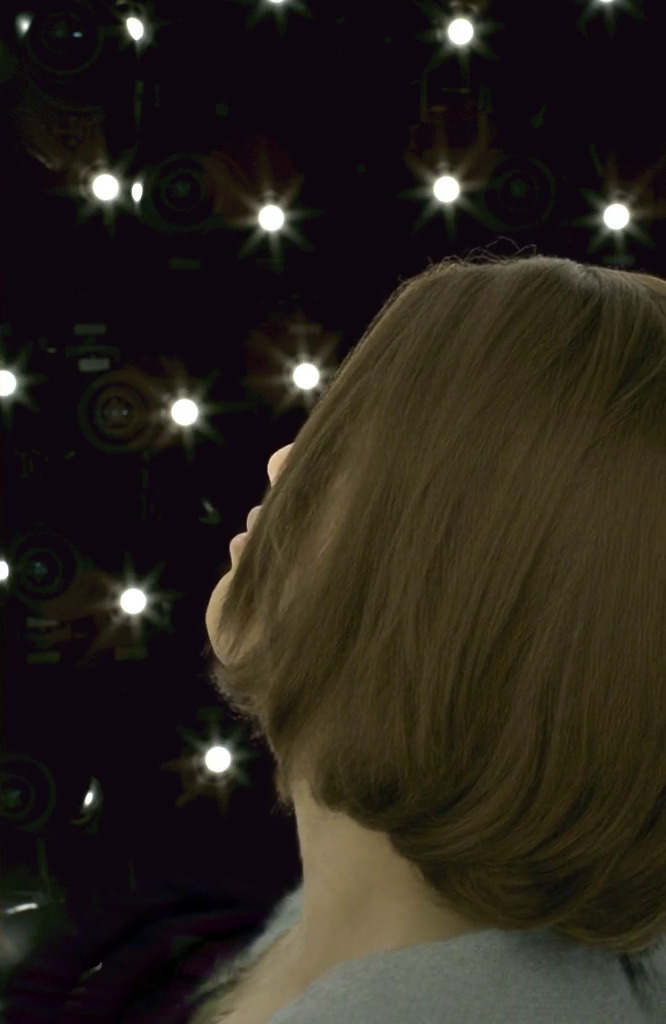} \\
 \adjincludegraphics[width=0.145\textwidth, trim={0 0 0 {0.15\height}}, clip]{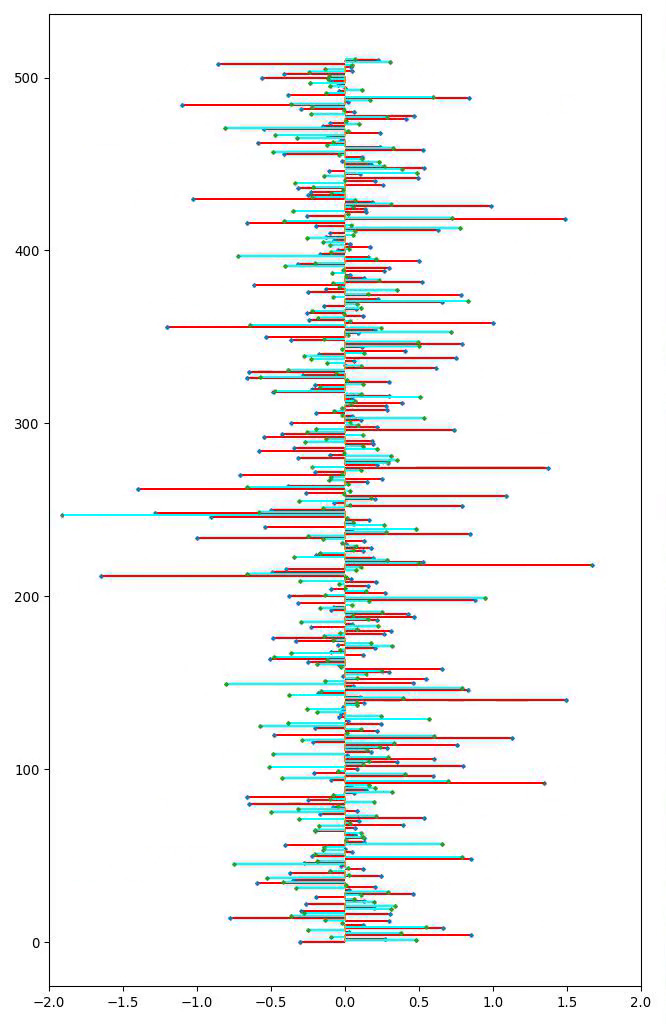} &
 \adjincludegraphics[width=0.145\textwidth, trim={{0.2\width} {0.12\height} 0 {0.2\height}}, clip]{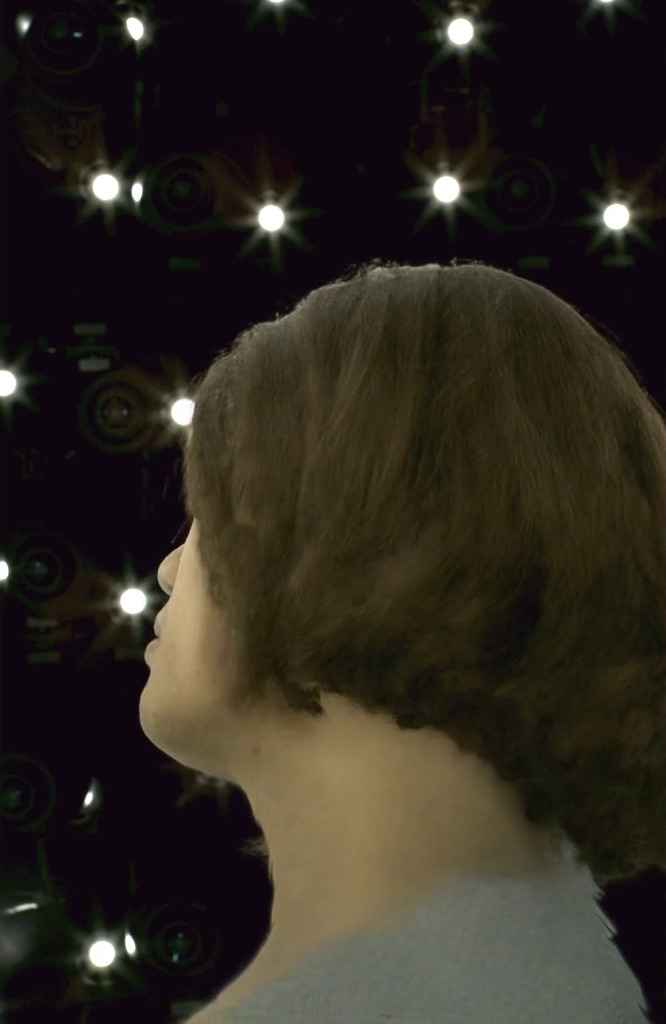} &
 \adjincludegraphics[width=0.145\textwidth, trim={{0.2\width} {0.12\height} 0 {0.2\height}}, clip]{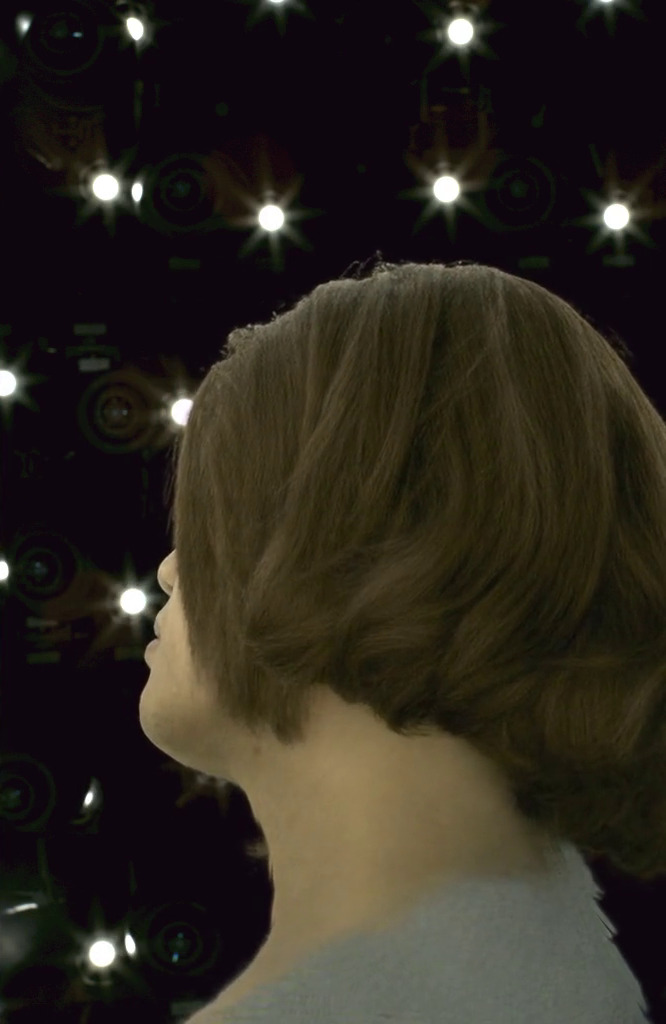}
\end{tabular}
\caption{\label{fig2:encprop}\textbf{\textcolor{red}{encprop} v.s. \textcolor{TealBlue}{ptsprop}.} \textcolor{TealBlue}{ptsprop} generates sharper results with less drifting than \textcolor{red}{encprop}.}
\end{figure}

\noindent\textbf{Ablation for the Point Autoencoder $\mathcal{E}+\mathcal{D}$.}
Here, we analyze how well the point autoencoder acts as a stabilizer for the dynamic model.
We compare two different models in Fig.~\ref{fig2:encprop}: \textcolor{red}{encprop}, that propagates the encoding directly, and \textcolor{TealBlue}{ptsprop} that propagates the regressed hair point cloud and generates the corresponding encoding from the point encoder.
The results are more salient with \textcolor{TealBlue}{ptsprop}.
The improvement of the \textcolor{TealBlue}{ptsprop} over the \textcolor{red}{encprop} is partially because the mapping from point cloud to encoding is an injective mapping and the point encoder serves as a noise canceller in the encoding space.
To further study this behavior, we perform a cycle test on the point encoder, where we add noise $n$ to a certain encoding {\color{red} $z$} and get a noisy version of the encoding {\color{cyan}$\hat{z}$}$=${\color{red}$z$}$+n$ and its corresponding noisy point cloud $\hat{p}$.
Then, we predict a cycled encoding {\color{blue}$\bar{z}$}$=\mathcal{E}(\mathcal{D}(${\color{cyan}$\hat{z}$}$))$.
We compare {\color{red}$z$}, {\color{cyan}$\hat{z}$} and {\color{blue}$\bar{z}$} in Fig.~\ref{fig2:encprop}.
{\color{red}$z$} and {\color{blue}$\bar{z}$} are consistently close while {\color{cyan}$\hat{z}$} jitters.
This result suggests that the remapping of {\color{cyan}$\hat{z}$} using $\mathcal{E}$ and $\mathcal{D}$ counteracts the noise $n$.

\begin{figure}[tb]
\setlength\tabcolsep{0pt}
\renewcommand{\arraystretch}{0}
\centering
\begin{tabular}{ccc}
 \textbf{\scriptsize sample 1} &
 \textbf{\scriptsize sample 2} &
 \textbf{\scriptsize sample 3} \\
 \adjincludegraphics[width=0.145\textwidth, trim={{0.3\width} {0.08\height} {0.2\width} {0.06\height}}, clip]{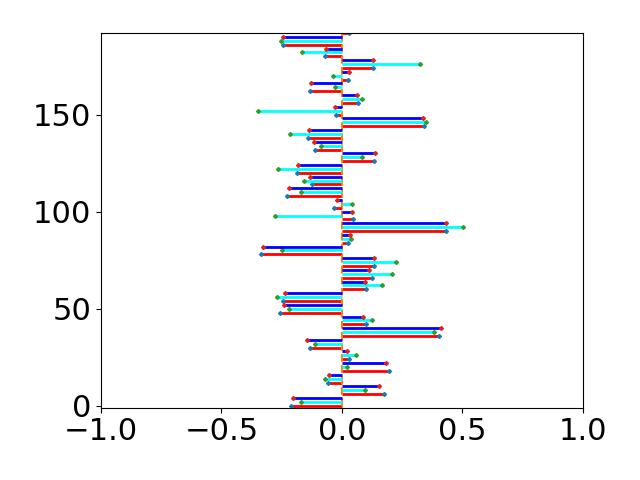} &
 \adjincludegraphics[width=0.145\textwidth, trim={{0.3\width} {0.08\height} {0.2\width} {0.06\height}}, clip]{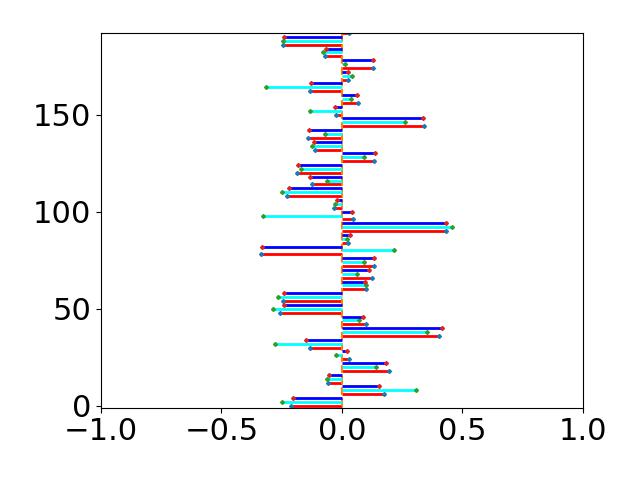} &
 \adjincludegraphics[width=0.145\textwidth, trim={{0.3\width} {0.08\height} {0.2\width} {0.06\height}}, clip]{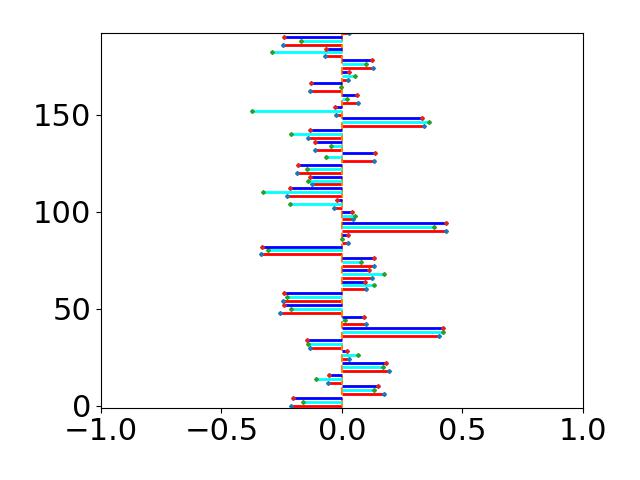}
\end{tabular}
\caption{\label{fig2:enc_sample}\textbf{Point Encoder $\mathcal{E}$ as a Stabilizer.} We sample several {\color{cyan}$\hat{z}$} and corresponding {\color{blue}$\bar{z}$}$=\mathcal{E}(\mathcal{D}(${\color{cyan}$\hat{z}$}$))$ with a fixed {\color{red}$z$} and visualize part of them above. As we can see, {\color{blue}$\bar{z}$} stays similar to {\color{red}$z$} while {\color{cyan}$\hat{z}$} jitters.}
\end{figure}

\noindent\textbf{Animation on Bald Head Sequences.}
We show animation results driven by head motions on in-the-wild phone video captures in Fig~\ref{fig0:anim_iphone}. 
We find our model generates reasonable motions of hair under natural head motions like swinging or nodding.
Please refer to the supplemental materials and videos for visualizations over time.

%% file: discuss.tex
\section{Discussion}
We present a two-stage data-driven pipeline for volumetric hair capture and animation.
The core of our method is a 3D volumetric autoencoder that we find useful for both, automatic hair state acquisition and stable hair dynamic generation.
The first stage of our pipeline simultaneously performs hair tracking, geometry and appearance reconstruction in a self-supervised manner via an autoencoder-as-a-tracker strategy.
The second stage leverages the hair states acquired from the first stage and creates a recurrent hair dynamics model that is robust to moderate drift with the autoencoder as a denoiser. 
We empirically show that our method performs stable tracking of hair on long and segmented video captures while preserving high fidelity for hair appearance. 
Our model also supports generating new animations in both, lab- and in-the-wild conditions and does not rely on hair observations.

\noindent\textbf{Limitations.}
Like many other data-driven methods, our method requires a large amount of diverse sstraining data and might fail with data that is far from the training distribution.
Our model is currently not relightable and can not animate new hairstyles.
One future direction is to separately model appearance and lighting, which can be learnt from varied lighting captures. 
Another interesting direction is to learn a morphable hair model for new hairstyle adaptation.

%% file: appendix.tex
\section{Appendix}

\subsection{Network Architecture}

Here we provide details about how we design our neural networks and further information about training.

\noindent\textbf{Encoder.} As training a point cloud encoder solely is extremely unstable, we first train an image encoder and use it as a teacher model to train the point cloud encoder. In practice, we train two encoders for our hair branch together. Here we first illustrate the structure of both encoders. We will go back to how we train them and use them later. One of the encoders is an image encoder which is a convolutional neural network (CNN) that takes multiple view images as input. We denote the image encoder as $\mathcal{E}_{img}$ The other one is $\mathcal{E}$ which is a PointNet encoder that takes either an unordered hair point cloud $p_t$ or a tracked hair point cloud $q_t$ as input. Positional encoding~\cite{mildenhall2020nerf} is applied to the raw point cloud coordinate before it is used as the input to the network. We find this is very effective to help the network capturing high frequency details. In practice, we use frequencies of $x^2$ where $x$ ranges from $1$ to $7$. We show the detailed architecture of $\mathcal{E}_{img}$ in Tab~\ref{tab:enc_img}. The architecture of the point cloud encoder $\mathcal{E}$ is shown in Tab~\ref{tab:enc_pt}. Both of the two encoders $\mathcal{E}$ and $\mathcal{E}_{img}$ can produce a latent vector in size of $256$, which are supposed to describe the same content. Their output will be passed to $\mathcal{E}_{\mu}$ and $\mathcal{E}_{\sigma}$ which are two linear layers that produce $\bm\mu$ and $\bm\sigma$ of $\bm{z_t}$ respectively.

\begin{table}[!ht]
\centering
\begin{tabular}{|c|c|c|}
\hline
   & \multicolumn{2}{c|}{Encoder $\mathcal{E}_{img}$}              \\ \hline
1  & \multicolumn{2}{c|}{\ttt{Conv2d}(3, 64)}        \\ \hline
2  & \multicolumn{2}{c|}{\ttt{Conv2d}(64, 64)}       \\ \hline
3  & \multicolumn{2}{c|}{\ttt{Conv2d}(64, 128)}      \\ \hline
4  & \multicolumn{2}{c|}{\ttt{Conv2d}(128, 128)}     \\ \hline
5  & \multicolumn{2}{c|}{\ttt{Conv2d}(128, 256)}     \\ \hline
6  & \multicolumn{2}{c|}{\ttt{Conv2d}(256, 256)}     \\ \hline
7  & \multicolumn{2}{c|}{\ttt{Conv2d}(256, 256)}     \\ \hline
8  & \multicolumn{2}{c|}{\ttt{Flatten}()}            \\ \hline
9  & \multicolumn{2}{c|}{\ttt{Linear}(256$\times n_{inpimg}\times$15, 256)} \\ \hline
\end{tabular}
\vspace{0.1cm}
\caption{\label{tab:enc_img}\textbf{Encoder $\mathcal{E}_{img}$ architecture}. Each \ttt{Conv2d} layer in the encoder has a kernel size of $3$, stride of $1$ and padding of $1$. Weight normalization~\cite{salimans2016weight} and untied bias are applied. After each layer, except for the last two parallel fully-connected layers, a Leaky ReLU~\cite{maas2013leakyrelu} activation with a negative slope of $0.2$ is applied. Then a downsample layer with a stride of $2$ is applied after every conv2d layer. The first linear layer takes the concatenation of all towers from different image views as input. $n_{inpimg}$ stands for much many views we take.
}
\end{table}

\begin{table}[!ht]
\centering
\begin{tabular}{|c|c|c|}
\hline
   & \multicolumn{2}{c|}{Encoder $\mathcal{E}$}              \\ \hline
1  & \multicolumn{2}{c|}{\ttt{Conv2d}(3, 128)}        \\ \hline
2  & \multicolumn{2}{c|}{\ttt{Conv2d}(128, 256)}       \\ \hline
3  & \multicolumn{2}{c|}{\ttt{Conv2d}(256, 256)}      \\ \hline
4  & \multicolumn{2}{c|}{\ttt{Conv2d}(256, 256)}     \\ \hline
5  & \multicolumn{2}{c|}{\ttt{Conv2d}(256, 512)}     \\ \hline
6  & \multicolumn{2}{c|}{\ttt{Conv2d}(512, 512)}     \\ \hline
7  & \multicolumn{2}{c|}{\ttt{Conv2d}(512, 512)}     \\ \hline
8  & \multicolumn{2}{c|}{\ttt{Conv2d}(512, 1024)}     \\ \hline
8  & \multicolumn{2}{c|}{\ttt{MAM pooling}()}            \\ \hline
9  & \multicolumn{2}{c|}{\ttt{Linear}(1024$\times$3, 512)} \\ \hline
10 & \multicolumn{2}{c|}{\ttt{Linear}(512, 256)}   \\ \hline
11 & \multicolumn{2}{c|}{\ttt{Linear}(256, 256)}   \\ \hline
\end{tabular}
\vspace{0.1cm}
\caption{\label{tab:enc_pt}\textbf{Encoder $\mathcal{E}$ architecture}. We use a $\mathcal{E}$ structure similar to PointNet~\cite{qi2017pointnet}. All \ttt{Conv2d} uses a kernel of $1$ and stride of $1$, which serves as a shared MLP. We only use \ttt{Conv2d} for simpler implementation. After each \ttt{Conv2d} layer, a Leaky ReLU~\cite{maas2013leakyrelu} activation with a negative slope of $0.2$ is applied. Then we use a MAM pool layer to aggregate features from all points. MAM stands for min, avarage and max pooling, which concatenates the results of min, average and max pooling into one. Then, two linear layers are applied to the output of MAM pooling and generate a 256 latent vector.
}
\end{table}

\noindent\textbf{Point Decoder.} We use a 3-layer MLP as the point decoder $\mathcal{D}$, which takes a 1d latent code $\bm{z_t}$ as input and outputs the coordinate of the corresponding tracked point cloud $q_t$. We show the architecture of $\mathcal{D}$ in Tab~\ref{tab:ptsbranch}.

\begin{table}[!ht]
\centering
\begin{tabular}{|c|c|c|}
\hline
   & \multicolumn{2}{c|}{Decoder $\mathcal{D}$}              \\ \hline
1  & \multicolumn{2}{c|}{\ttt{Linear}(256, 256)} \\ \hline
2 & \multicolumn{2}{c|}{\ttt{Linear}(256, 256)}   \\ \hline
2 & \multicolumn{2}{c|}{\ttt{Linear}(256, 4096$\times$3)}   \\ \hline
\end{tabular}
\vspace{0.1cm}
\caption{\label{tab:ptsbranch}\textbf{Decoder $\mathcal{D}$ architecture}. We use an MLP with three \ttt{Linear} layers as the decoder $\mathcal{D}$. After each layer except the last layer, a Leaky ReLU~\cite{maas2013leakyrelu} activation with a negative slope of 0.2 is applied.
}
\end{table}

\noindent\textbf{Volume Decoder.} The volumetric model is a stack of 2D deconv layers. We align the x-axis and y-axis of each volume and put them onto a 2D imaginary UV-space. Then we convolve on them to regress the z-axis content for each of the x,y position. We show the architecture of the volume decoder in Tab~\ref{tab:vol_dec}. In our setting, we have two seperate volume decoder for both RGB volume and alpha volume.

\begin{table}[]
\centering
\begin{tabular}{|c|cc|}
\hline
  & \multicolumn{2}{c|}{Volume Decoder}                                                                                       \\ \hline
  & \multicolumn{1}{c|}{global encoding $\bm{z}_t$} & \multirow{2}{*}{\begin{tabular}[c]{@{}c@{}}per-point \\ hair feature\end{tabular}} \\ \cline{1-2}
  & \multicolumn{1}{c|}{repeat}          &                                                                                    \\ \hline
  & \multicolumn{2}{c|}{concat}                                                                                               \\ \hline
1 & \multicolumn{2}{c|}{$\ttt{Linear}$(320, 512)}                                                                             \\ \hline
2 & \multicolumn{2}{c|}{$\ttt{deconv2d}$(512, 256)}                                                                           \\ \hline
3 & \multicolumn{2}{c|}{$\ttt{conv2d}$(256, 256)}                                                                             \\ \hline
4 & \multicolumn{2}{c|}{$\ttt{deconv2d}$(256, 256)}                                                                           \\ \hline
5 & \multicolumn{2}{c|}{$\ttt{conv2d}$(256, 256)}                                                                             \\ \hline
6 & \multicolumn{2}{c|}{$\ttt{deconv2d}$(256, 128)}                                                                           \\ \hline
7 & \multicolumn{2}{c|}{$\ttt{conv2d}$(128, 128)}                                                                             \\ \hline
8 & \multicolumn{2}{c|}{$\ttt{deconv2d}$(128, 16$\times$ch)}                                                                  \\ \hline
\end{tabular}
\caption{\label{tab:vol_dec}\textbf{Architecture of the Volume Decoder}. We first repeat the global encoding $\bm{z}_t$ into the shape of the per-point hair feature. The per-point hair feature is a tensor that is shared across all time frames. We then concatenate those two into one. Each layer except for the last one is followed by a Leaky ReLU layer with a negative slope of $0.2$. Each $\ttt{deconv2d}$ layer has a filter size of 4, stride size of 2 and padding size of 1. Each $\ttt{conv2d}$ layer has a filter size of 3, stride size of 1 and padding size of 1. ch stands for the channel size of the output. It is set to 3 if it is an rgb decoder and 1 for a alpha decoder.}
\end{table}

\noindent\textbf{Dynamic Model.}
We use three different inputs to the dynamic model $\mathtt{T^2M}$, namely the hair encoding $\bm{z}_{t-1}$ at the previous frame $t-1$, the head velocity $\bm{h}_{t-1}$ and $\bm{h}_{t-2}$ from the previous two frames $t-1$ and $t-2$, and the head relative gravity direction $\bm{g}_t$ at the current frame $t$.
We first encode $\{\bm{h}_{t-1}, \bm{h}_{t-2}\}$ and $\bm{g}_t$ into two 1d vectors with 128 dimensions respectively. Then, we concatenate them together with encoding $\bm{z}_{t-1}$ as the input to another MLP to regress the next possible hair state encoding $\bm{z}_{t}$.
As in Tab.~\ref{tab:t2m}, we show the flow of $\mathtt{T^2M}$. For the head velocity branch, we first extract the per-vertex velocity $\bm{h}_{t-1}=\bm{x}_{t}-\bm{x}_{t-1}$ where $\bm{x}_{t}$ is the coordiante of the tracked head mesh at frame $t$.
To be noted, here the $\bm{h}_{t-1}$ contains only the information of the rigid head motion but not any other non-rigid motion like expression change.
This representation of head motion is redundant theoretically, but we find it helps our network to converge better comparing to just using the pure 6-DoF head rotation and translation.
We then reshape it and use it as the input to a two layer MLP to extract a 1d encoding of size 128. For the gravity branch, we first encode the gravity direction $\bm{g}_t$ with cosine encoding~\cite{mildenhall2020nerf}.
The output of the dynamic model is the mean $\mu_{t+1}$ and standard deviation $\sigma_{t+1}$ of the predicted hair state $\bm{z}_{t+1}$. 

\begin{table}[]
\resizebox{\columnwidth}{!}{

\begin{tabular}{|c|cccc|}
\hline
  & \multicolumn{4}{c|}{Temporal Transfer Module ($\mathtt{T^2M}$)}                                                                                     \\ \hline
1 & \multicolumn{2}{c|}{head velocity $\{\bm{h}_{t-1}, \bm{h}_{t-2}\}$}                   & \multicolumn{1}{c|}{head relative gravity $\bm{g}_{t}$}            & hair state $\bm{z}_{t-1}$      \\ \hline
2 & \multicolumn{2}{c|}{$\ttt{Linear}$(7306$\times$3, 256)} & \multicolumn{1}{c|}{\multirow{2}{*}{cosine encoding}} & \multirow{2}{*}{} \\ \cline{1-3}
3 & \multicolumn{2}{c|}{$\ttt{Linear}$(256, 128)}           & \multicolumn{1}{c|}{}                                 &                   \\ \hline
4 & \multicolumn{4}{c|}{$\ttt{Linear}$(539, 256)}                                                                                       \\ \hline
5 & \multicolumn{4}{c|}{$\ttt{Linear}$(256, 256)}                                                                                       \\ \hline
6 & \multicolumn{4}{c|}{$\ttt{Linear}$(256, 256)}                                                                                       \\ \hline
7 & \multicolumn{2}{|c|}{\ttt{Linear}(256, 256)}     & \multicolumn{2}{|c|}{\ttt{Linear}(256, 256)}   \\ \hline
\end{tabular}
}
\caption{\label{tab:t2m}\textbf{Temporal Transfer Module ($\mathtt{T^2M}$)}. We first encode the head velocity $\{\bm{h}_{t-1}, \bm{h}_{t-2}\}$  and head relative gravity $\bm{g}_{t}$ into 1d vectors, with a 2-layer MLP and cosine encoding respectively. Then we concatenate hair state $\bm{z}_{t-1}$ with those vectors to serve as the input to another MLP. The last two layers will be regressing the mean $\mu_{t+1}$ and standard deviation $\sigma_{t+1}$ of the predicted hair state $\bm{z}_{t+1}$. All $\ttt{Linear}$ expect for the last two are followed by a Leaky ReLu activation with a negative slope of $0.2$.
}
\end{table}

\subsection{Training details}

\noindent\textbf{Dataset and Capture Systems.} 
Following the setting in HVH~\cite{wang2021hvh}, we also captured several video sequences with scripted hair motion performed under different hair styles for animation tests.
During the capture, we ask the participants to put on different kind of hair wigs and perform a varity of head motions like nodding, swinging and tilting for multiple times under both slow and fast speed.
To collect a demonstration set for animation, we also ask the participants to put on a hair net (bare head) and perform the same set of motions as when they are wearing a hair wig.

\noindent\textbf{Hair Point Flow Estimation.} There are three steps for computing the hair point flow, namely per-point feature descriptor extraction, feature matching and flow filtering. In the first step, we compute a per-point feature descriptor based on the distribution of each point's local neighboring. In the second step, we match the points from two adjacent time steps based on the similarity between their feature descriptor. In the last step, we filter out outlier flows that are abnormal.

\begin{figure}
    \centering
    \adjincludegraphics[width=0.5\columnwidth]{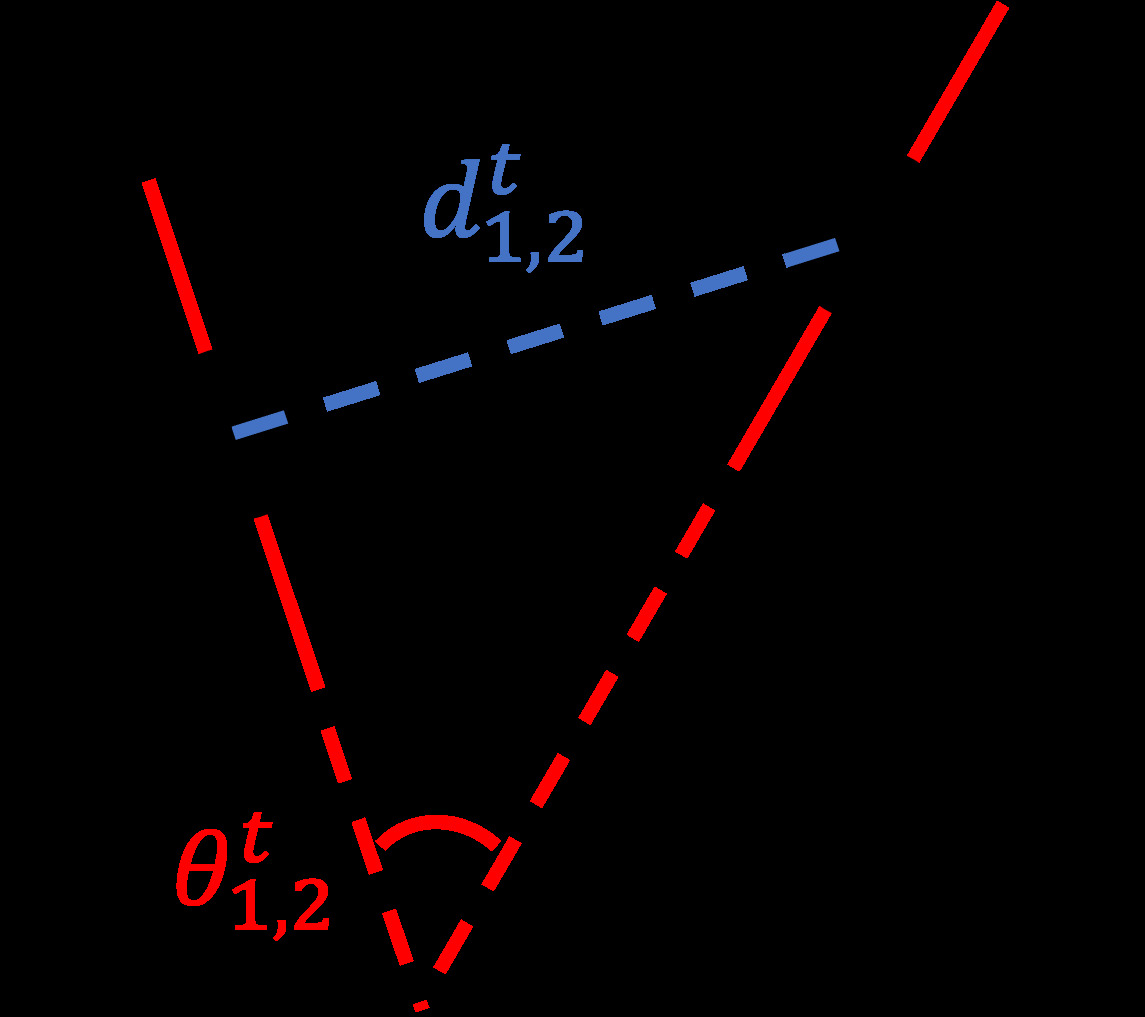}
    \caption{\label{fig_supp:lfh}}
\end{figure}

To compute the point feature descriptor, we construct Line Feature Histograms~(LFH) inspired by Point Feature Histograms~(PFH)~\cite{rusu2008persistent}. 
The LFH is a historgram of a 4-tuple that describes the spatial relationship between a certain point $\bm{p}_1^t$ and its neighboring poit $\bm{p}_2^t$.
As shown in Fig.~\ref{fig_supp:lfh}, we visualize two points $\bm{p}_1^t\in \mathbb{R}^3$ and $\bm{p}_2^t\in\mathbb{R}^3$ from the same time step $t$. Given $\bm{p}_1^t$ and $\bm{p}_2^t$, we define the following four properties that describe their spatial relationship. The first one is the relative position of $\bm{p}_2^t$ with respect to $\bm{p}_1^t$, which is $\bm{d}_{1,2}^t = \bm{p}_2^t - \bm{p}_1^t$. Then we could compute the relative distance as $||\bm{d}_{1,2}^t||_2\in\mathbb{R}$. The second term is the angle $\theta_{1,2}^t$ between $\bm{dir}(\bm{p}_1^t)$ and  $\bm{dir}(\bm{p}_2^t)$, where $\bm{dir}(\bm{x})$ is the line direction of $\bm{x}$ from ~\cite{nam2019lmvs}. The last two terms are the angles $\alpha_{1,2}^t$ and $\beta_{1,2}^t$ between $(\bm{dir}(\bm{p}_1^t), \bm{d}_{1,2}^t)$ and $(\bm{dir}(\bm{p}_2^t), \bm{d}_{1,2}^t)$ respectively. For all 
intersections, we take the acute angle, which means $\theta_{1,2}^t, \alpha_{1,2}^t$ and $\beta_{1,2}^t$ are in $[0, \pi/2]$. Thus, the 4-tuple we used to create $LFH(\bm{p}_1^t)$ is $(||\bm{d}_{1,2}^t||_2, \theta_{1,2}^t, \alpha_{1,2}^t, \beta_{1,2}^t)$ and we normalize the histogram by its l2 norm. The designed LFH has three good properties. As we use the normalized feature, it is density invariant. Since $\theta_{1,2}^t, \alpha_{1,2}^t$ and $\beta_{1,2}^t$ are always acute angles, the feature is also rotation and flip invariant, where if we flip or rotate $\bm{dir}(\bm{p}_1^t)$ the histogram are still the same. This design helps us getting a more robust feature descriptor for matching. We set the resolution for each entries of the 4-tuple to be 4 and it results in a descriptor in size of 256.

In the second step, we compute the correspondence between points from adjacent time frames $t$ and $t+t_\delta$ where $t_\delta\in\{-1, 1\}$. We use the method from Rusu \textit{et al.}~\cite{rusu2008aligning} to compute the correspondence between two point clouds from $t$ and $t+t_\delta$. To further validate the flow we get, we use several heuristic to filter out some obvious outliers. We first discard all the flows that have a large magnitude. As the flow is computed between two adjacent frames, it is supposed to be not abrupt. The second heuristic we use to filter the outliers is cycle consistency, where we compute the flow both forward and backward to see if we can map back to the origin. If the mapped back point departs too far away from the origin, we discard their flow. 

\noindent\textbf{Training of Encoder.} As mentioned before, we train two encoders $\mathcal{E}$ and $\mathcal{E}_{img}$ together. In practice, we find that directly training $\mathcal{E}$ is not very stable and might lead to not able to converge. Thus, we learn the two encoders in a teach-student manor, where we use $\mathcal{E}_{img}$ as a teach model to train $\mathcal{E}$. We denote $\bm{x}_{img, t}$ as the output of $\mathcal{E}_{img}$ and $\bm{x}_{pt, t}$ as the output of $\mathcal{E}$. Then, we formulate the following MSE loss to enforce the $\mathcal{E}$ to output similarly to $\mathcal{E}_{img}$:

\begin{align*}
    \mathcal{L}_{ts} = ||\bm{x}_{img, t} - \bm{x}_{pt, t}||_2,
\end{align*}

\noindent where we restraint the gradient from $\mathcal{L}_{ts}$ from back-propagating to $\mathcal{E}_{img}$ while training.

\subsection{Ablation on Different Encoders}

We show quantitative evaluations on rendering quality of different encoders on both SEEN and UNSEEN sequences in Tab.~\ref{tab:abl_enc}. As we can see, our $\mathcal{E}$ performs similarly to the $\mathcal{E}_{img}$ on the novel views of the SEEN sequence. This is as expected due to the nation of teach-student model and we train our model on the SEEN sequence with the training views. On the UNSEEN sequence, we find our $\mathcal{E}$ performs better than $\mathcal{E}_{img}$. The reason could be that there are smaller domain gap between the point clouds from SEEN sequence and UNSEEN sequence while the multi-view images might be varying a lot due to the head motion. And the CNN is not good for handling such changes due to the head motion while point encoder can process point clouds with better structure awareness.

\begin{table}[]
\resizebox{\columnwidth}{!}{
\begin{tabular}{|c|c|c|c|c|}
\hline
                       & MSE$\downarrow$   & PSNR$\uparrow$  & SSIM$\uparrow$   & LPIPS$\downarrow$   \\ \hline
$\mathcal{E}$ on SEEN & 29.48 & 34.05 & 0.9657 & 0.1109 \\ \hline
$\mathcal{E}_{img}$ on SEEN   & 29.44 & 34.05 & 0.9657 & 0.1109 \\ \hline
$\mathcal{E}$ on UNSEEN  & 34.97 & 33.21 & 0.9587 & 0.1209 \\ \hline
$\mathcal{E}_{img}$ on UNSEEN    & 37.36 & 32.94 & 0.9559 & 0.1333 \\ \hline
\end{tabular}
}
\caption{\label{tab:abl_enc}\textbf{Metrics on Novel Views.} We show quantitative results of different encoders under both SEEN and UNSEEN sequence of the same hair styles.}
\end{table}

\subsection{Ablation on Different Designs of the Dynamic Model}

\begin{figure}[htb]
    \centering
    \includegraphics[width=\columnwidth]{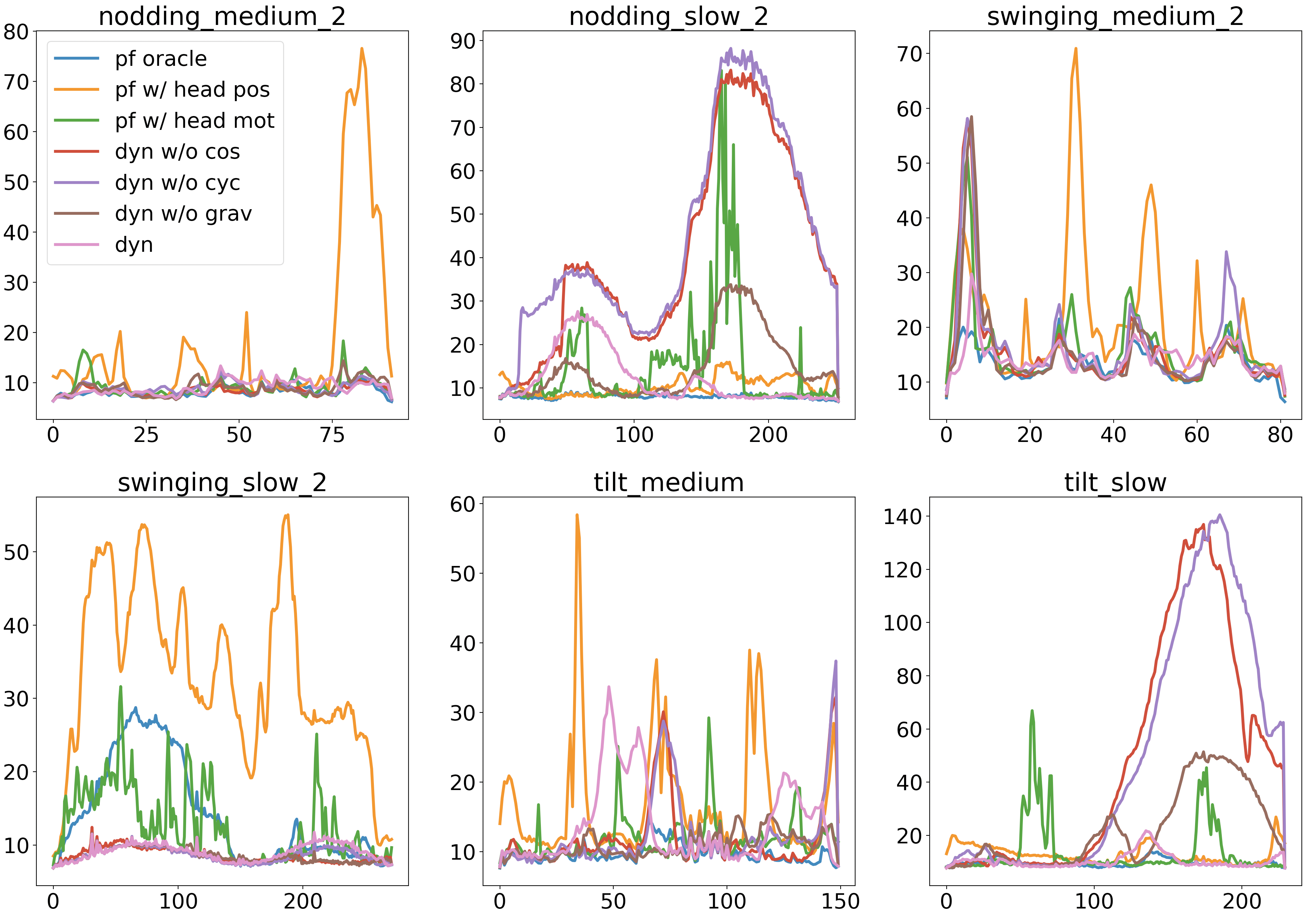}
    \caption{\label{fig_app:cham_plot}\textbf{ChamDist v.s. time.} We plot Chamfer distance v.s. time of different dynamic models to show drifting.}
\end{figure}

We show the comparisons of different dynamic models and per-frame driven models in Tab.~\ref{tab_app:drive_test}.
We find that our model offers a significant improvement over the per-frame driven model that takes head pose or motion as input.
This is, because innately the hair motion is not only determined by the head pose or the previous history of head pose but also the initial status of the hair.
In Fig.~\ref{fig_app:cham_plot}, we visualize how each model drifts by plotting the Chamfer distance between the regressed point cloud and the ground truth point cloud.
We find that adding head relative gravity direction can improve the model performance on slow motions.

\begin{table}[h!tb]
\resizebox{\columnwidth}{!}
{
\begin{tabular}{c|cccc|c}
               & MSE($\downarrow$) & PSNR($\uparrow$) & SSIM($\uparrow$) & LPIPS($\downarrow$) & ChamDis($\downarrow$) \\ \hline
pf w/ hair img & 37.36 & 32.94 & 0.9559 & 0.1333 & 10.47 \\
pf w/ hair pts & 34.97 & 33.21 & 0.9587 & 0.1209 & 10.46 \\ \hline
pf w/ head pos & 47.43 & 32.01 & 0.9458 & 0.1522 & 18.94 \\
pf w/ head mot & 40.25 & 32.64 & 0.9508 & 0.1333 & 13.31 \\ \hline
dyn w/o cos    & 44.96 & 32.26 & 0.9458 & 0.1327 & 25.49 \\
dyn w/o cyc    & 45.22 & 32.23 & 0.9453 & 0.1335 & 26.79 \\
dyn w/o grav   & 40.12 & 32.64 & 0.9504 & 0.1268 & 13.76 \\
dyn            & 38.49 & 32.80 & 0.9532 & 0.1211 & 11.12 \\ \hline
\end{tabular}
}
\caption{\label{tab_app:drive_test} \textbf{Ablation of Different Dynamic Models.}}
\end{table}

\subsection{Effect of the Initialization}
\begin{figure}[h!tb]
\setlength\tabcolsep{0pt}
\renewcommand{\arraystretch}{0}
\centering
\begin{tabular}{ccccc}
 & 
 \multicolumn{2}{c}{\textbf{time 0}} & 
 \multicolumn{2}{c}{\textbf{time n}} \\
 & 
 \textbf{\scriptsize hs1 init.} &
 \textbf{\scriptsize hs2 init.} &
 \textbf{\scriptsize hs1 init.} &
 \textbf{\scriptsize hs2 init.} \\
 \parbox{0.04\textwidth}{\centering hs1\\\centering mod.} & 
 {\setlength{\fboxsep}{0pt}\setlength{\fboxrule}{1pt}\fcolorbox{green}{white}{\adjincludegraphics[width=0.11\textwidth, trim={{0.3\width} {0.2\height} 0 {0.2\height}}, clip, valign=m]{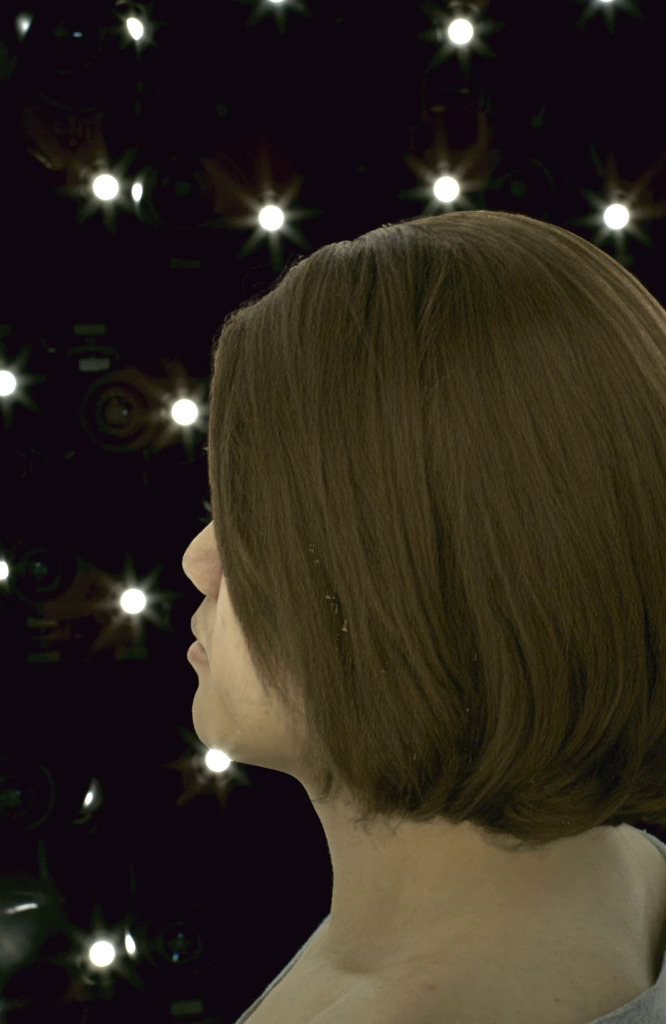}}} &
 {\setlength{\fboxsep}{0pt}\setlength{\fboxrule}{1pt}\fcolorbox{orange}{white}{\adjincludegraphics[width=0.11\textwidth, trim={{0.3\width} {0.2\height} 0 {0.2\height}}, clip, valign=m]{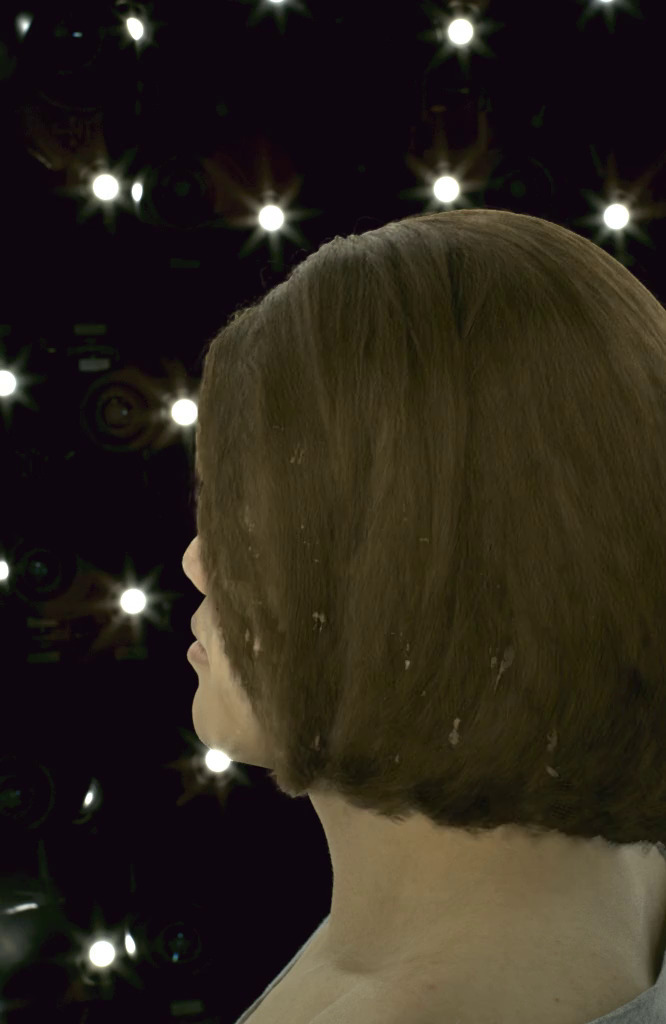}}} &
 {\setlength{\fboxsep}{0pt}\setlength{\fboxrule}{1pt}\fcolorbox{green}{white}{\adjincludegraphics[width=0.11\textwidth, trim={{0.3\width} {0.2\height} 0 {0.2\height}}, clip, valign=m]{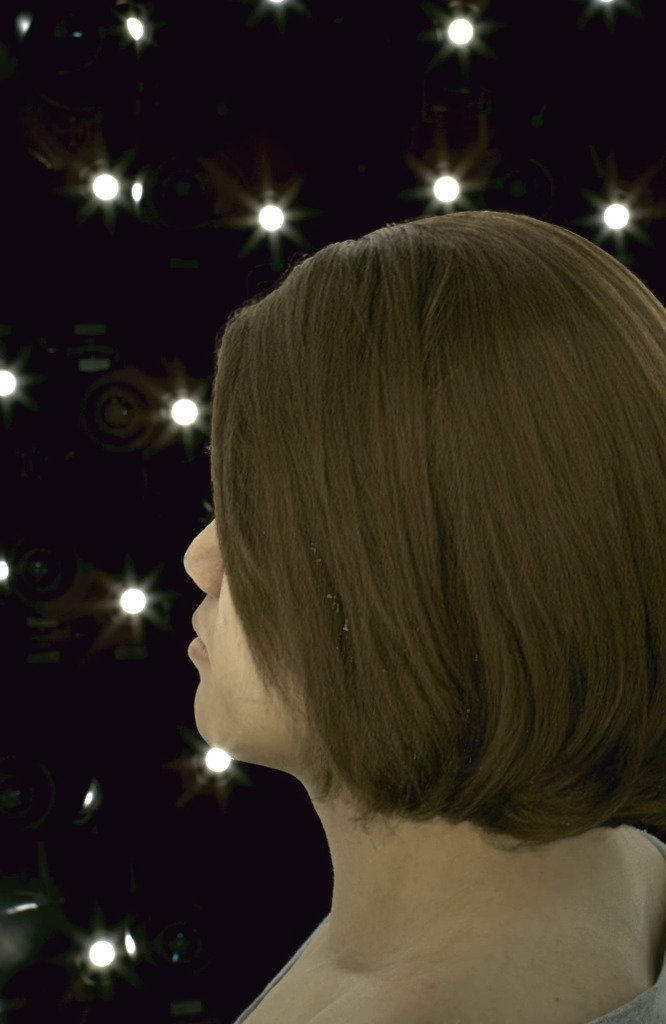}}} &
 {\setlength{\fboxsep}{0pt}\setlength{\fboxrule}{1pt}\fcolorbox{orange}{white}{\adjincludegraphics[width=0.11\textwidth, trim={{0.3\width} {0.2\height} 0 {0.2\height}}, clip, valign=m]{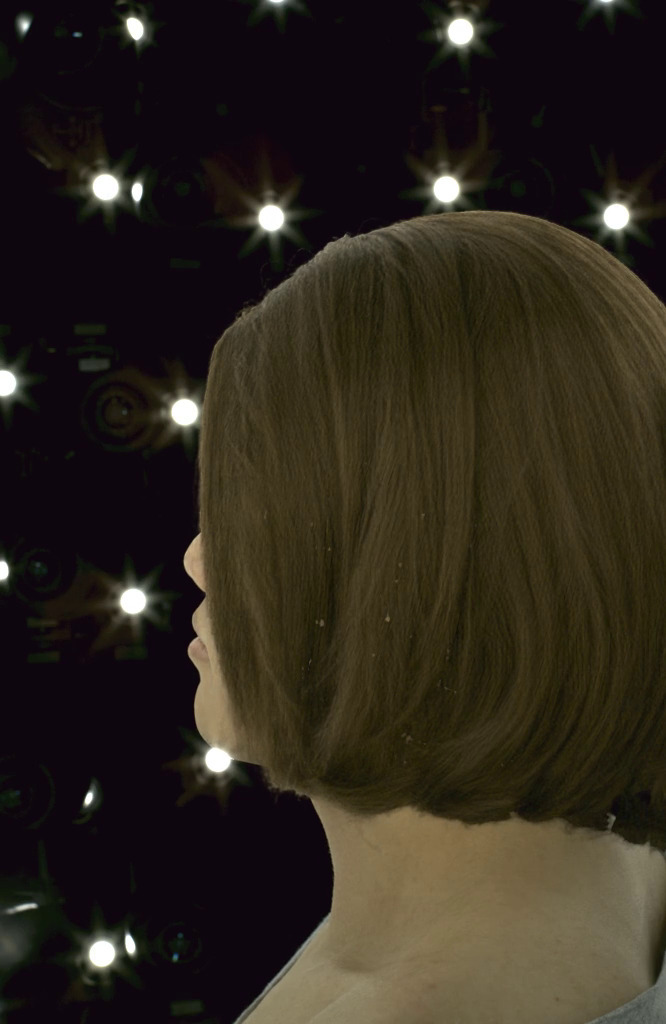}}} \\
 \parbox{0.04\textwidth}{\centering hs2\\\centering mod.} & 
 {\setlength{\fboxsep}{0pt}\setlength{\fboxrule}{1pt}\fcolorbox{orange}{white}{\adjincludegraphics[width=0.11\textwidth, trim={0 {0.2\height} {0.3\width} {0.2\height}}, clip, valign=m]{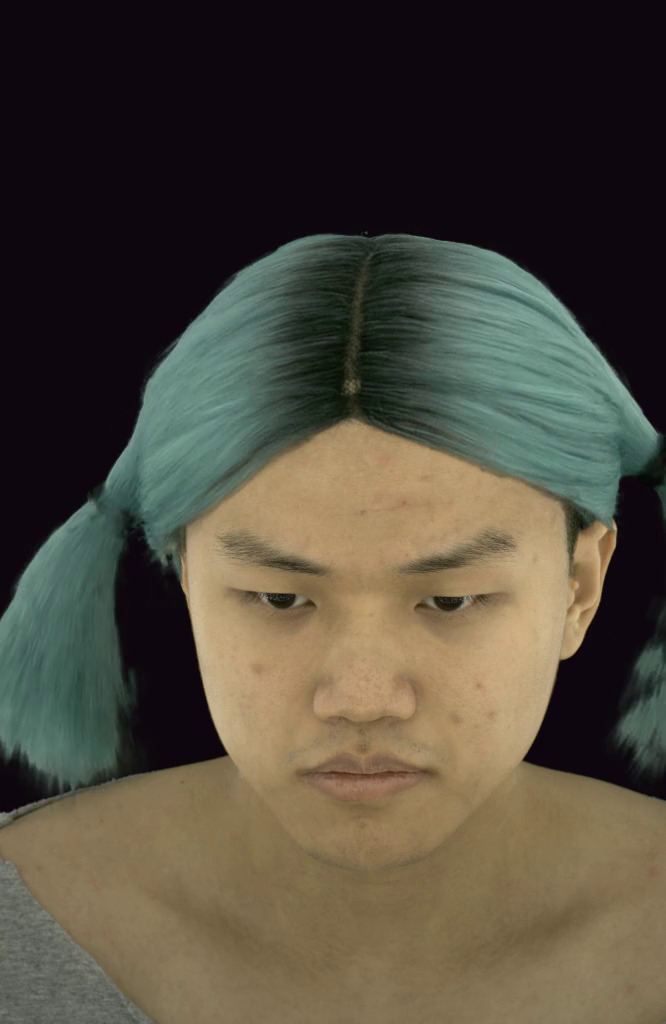}}} &
 {\setlength{\fboxsep}{0pt}\setlength{\fboxrule}{1pt}\fcolorbox{green}{white}{\adjincludegraphics[width=0.11\textwidth, trim={0 {0.2\height} {0.3\width} {0.2\height}}, clip, valign=m]{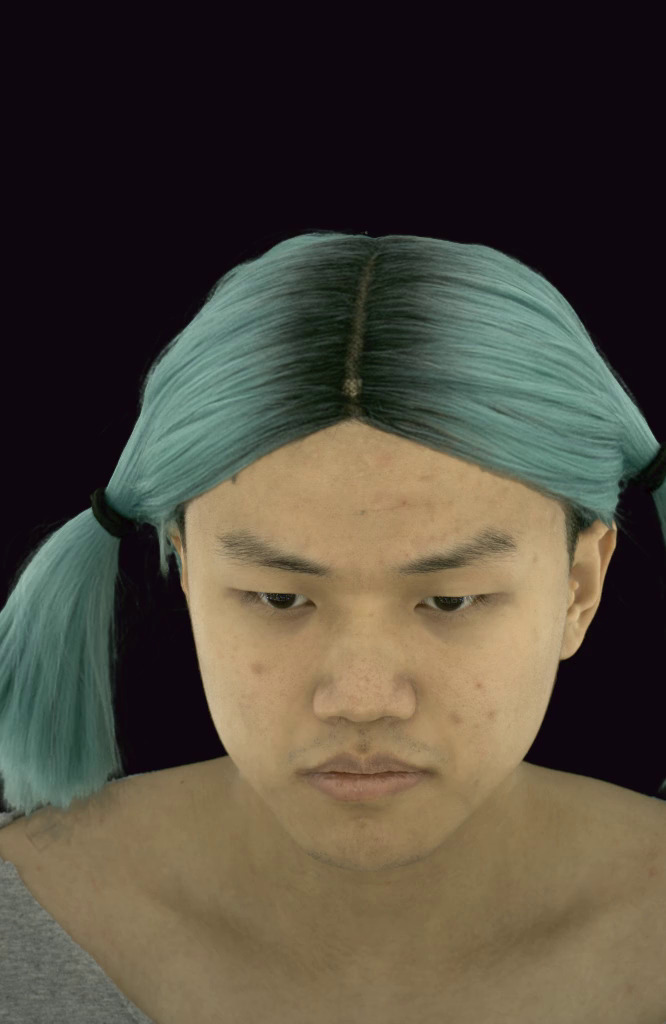}}} &
 {\setlength{\fboxsep}{0pt}\setlength{\fboxrule}{1pt}\fcolorbox{orange}{white}{\adjincludegraphics[width=0.11\textwidth, trim={0 {0.2\height} {0.3\width} {0.2\height}}, clip, valign=m]{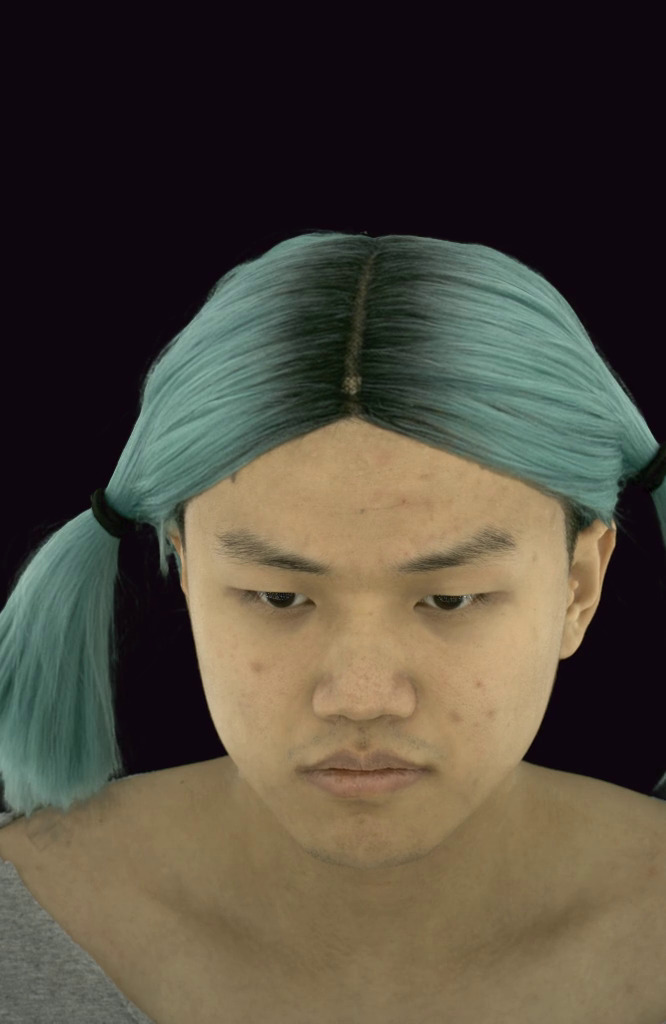}}} &
 {\setlength{\fboxsep}{0pt}\setlength{\fboxrule}{1pt}\fcolorbox{green}{white}{\adjincludegraphics[width=0.11\textwidth, trim={0 {0.2\height} {0.3\width} {0.2\height}}, clip, valign=m]{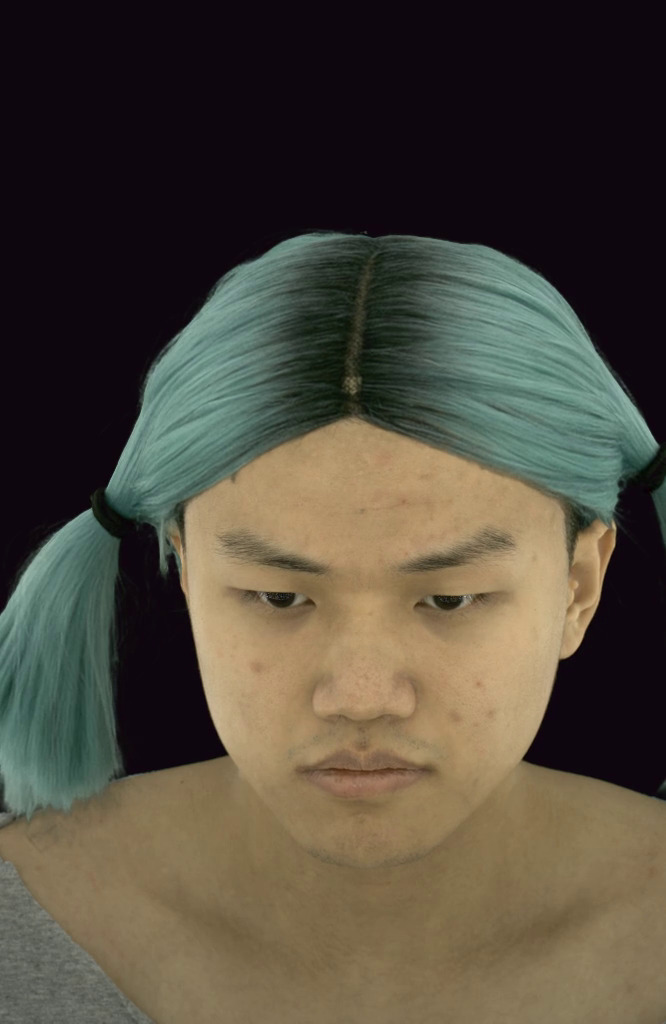}}}
\end{tabular}
\caption{\label{fig2:init_test} \textbf{Effect of the Initialization.} We initialize two different models (hs1 mod. and hs2 mod.) with two different hair point clouds (hs1 and hs2) in two time steps. The green box indicates matched initialization while orange indicates mismatched initialization. Although the mismatched initialization starts shows blurry results at first, the model automatically corrects itself when there is no head motion.}
\end{figure}

We test how robust our model is to the initialization of the hair point cloud.
In Fig.~\ref{fig2:init_test}, we show animation results from models of two different hair styles~\textbf{(hs)} with different initialization hair point clouds.
The results look sharp when the model is matched with the correct hair style, but blurry when we use mismatched hair point clouds for initialization.
However, we find our model self-rectifies and returns to a stable state after a certain number of iterations.
This could be partially due to the model prior stored in the point encoder.

\subsection{Further Ablation on the Point Encoder}

To further study the point encoder $\mathcal{E}$'s ability on denoising the encoding, we tested the encoder with inputs that containing different level of noise. Similar to the study we did in the main paper, we first extract a fixed encoding {\color{red} $z$} and add noise $n$ to it as {\color{cyan}$\hat{z}$}$=${\color{red}$z$}$+n$ and do noise removal as {\color{blue}$\bar{z}$}$=\mathcal{E}(\mathcal{D}(${\color{cyan}$\hat{z}$}$))$. We extend this by adding different level of noise by multiplying $n$ with different scalars. Please refer to the video navigation page for more details.

\subsection{Further Ablation on Novel View Synthesis}

We compare our method with MVP~\cite{steve_mvp} on the longer sequences we captured with scripted head motion. reconstruction related metrics are shown in Tab.\ref{tab_app:nvs_long}. We found that NeRF-based methods can not fit to longer sequences properly. This problem might be due to the large range of motion exhibited in the videos as well as the length of the video. HVH is not applicable because it does not support hair tracking across segmented sequences of different hair motion. Compared to MVP, we achieve better reconstruction accuracy and improved perceptual similarity between rendered image and ground truth with the hair specific modeling in our design.

\begin{table}[tb]
\begin{tabular}{c|cccc|}
\cline{2-5}
     & \multicolumn{1}{c|}{MSE($\downarrow$)} & \multicolumn{1}{c|}{PSNR($\uparrow$)} & \multicolumn{1}{c|}{SSIM($\uparrow$)} & LPIPS($\downarrow$)  \\ \hline
MVP  &         66.21                 &           30.36               &              0.9291             &      0.2830  \\
Ours & \textbf{29.44}                    & \textbf{34.05}                     & \textbf{0.9657}                    & \textbf{0.1109}
\end{tabular}
\caption{\label{tab_app:nvs_long}}
\end{table}

\subsection{Animation on Bald Head Sequences}
We show animation results driven by head motions both on lab conditioned multi-view video captures and in-the-wild phone video captures.
For results on in-the-wild phone video captures, please refer to the supplemental videos.
For phone captures, we ask the participants to face the frontal camera of the phone and perform different head motions.
Then, we apply the face tracking algorithm in ~\cite{ica_chen} to obtain face tracking data that serves as the input to our method.
The initial hair state of the phone animation is sampled from the lab captured dataset.
We find our model generates reasonable motions of hair under head motions like swinging, nodding and etc.

\begin{figure}[tb]
\setlength\tabcolsep{0pt}
\renewcommand{\arraystretch}{0}
\centering
\begin{tabular}{ccc}
 \adjincludegraphics[width=0.16\textwidth, trim={0 0 0 {0.2\height}}, clip]{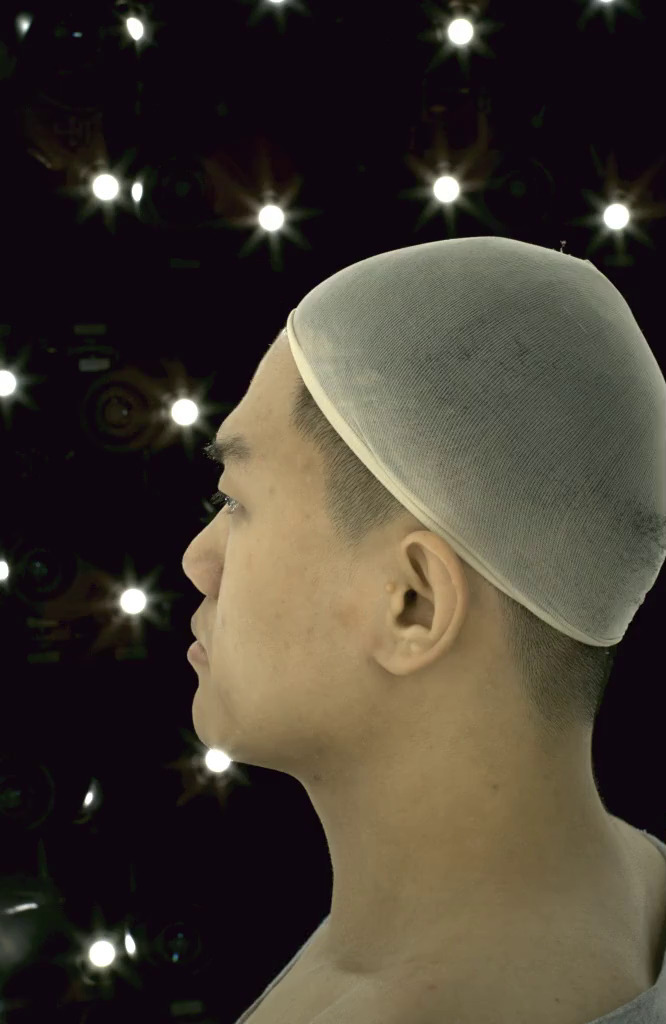} &
 \adjincludegraphics[width=0.16\textwidth, trim={0 0 0 {0.2\height}}, clip]{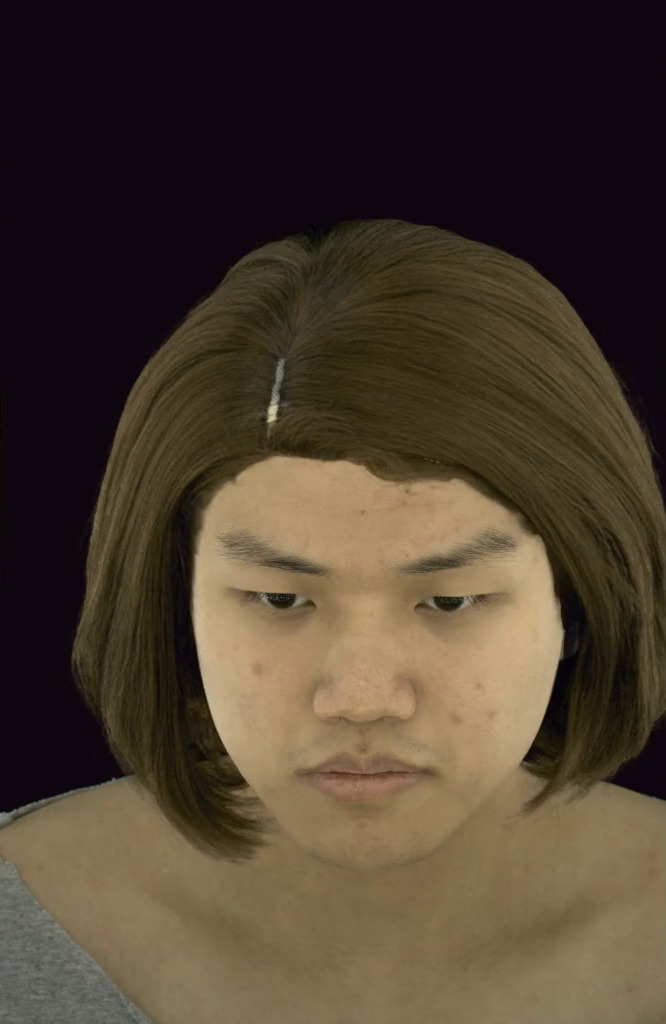} &
 \adjincludegraphics[width=0.16\textwidth, trim={0 0 0 {0.2\height}}, clip]{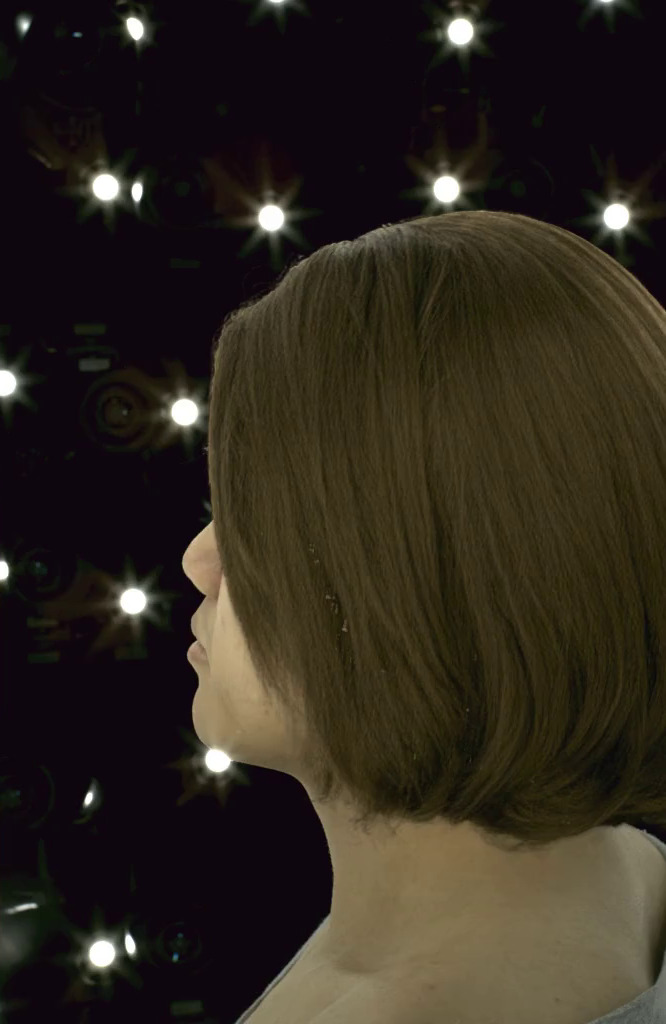} \\
 \adjincludegraphics[width=0.16\textwidth, trim={0 0 0 {0.2\height}}, clip]{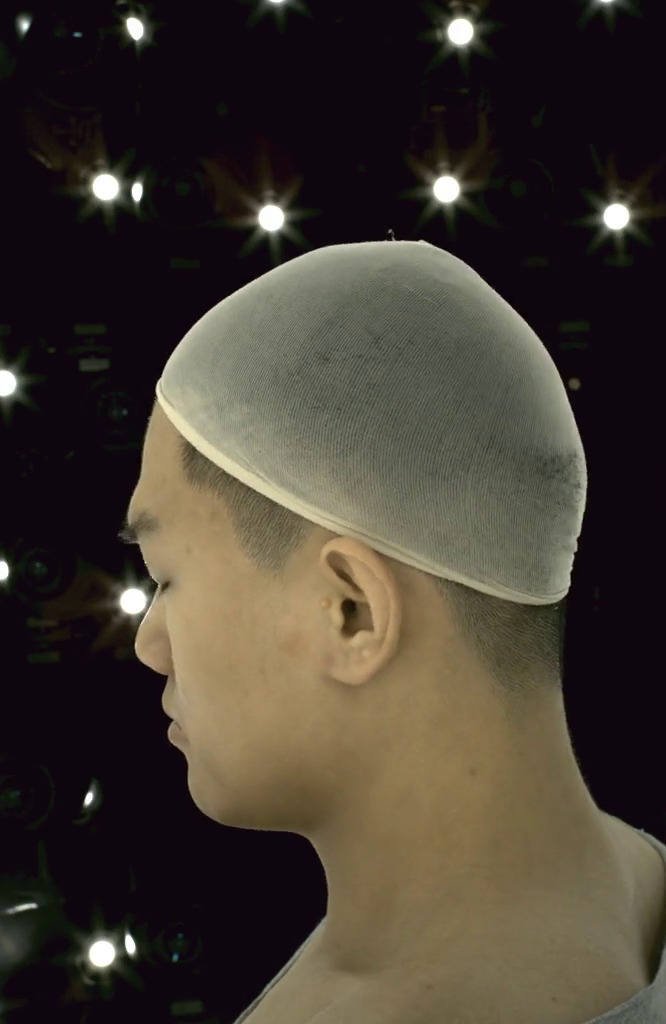} &
 \adjincludegraphics[width=0.16\textwidth, trim={0 0 0 {0.2\height}}, clip]{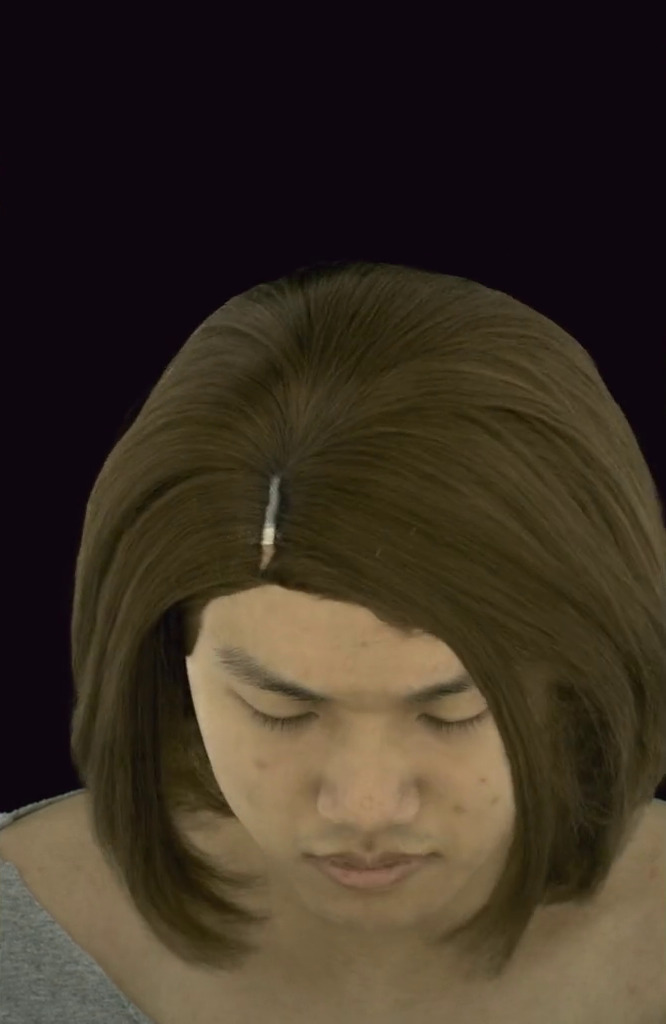} &
 \adjincludegraphics[width=0.16\textwidth, trim={0 0 0 {0.2\height}}, clip]{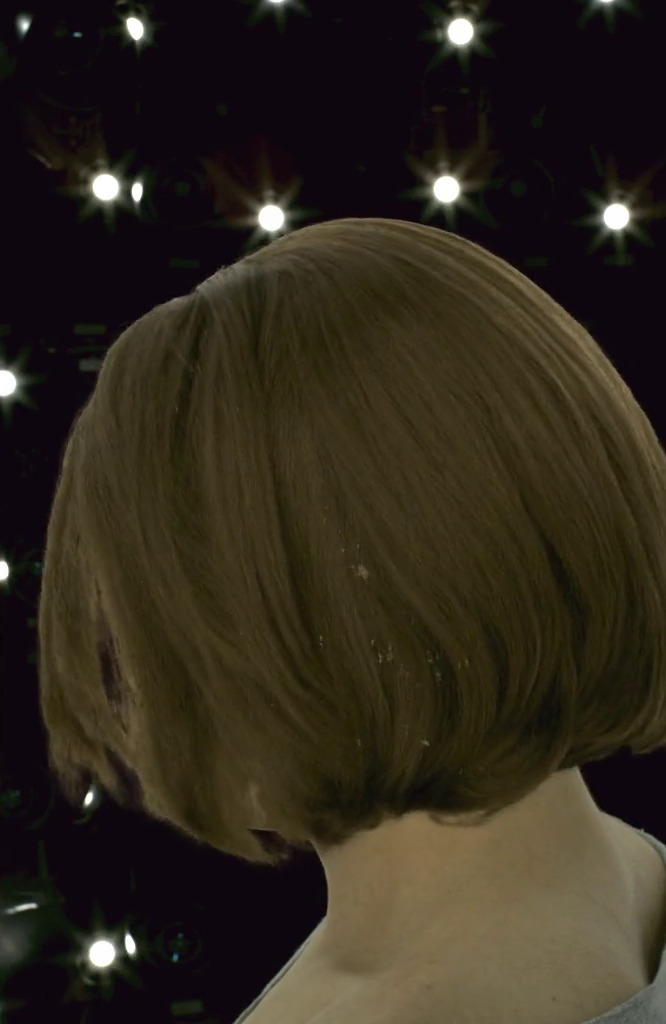} \\
 \adjincludegraphics[width=0.16\textwidth, trim={0 0 0 {0.2\height}}, clip]{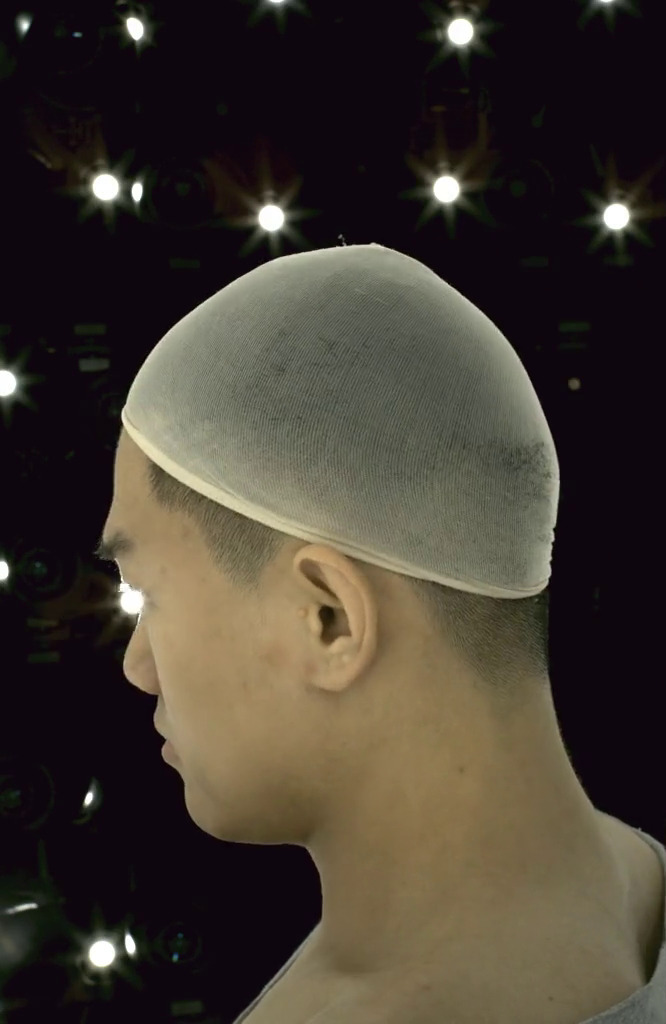} &
 \adjincludegraphics[width=0.16\textwidth, trim={0 0 0 {0.2\height}}, clip]{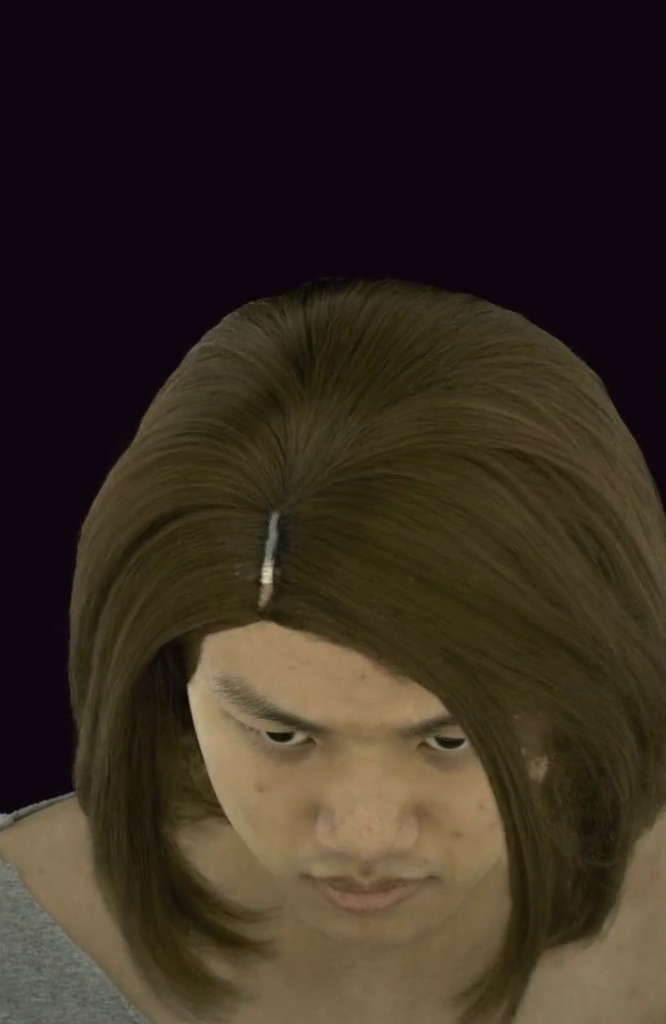} &
 \adjincludegraphics[width=0.16\textwidth, trim={0 0 0 {0.2\height}}, clip]{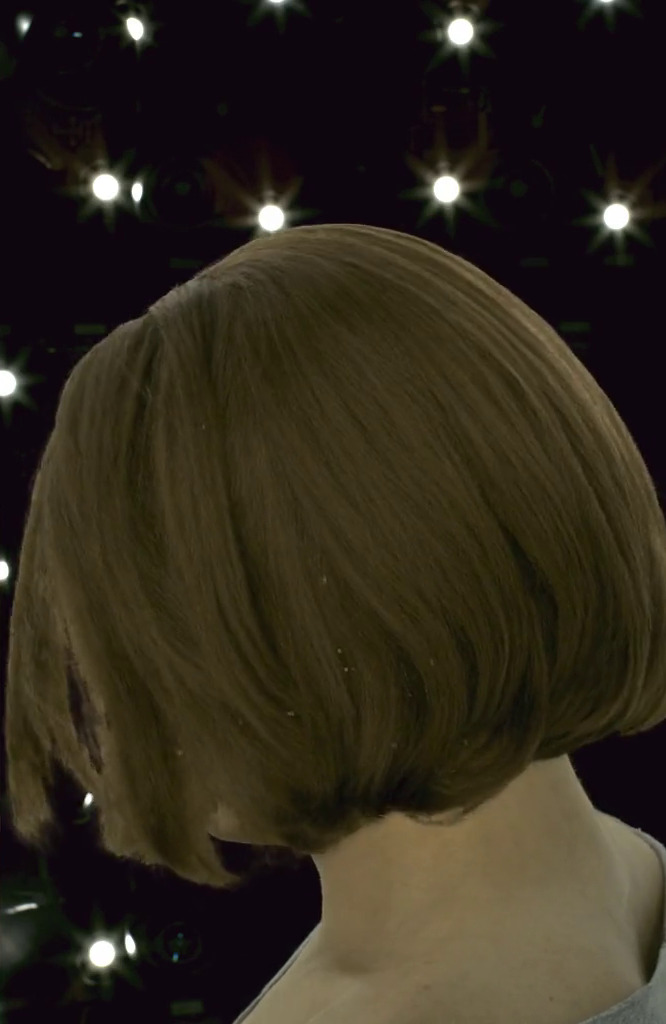} \\
 \adjincludegraphics[width=0.16\textwidth, trim={0 0 0 {0.2\height}}, clip]{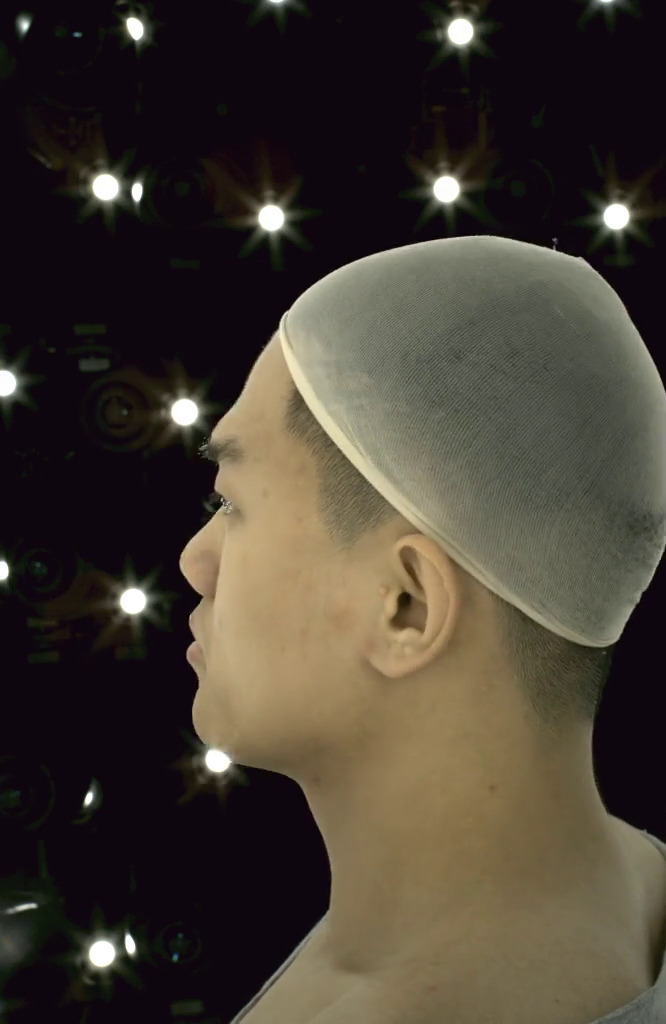} &
 \adjincludegraphics[width=0.16\textwidth, trim={0 0 0 {0.2\height}}, clip]{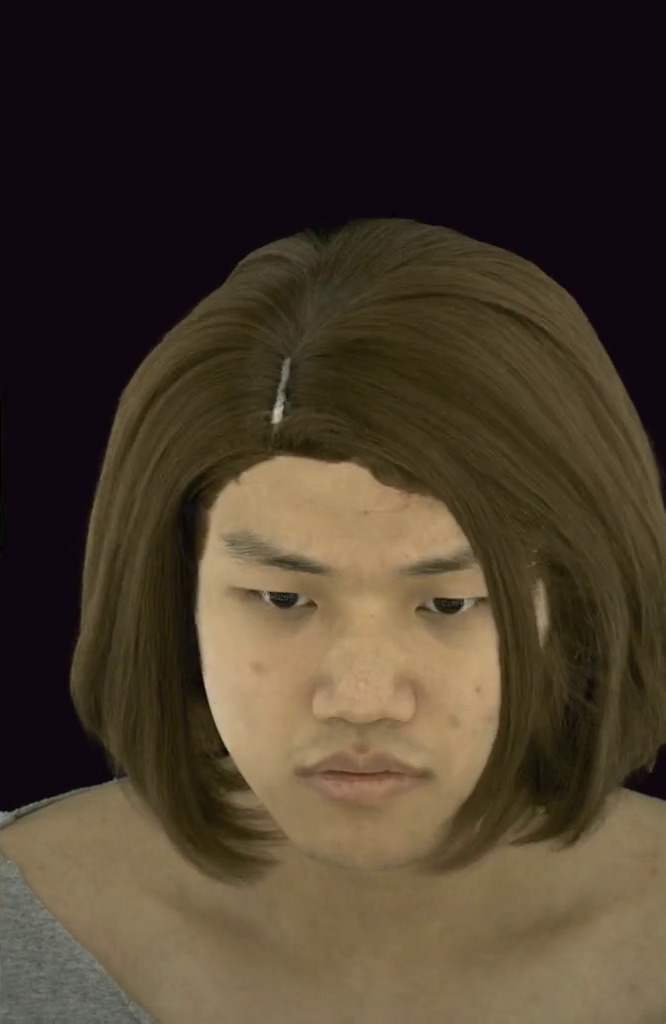} &
 \adjincludegraphics[width=0.16\textwidth, trim={0 0 0 {0.2\height}}, clip]{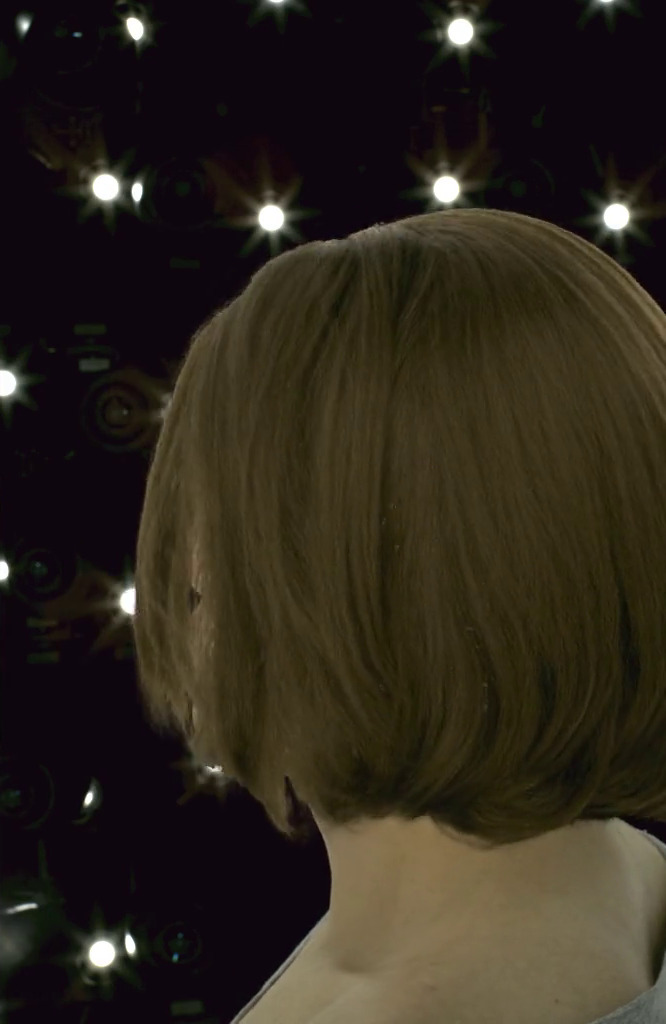} 
\end{tabular}
\caption{\label{fig:anim_mugsy_1}\textbf{Animation on Bald Sequence.}}
\end{figure}

\begin{figure}[tb]
\setlength\tabcolsep{0pt}
\renewcommand{\arraystretch}{0}
\centering
\begin{tabular}{ccc}
 \adjincludegraphics[width=0.48\textwidth, trim={0 0 0 {0.2\height}}, clip]{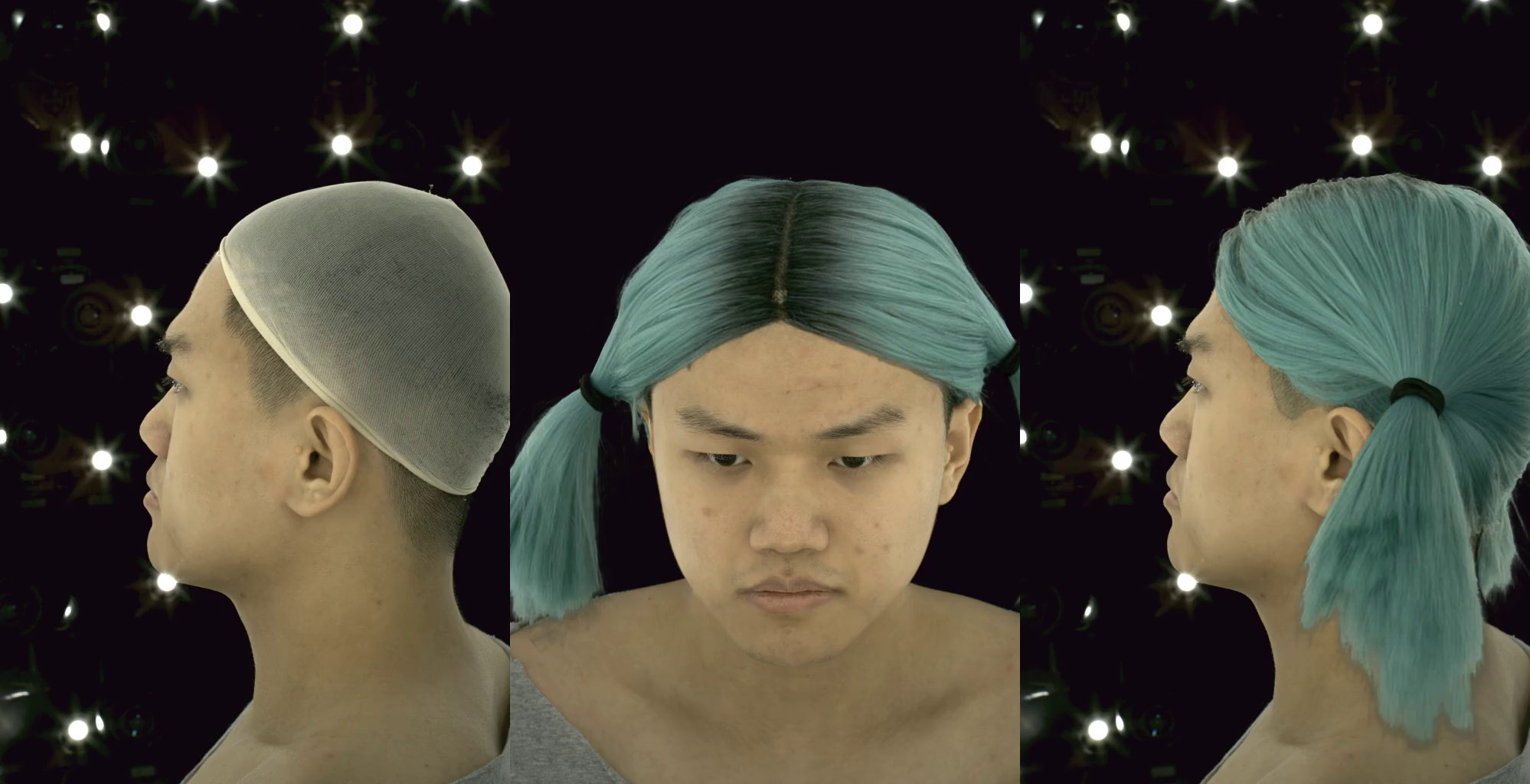} \\
 \adjincludegraphics[width=0.48\textwidth, trim={0 0 0 {0.2\height}}, clip]{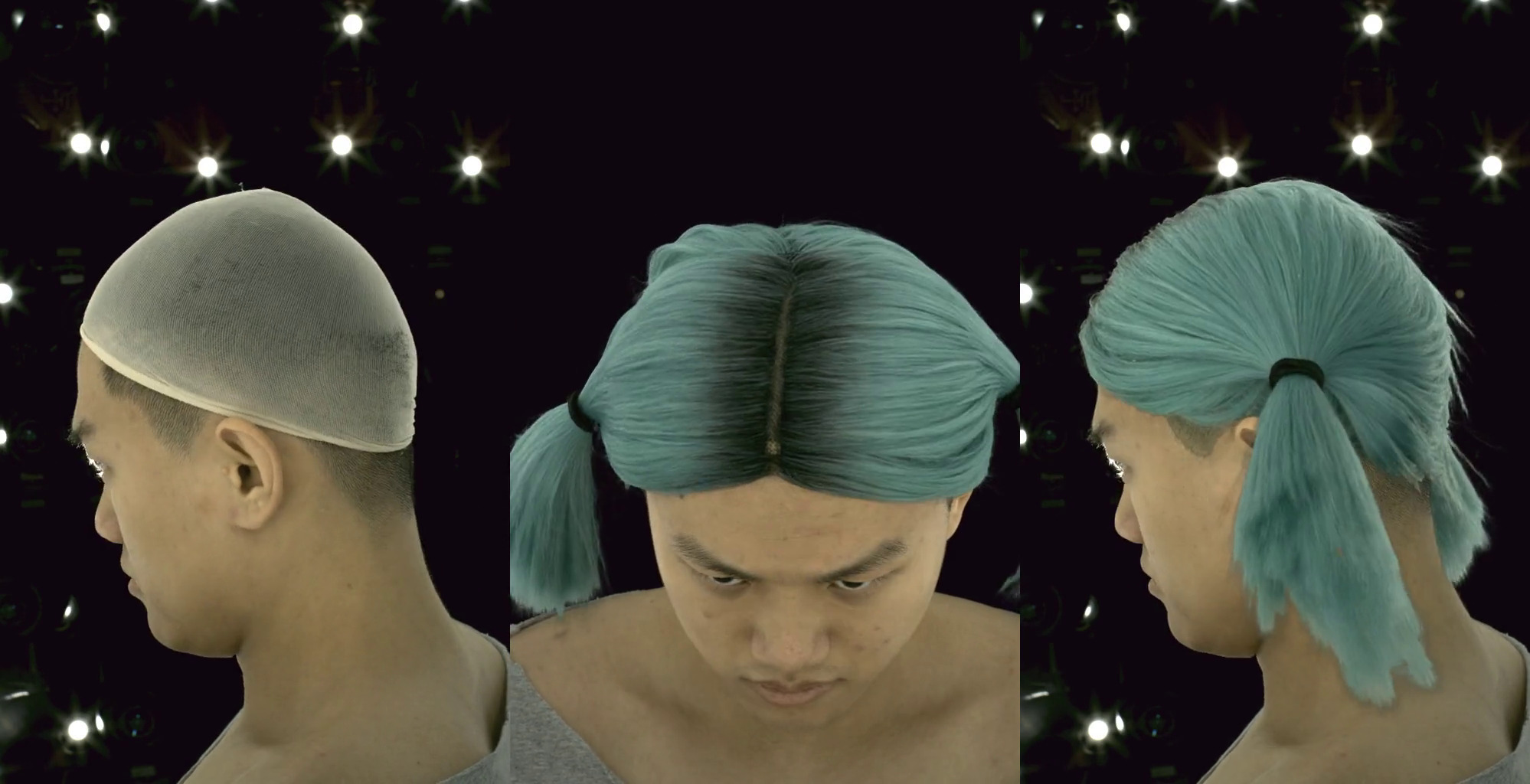} \\
 \adjincludegraphics[width=0.48\textwidth, trim={0 0 0 {0.2\height}}, clip]{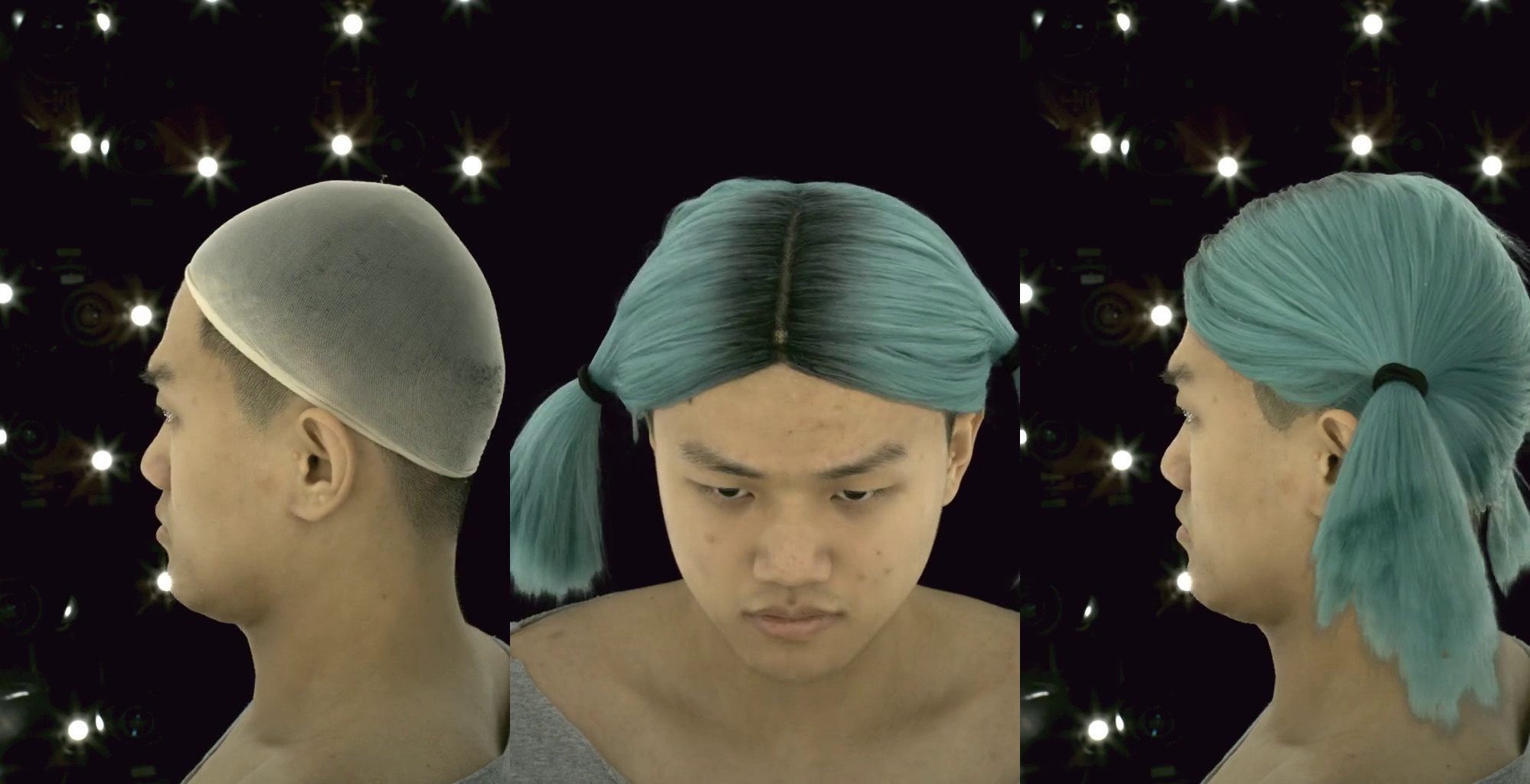} \\
 \adjincludegraphics[width=0.48\textwidth, trim={0 0 0 {0.2\height}}, clip]{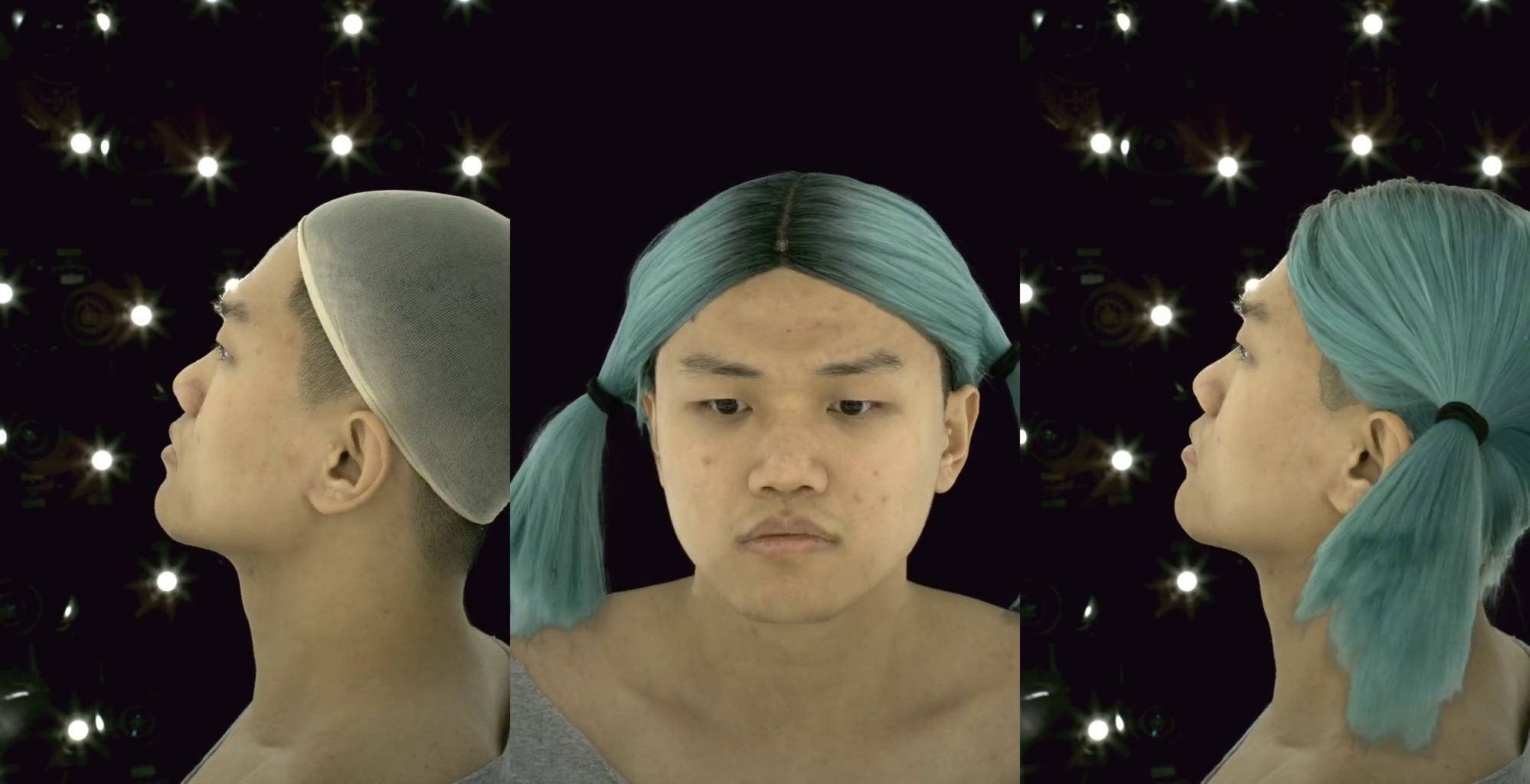}  
\end{tabular}
\caption{\label{fig:anim_mugsy_2}\textbf{Animation on Bald Sequence.}}
\end{figure}

We also test our model on lab conditioned multi-view video captures. As shown in Figs.~\ref{fig:anim_mugsy_1},~\ref{fig:anim_mugsy_2}, our model generates reasonable hair motions with respect to the head motion while preserving multi-view consistency.

\begin{figure}[tb]
\setlength\tabcolsep{0pt}
\renewcommand{\arraystretch}{0}
\centering
\begin{tabular}{ccc}
 \adjincludegraphics[width=0.48\textwidth, trim={0 0 0 {0.1\height}}, clip]{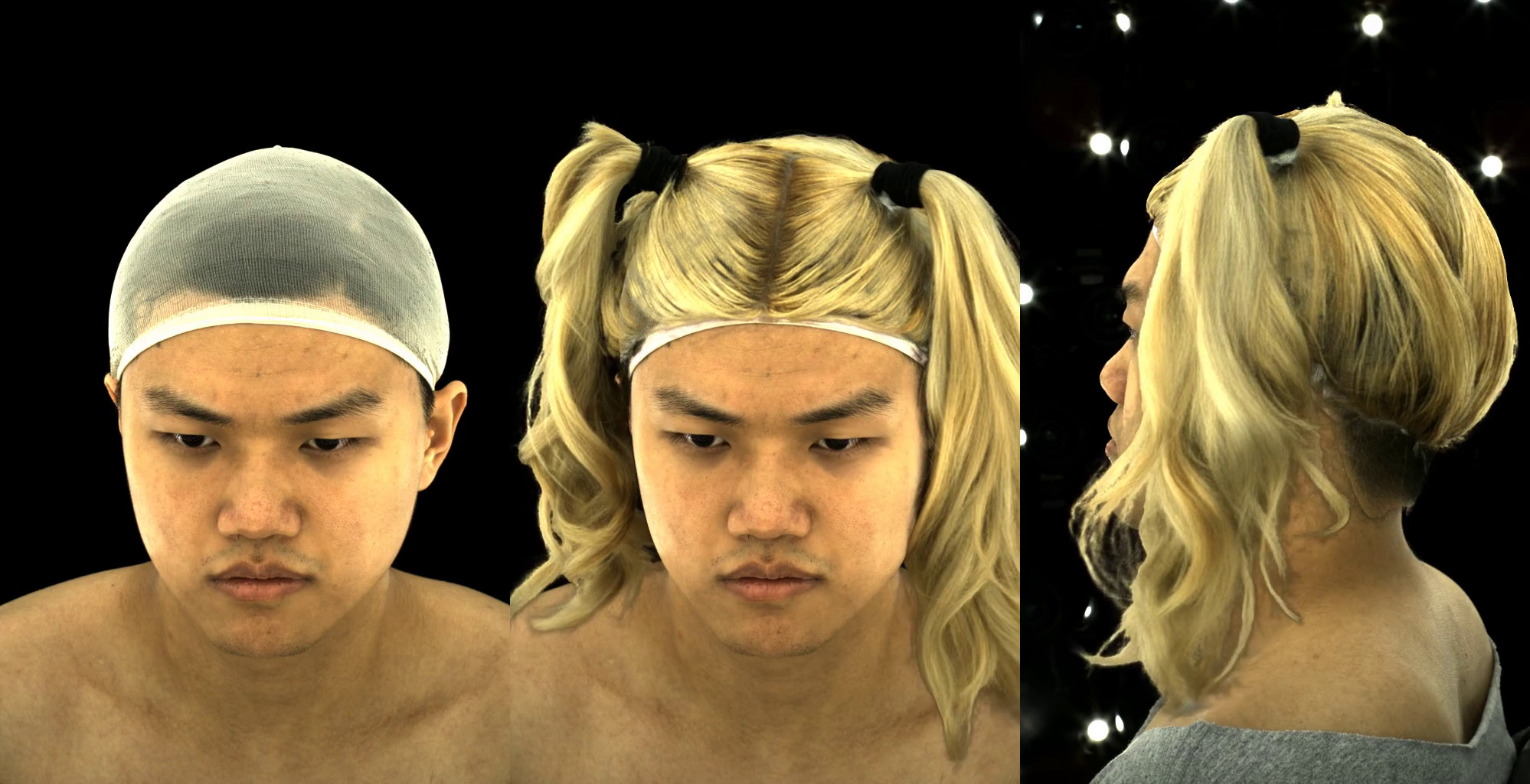} \\
 \adjincludegraphics[width=0.48\textwidth, trim={0 0 0 {0.1\height}}, clip]{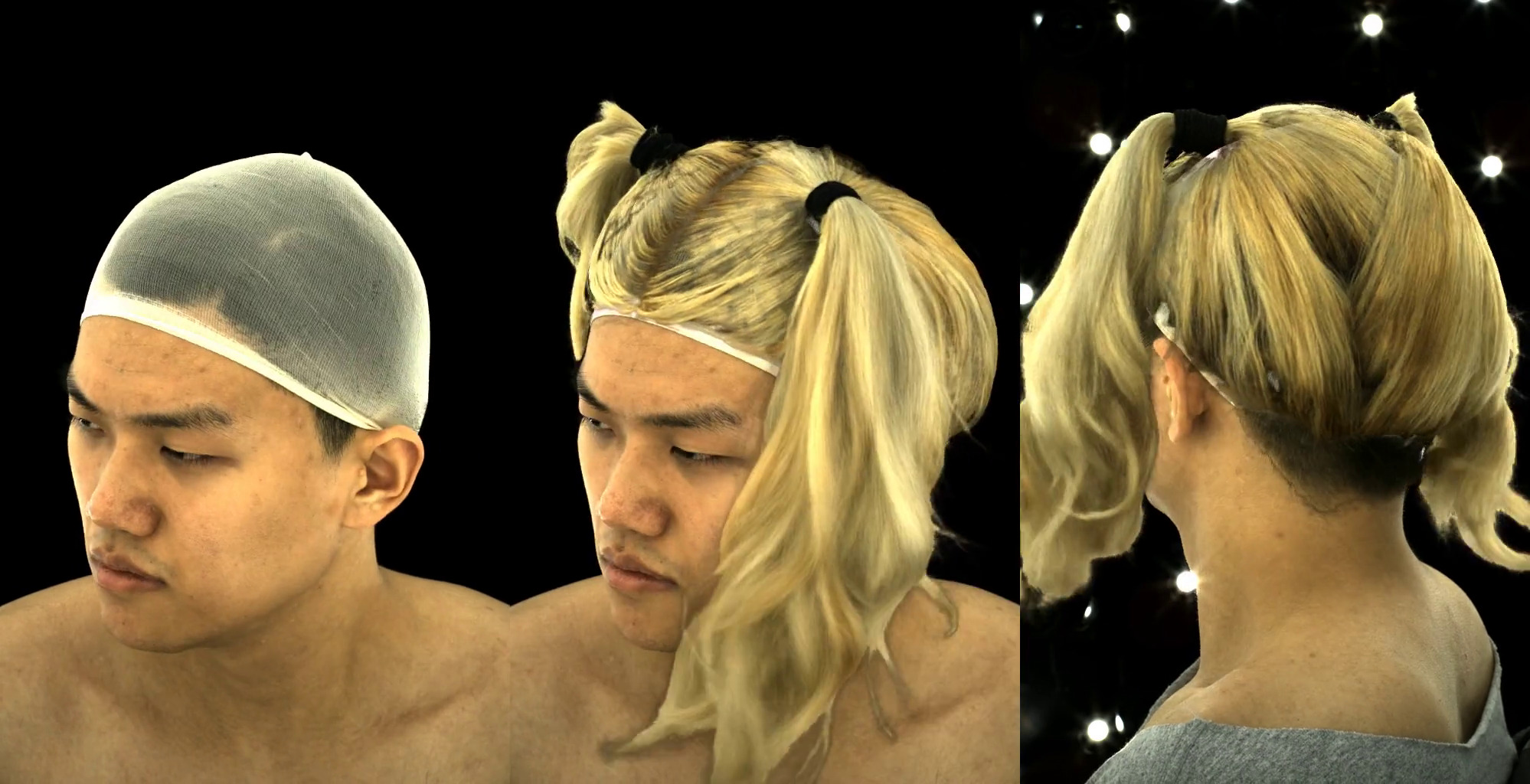} \\
 \adjincludegraphics[width=0.48\textwidth, trim={0 0 0 {0.1\height}}, clip]{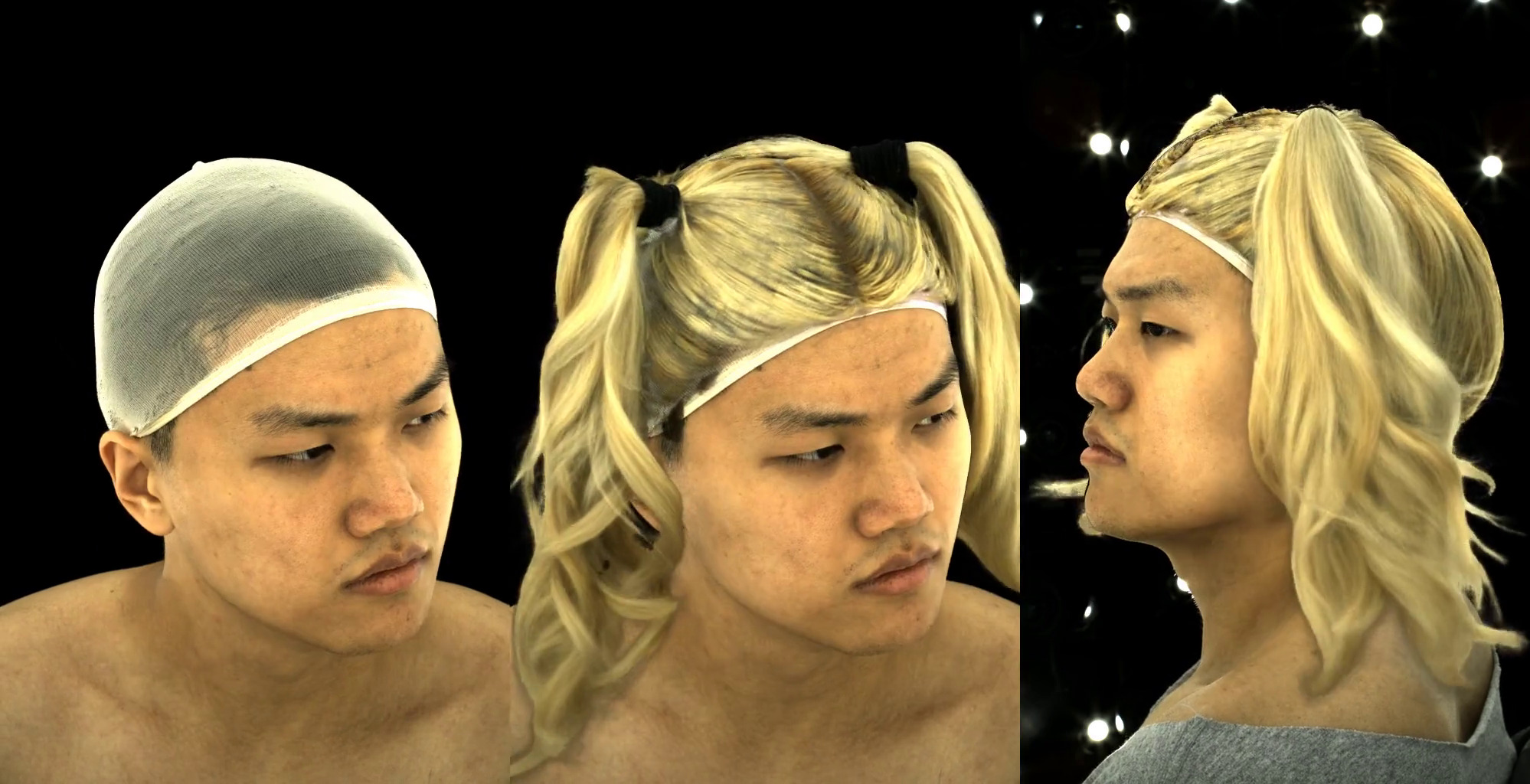} \\
 \adjincludegraphics[width=0.48\textwidth, trim={0 0 0 {0.1\height}}, clip]{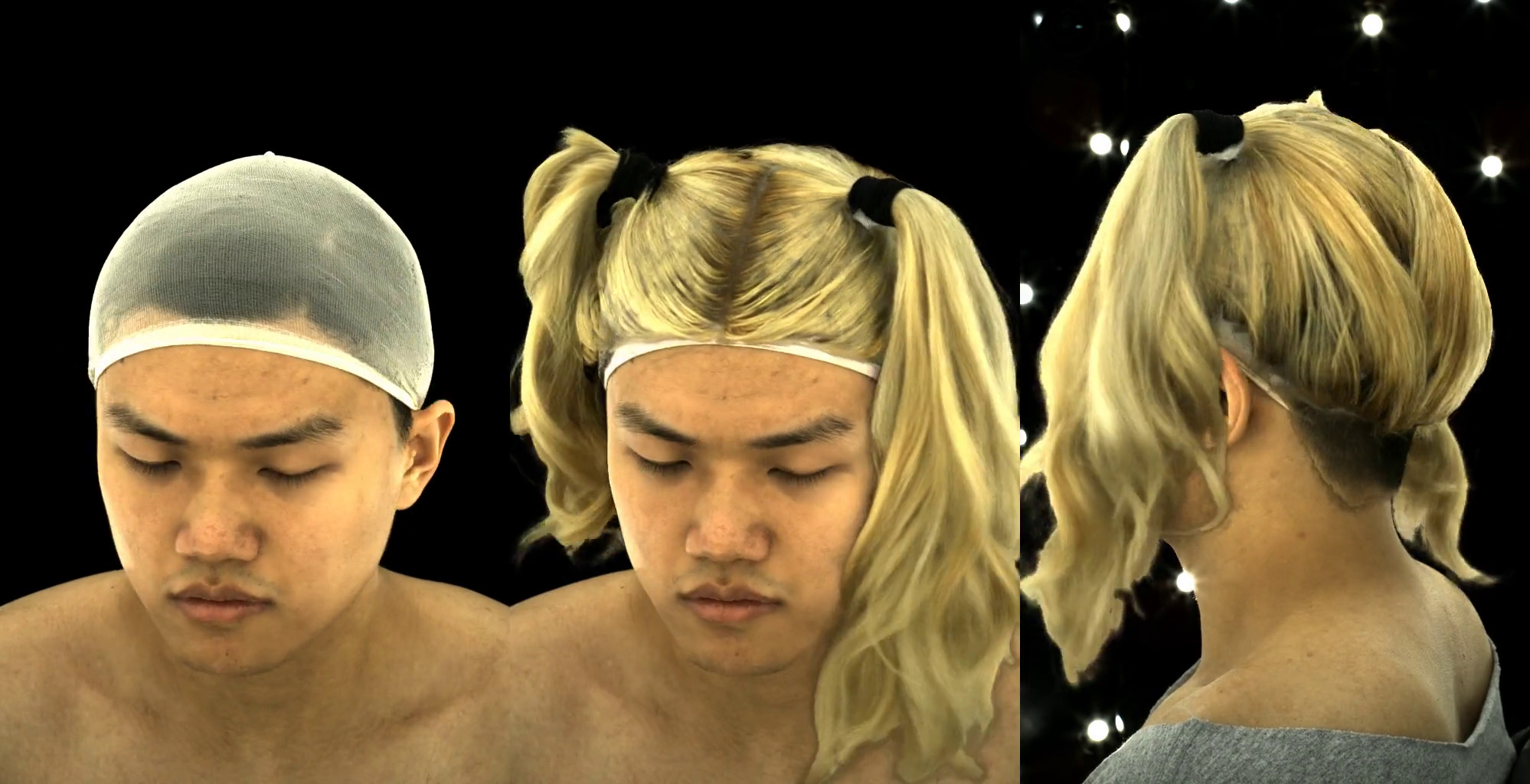}  
\end{tabular}
\caption{\label{fig:anim_mugsy_3}\textbf{Animation on Bald Sequence.}}
\end{figure}

\begin{figure}[tb]
\setlength\tabcolsep{0pt}
\renewcommand{\arraystretch}{0}
\centering
\begin{tabular}{ccc}
 \adjincludegraphics[width=0.48\textwidth, trim={0 0 0 {0.1\height}}, clip]{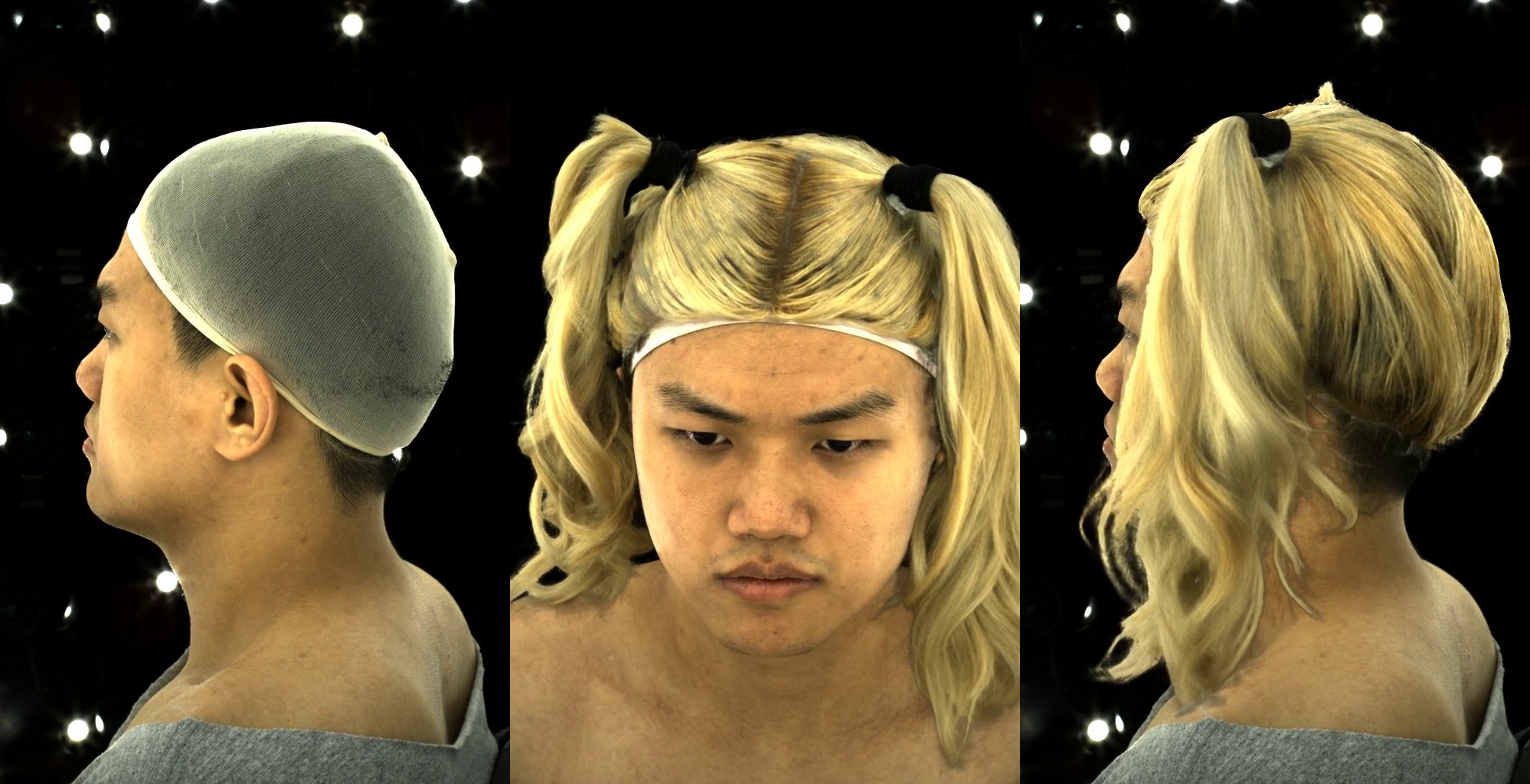} \\
 \adjincludegraphics[width=0.48\textwidth, trim={0 0 0 {0.1\height}}, clip]{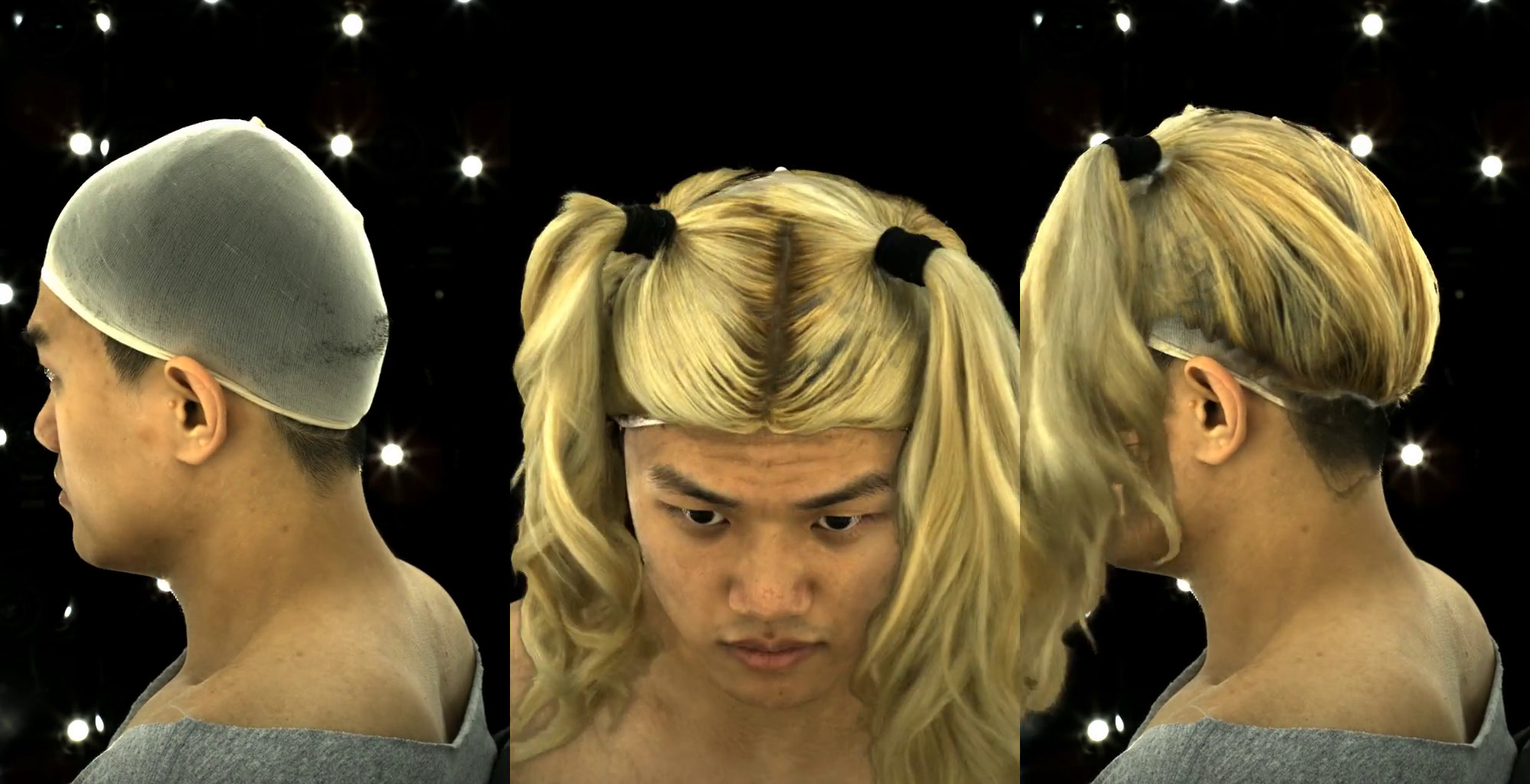} \\
 \adjincludegraphics[width=0.48\textwidth, trim={0 0 0 {0.1\height}}, clip]{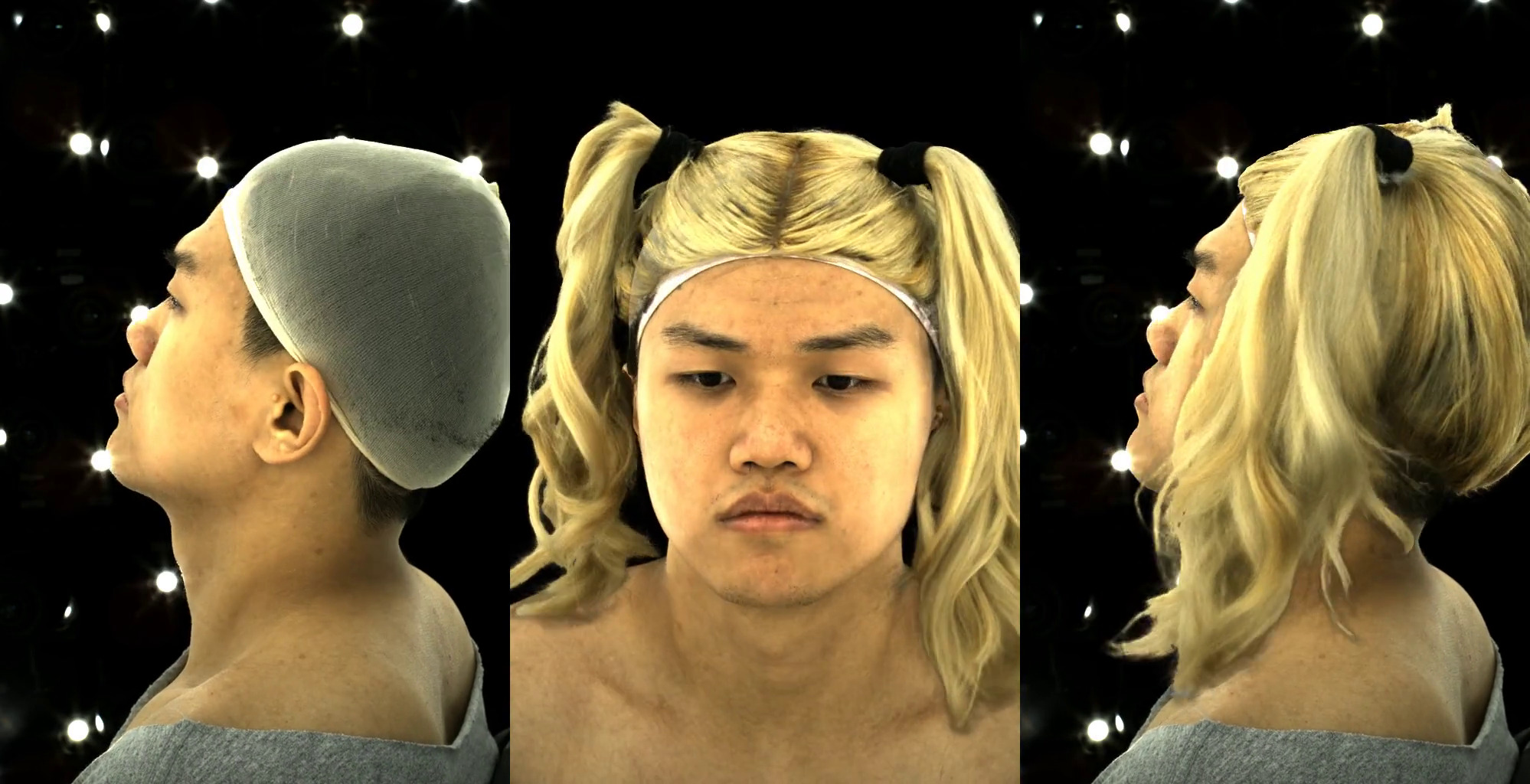} \\
 \adjincludegraphics[width=0.48\textwidth, trim={0 0 0 {0.1\height}}, clip]{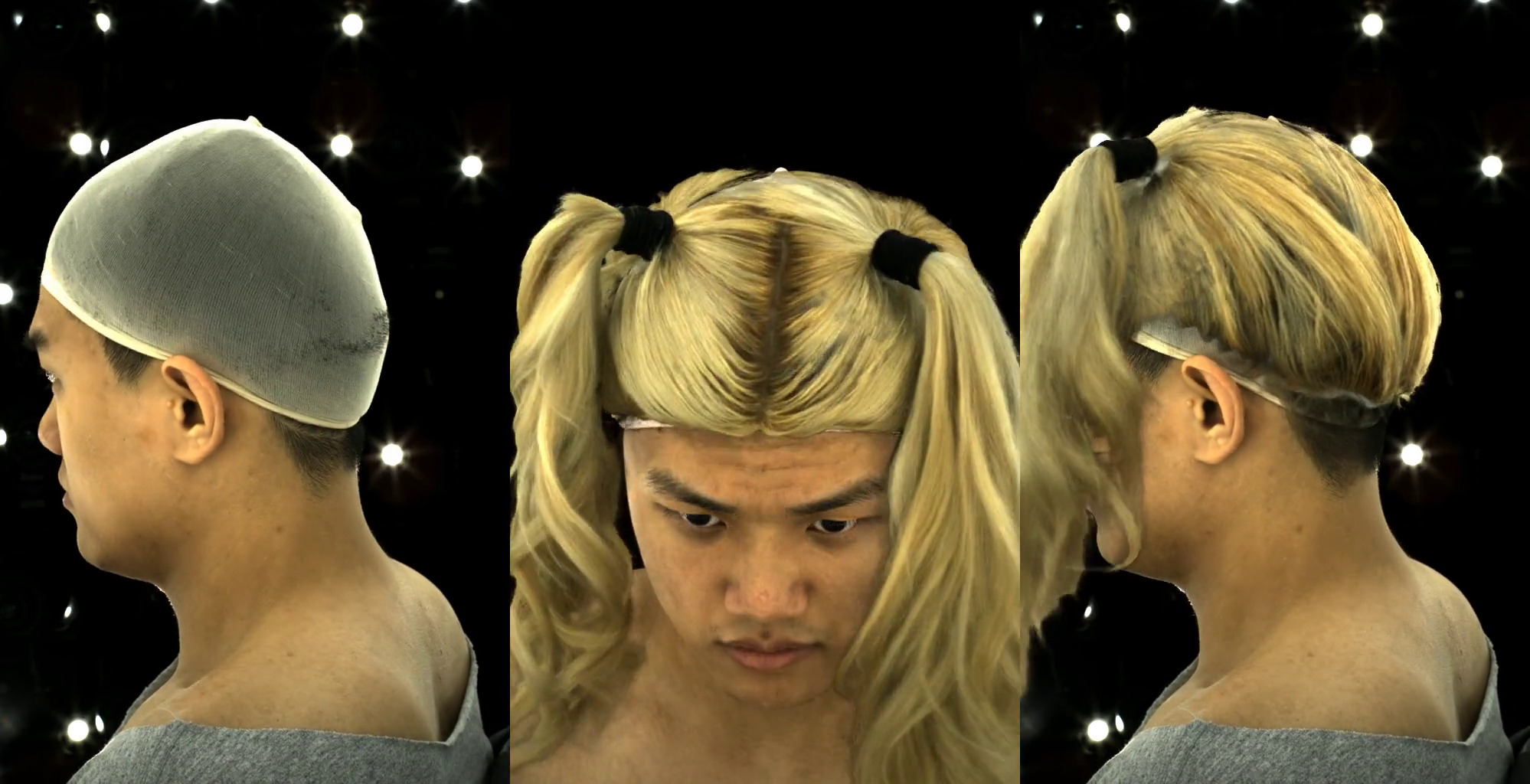}  
\end{tabular}
\caption{\label{fig:anim_mugsy_4}\textbf{Animation on Bald Sequence.}}
\end{figure}

\begin{figure}[tb]
\setlength\tabcolsep{0pt}
\renewcommand{\arraystretch}{0}
\centering
\begin{tabular}{ccc}
 \adjincludegraphics[width=0.48\textwidth, trim={0 0 0 {0.1\height}}, clip]{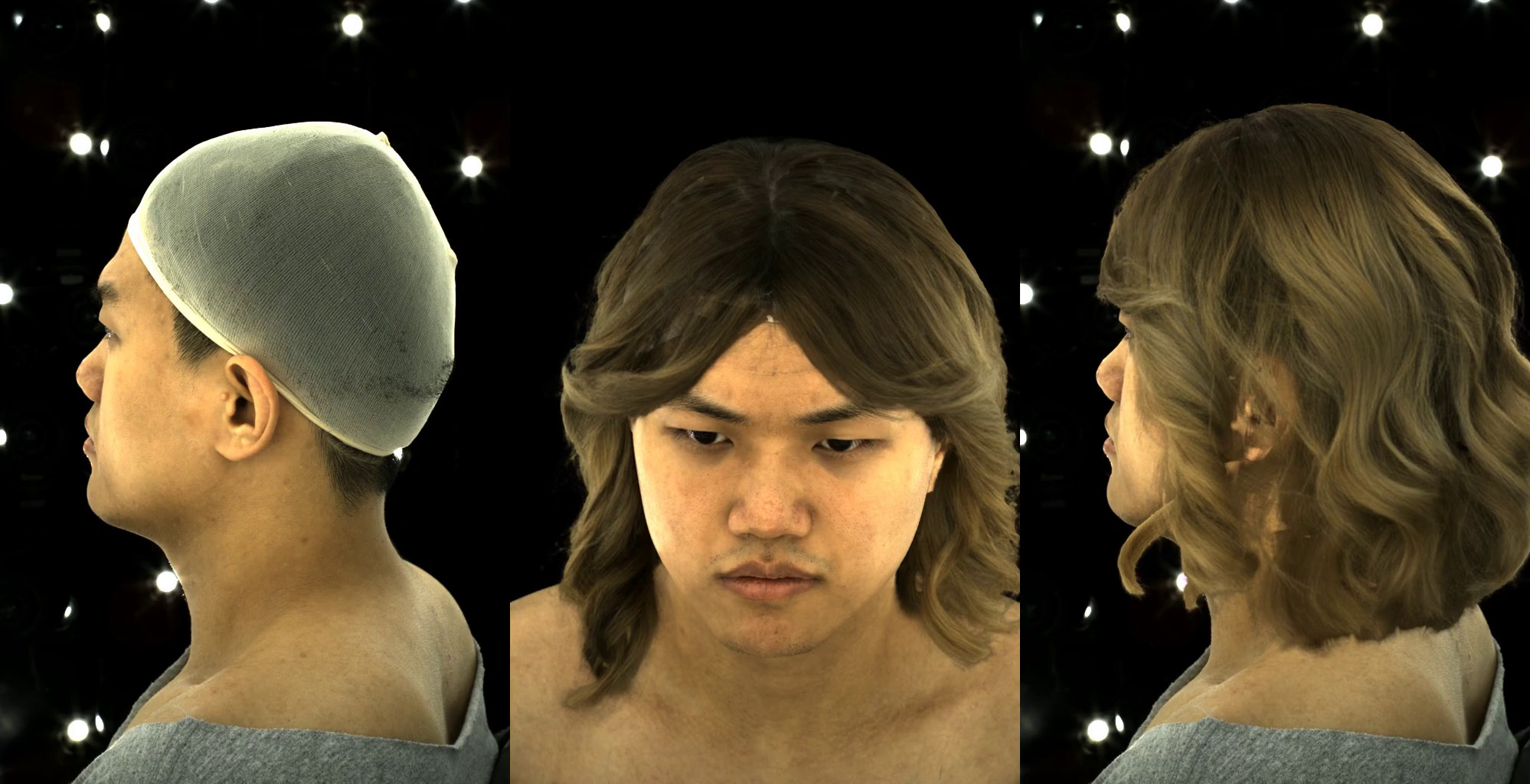} \\
 \adjincludegraphics[width=0.48\textwidth, trim={0 0 0 {0.1\height}}, clip]{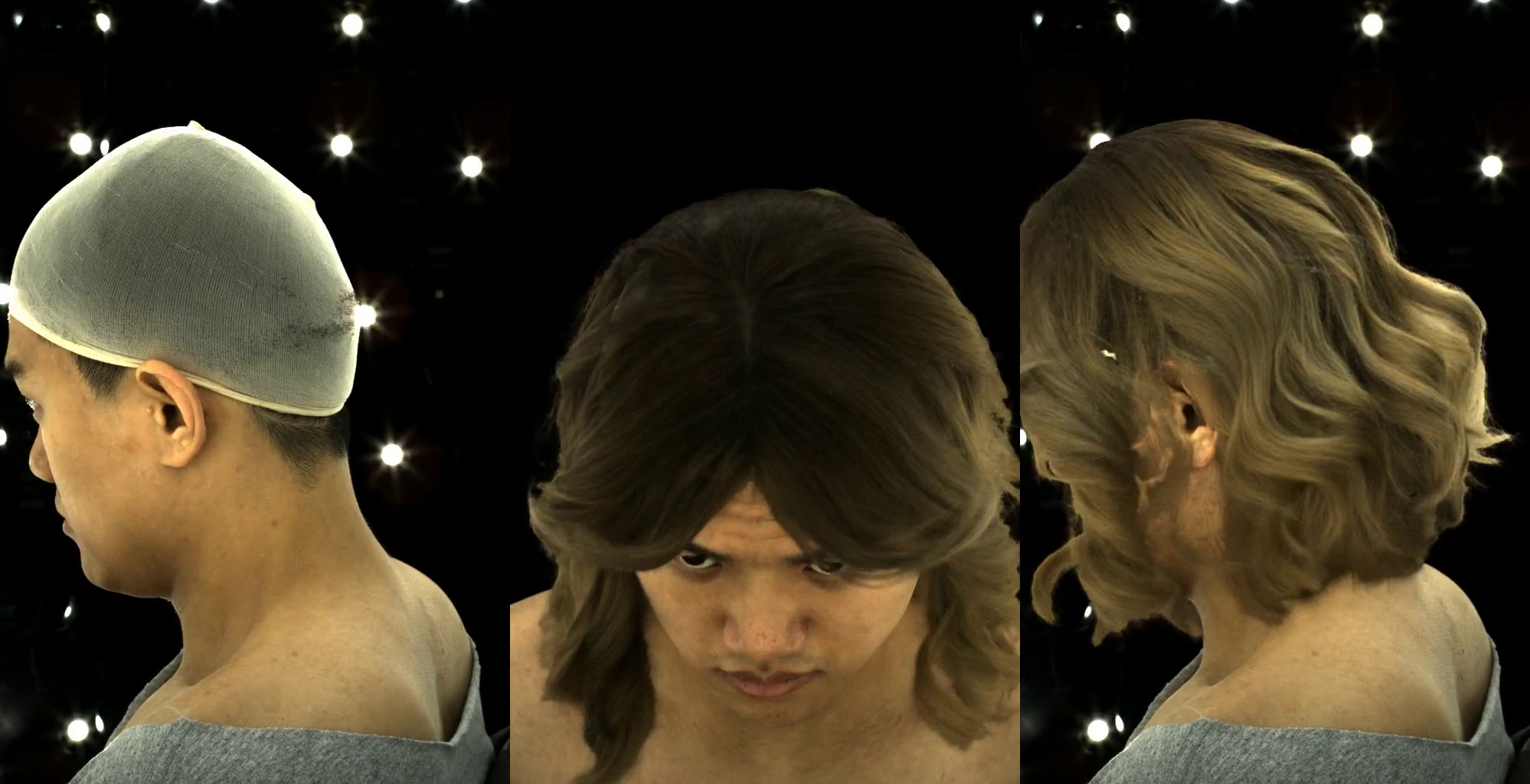} \\
 \adjincludegraphics[width=0.48\textwidth, trim={0 0 0 {0.1\height}}, clip]{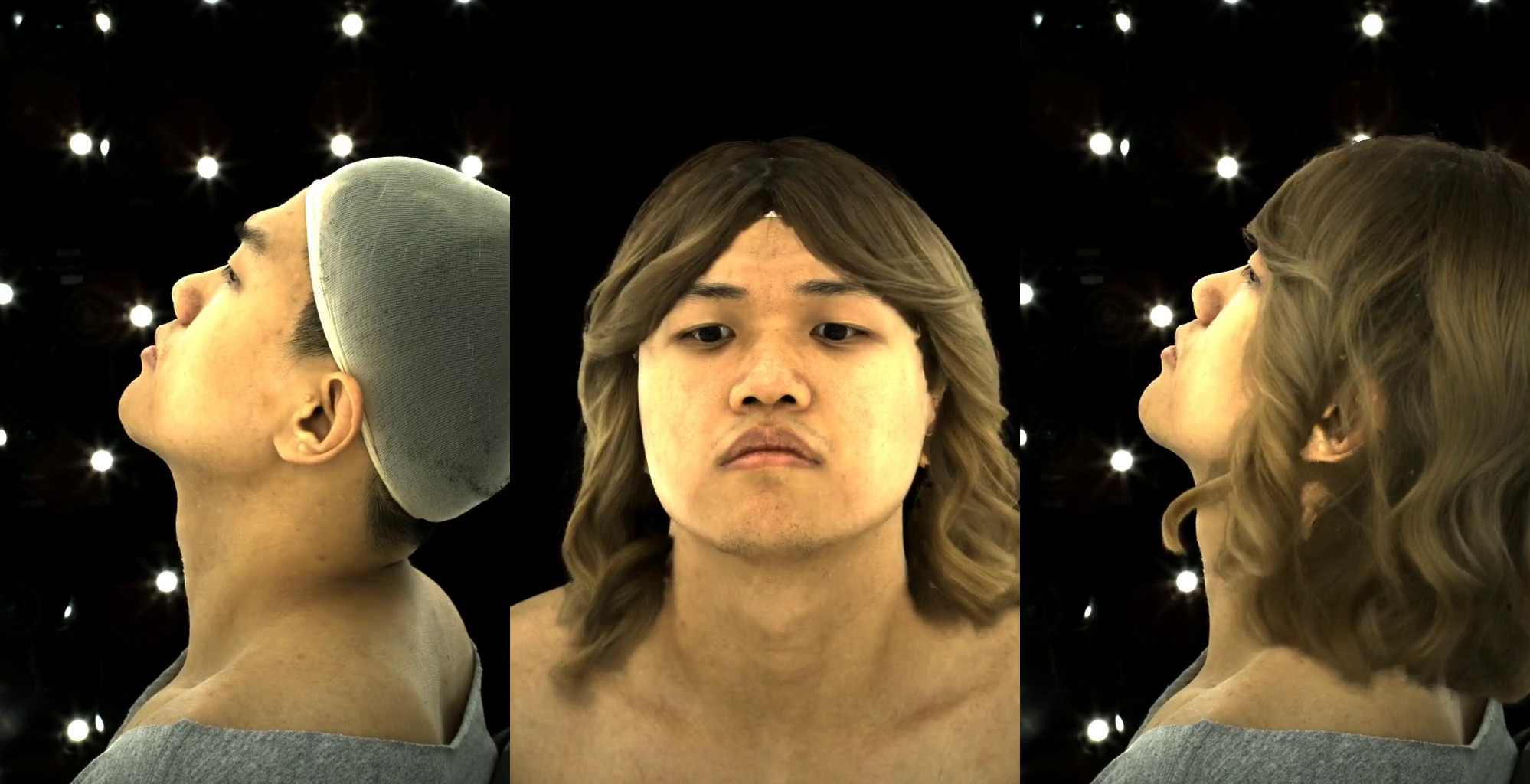} \\
 \adjincludegraphics[width=0.48\textwidth, trim={0 0 0 {0.1\height}}, clip]{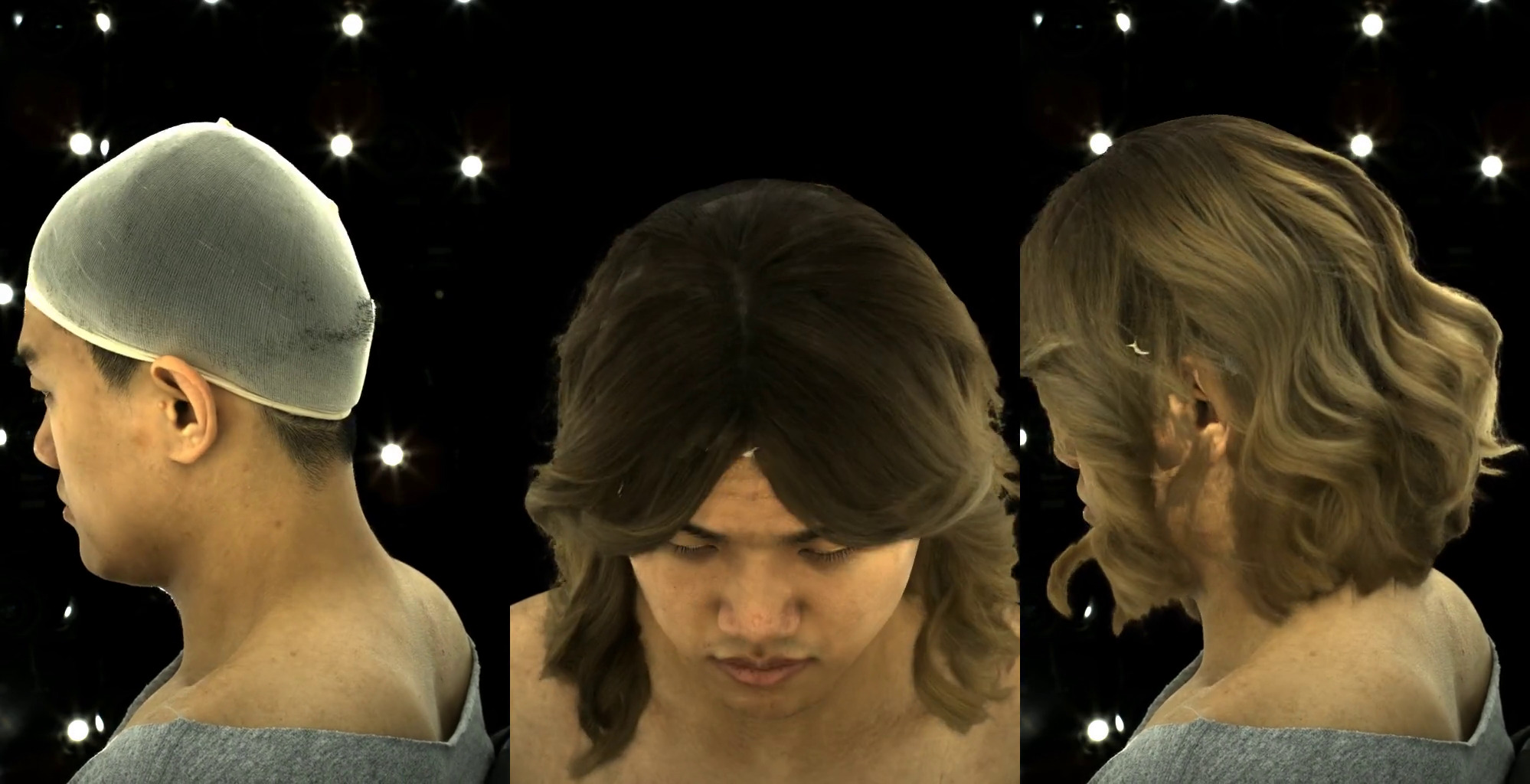}  
\end{tabular}
\caption{\label{fig:anim_mugsy_5}\textbf{Animation on Bald Sequence.}}
\end{figure}

\begin{figure}[tb]
\setlength\tabcolsep{0pt}
\renewcommand{\arraystretch}{0}
\centering
\begin{tabular}{ccc}
 \adjincludegraphics[width=0.48\textwidth, trim={0 0 0 {0.1\height}}, clip]{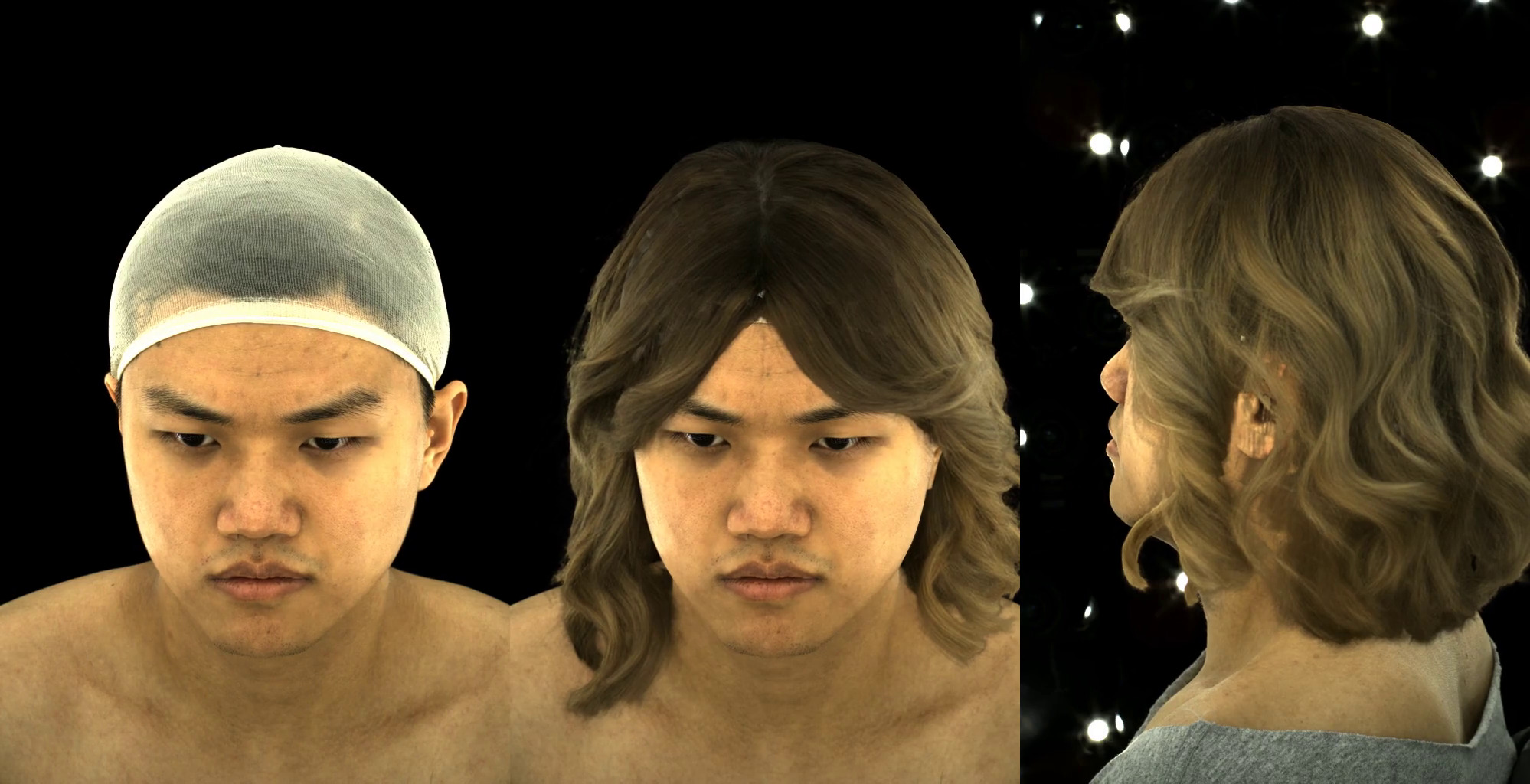} \\
 \adjincludegraphics[width=0.48\textwidth, trim={0 0 0 {0.1\height}}, clip]{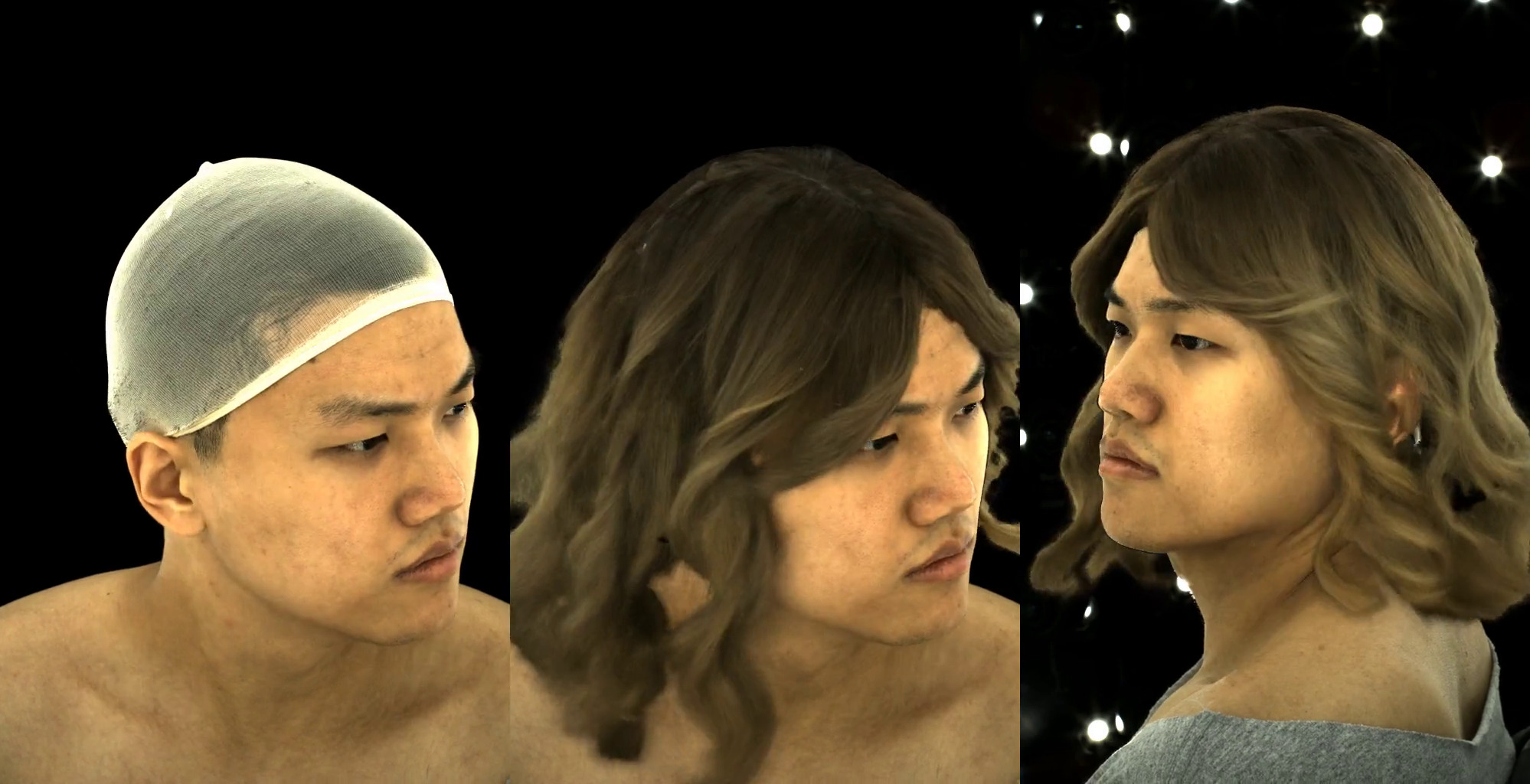} \\
 \adjincludegraphics[width=0.48\textwidth, trim={0 0 0 {0.1\height}}, clip]{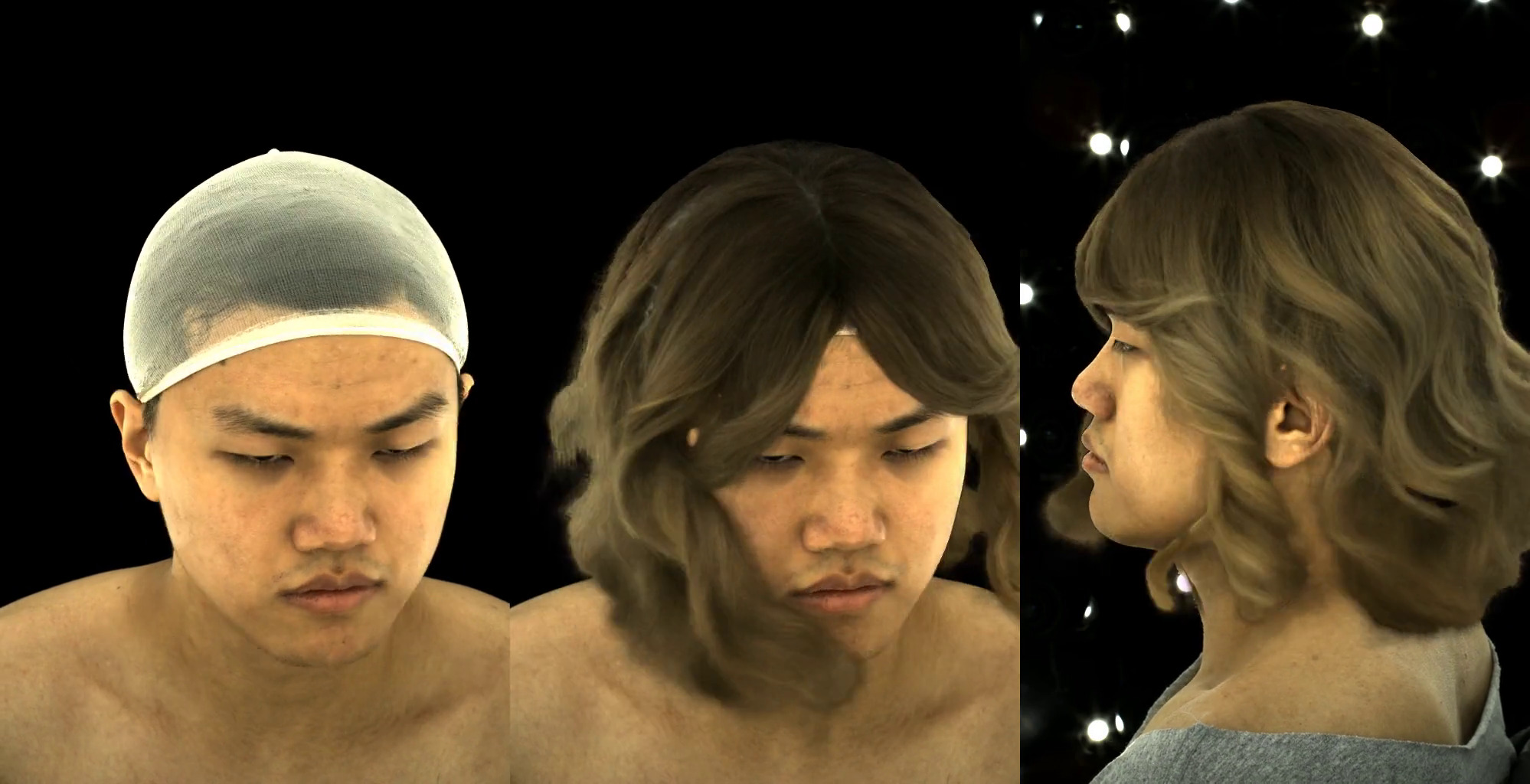} \\
 \adjincludegraphics[width=0.48\textwidth, trim={0 0 0 {0.1\height}}, clip]{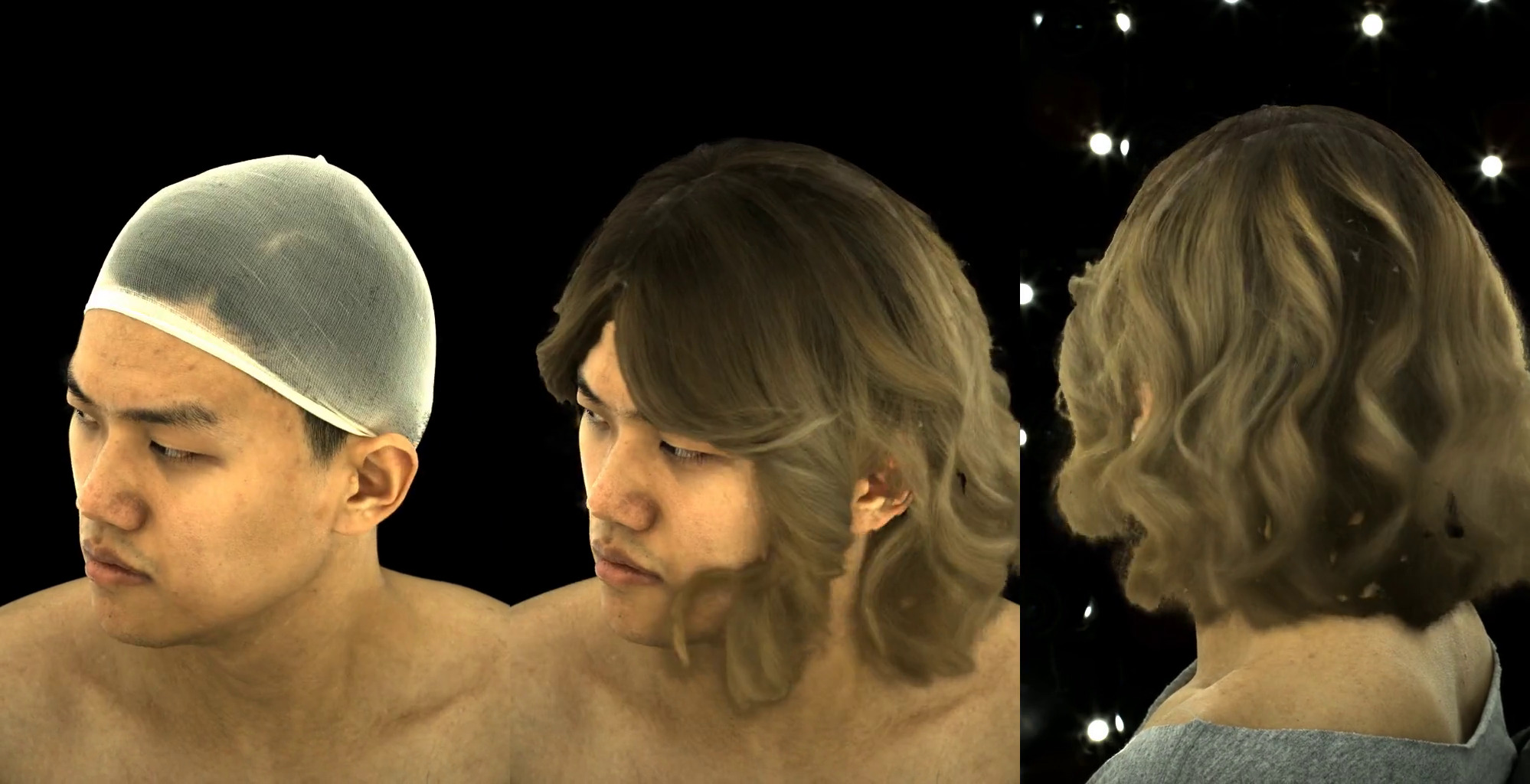}  
\end{tabular}
\caption{\label{fig:anim_mugsy_6}\textbf{Animation on Bald Sequence.}}
\end{figure}